\documentclass[sn-basic]{sn-jnl}

\usepackage{amsmath,amssymb,amsfonts}%
\usepackage{amsthm}%
\usepackage{mathrsfs}%
\usepackage[title]{appendix}%
\usepackage{xcolor}%
\usepackage{textcomp}%
\usepackage{manyfoot}%
\usepackage{booktabs}%
\usepackage{algorithm}%
\usepackage{algorithmicx}%
\usepackage{algpseudocode}%
\usepackage{listings}%
\usepackage{array}
\usepackage{stfloats}
\usepackage{verbatim}
\usepackage{wrapfig}
\usepackage{extarrows}
\usepackage{tikz}
\usepackage{bookmark}
\usepackage{mathtools}
\usepackage{enumitem}
\usepackage{anyfontsize}
\usepackage{bbding}
\usepackage{tcolorbox}
\usepackage{makecell, multirow}
\usepackage{graphicx}
\usepackage{tabulary,array}

\usepackage{lipsum}

\newtheorem{definition}{Definition}%

\def\revision{\textcolor{black}}
\def\revise{\textcolor{black}}

\begin{document}

\title[PROUD: PaRetO-gUided Diffusion Model for Multi-objective Generation]{PROUD: PaRetO-gUided Diffusion Model for Multi-objective Generation}







\author[1,2]{\fnm{Yinghua} \sur{Yao}}\email{eva.yh.yao@gmail.com}
\author*[1,2]{\fnm{Yuangang} \sur{Pan}}\email{yuangang.pan@gmail.com}
\author[1,2]{\fnm{Jing} \sur{Li}}\email{j.lee9383@gmail.com}
\author[1,2]{\fnm{Ivor} \sur{Tsang}}\email{ivor.tsang@gmail.com}
\author[3]{\fnm{Xin} \sur{Yao}}\email{xinyao@ln.edu.hk}

\affil[1]{\orgdiv{Centre for Frontier AI Research}, \orgname{Agency for Science, Technology and Research (A*STAR)}, \orgaddress{\postcode{138632}, \country{Singapore}}}

\affil[2]{\orgdiv{Institute of High Performance Computing}, \orgname{Agency for Science, Technology and Research (A*STAR)}, \orgaddress{\postcode{138632}, \country{Singapore}}}


\affil[3]{\orgdiv{Department of Computing and Decision Sciences}, \orgname{Lingnan University}, \orgaddress{\city{Hong Kong}}}


\abstract{
Recent advancements in the realm of deep generative models focus on generating samples that satisfy multiple desired properties. However, prevalent approaches optimize these property functions independently, thus omitting the trade-offs among them. 
In addition, the property optimization is often improperly integrated into the generative models, resulting in an unnecessary compromise on generation quality (i.e., the quality of generated samples).
To address these issues, we formulate a constrained optimization problem. It seeks to optimize generation quality while ensuring that generated samples reside at the Pareto front of multiple property objectives. Such a formulation enables the generation of samples that cannot be further improved simultaneously on the conflicting property functions and preserves good quality of generated samples.
Building upon this formulation, we introduce the PaRetO-gUided Diffusion model (PROUD), wherein the gradients in the denoising process are dynamically adjusted to enhance generation quality while the generated samples adhere to Pareto optimality. Experimental evaluations on image generation and protein generation tasks demonstrate that our PROUD consistently maintains superior generation quality while approaching Pareto optimality across multiple property functions compared to various baselines.
}




\maketitle

\section{Introduction}\label{sec1}

Deep generative models have been developing prosperously over the last decade, with advances in variational autoencoders~\citep{DBLP:journals/corr/KingmaW13}, generative adversarial networks~\citep{goodfellow2014generative,zhang2023robust}, normalizing flows~\citep{papamakarios2021normalizing}, energy-based models~\citep{song2021train}, and diffusion models~\citep{song2019generative,ho2020denoising}. Particularly, controllable generative models can generate samples that satisfy multiple properties of interest, showing great promise in various applications, such as material design~\citep{jin2020multi,tagasovska2022pareto} and controlled text/image generation~\citep{dathathri2019plug,liao2020towards}. {These properties of interest vary depending on the specific application domains.} For example, in protein design, the properties can refer to specified structural or functional characteristics, such as solubility or binding affinity~\citep{watson2023novo}. In image generation, the properties can refer to certain attributes or features, such as specified hairstyle \& makeup~\citep{wang2023stylediffusion}, or specified color patches~\citep{liu2021sampling}. In addition, it is considered imperative that generated samples should reside in the same data manifold\footnote{This relates to the manifold hypothesis that many real-world high-dimensional datasets lie on low-dimensional latent manifolds in the high-dimensional space~\citep{fefferman2016testing}} as training samples for data naturalness concerns~\citep{gruver2023protein}.

\begin{figure}
    \begin{minipage}{0.64\linewidth}
	\centerline{\includegraphics[width=1\textwidth]{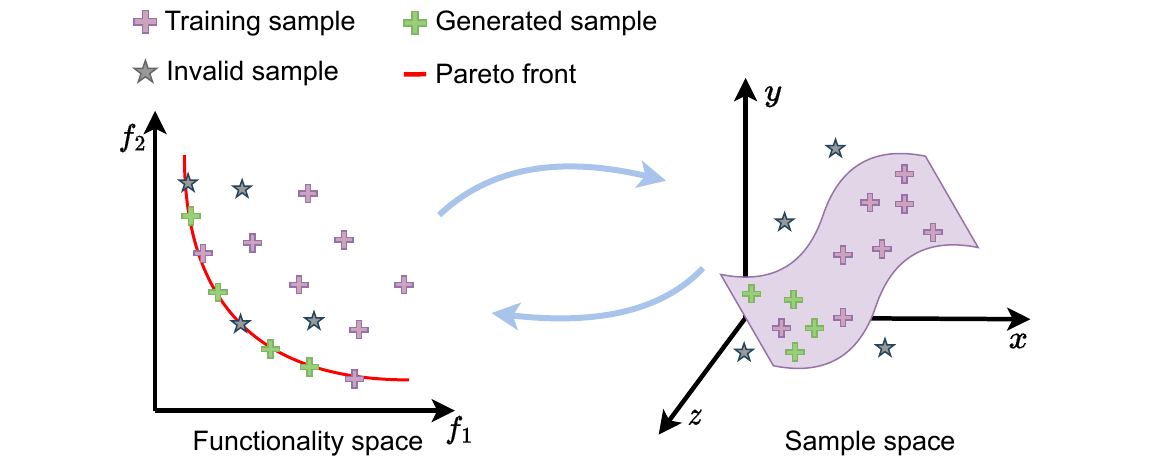}}
 	\centerline{\small(a)}
	\end{minipage}
    \begin{minipage}{0.35\linewidth}
	\centerline{\includegraphics[width=1\textwidth]{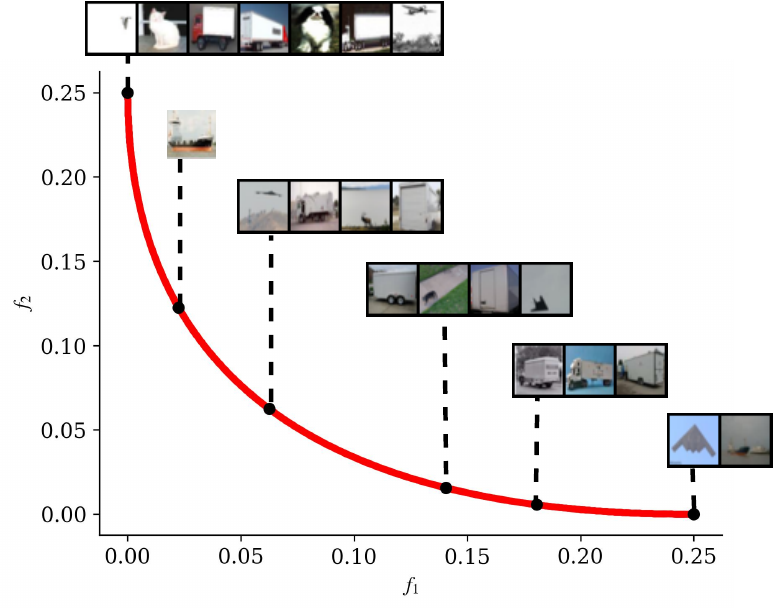}}
 	\centerline{\small(b)}
	\end{minipage}
    \caption{\label{fg:motivation} {(a) Diagram of multi-objective generation (best viewed in color). Our multi-objective generation aims to produce samples that simultaneously lie on the Pareto front in the functionality space (\textbf{Left Panel}) and remain within the manifold $\mathcal{X}$ in the sample space (\textbf{Right Panel}), i.e., the green cross. (b)~Visualization of the image generation task optimized with two objectives on CIFAR10. \revise{Images are directly taken from the original CIFAR10 dataset (see full resolution images in Fig.~\ref{fg:cifar_2obj_pf_img})}, whose objective values lie on the Pareto front, namely, $\{x|x\in X, F(x) =[f_1^{\ast}, f_2^{\ast}] \in F^\ast\}$, where $F^{\ast}$ denotes the points on the Pareto front.}}	\vskip-0.05in
    
\end{figure}

{Before delving into details, we first establish the problem setting.} Given a dataset~$X \subseteq \mathcal{X}$, where $\mathcal{X} \subset \mathbb{R}^d$ denotes a low-dimensional manifold in the high-dimensional space~$\mathbb{R}^d$. Suppose we have $m$ objective functions~$F(x)=[f_1(x), f_2(x), \ldots, f_m(x)]$, each of which returns a property value for the sample~$x\in\mathcal{X}$. The aim of multi-objective generation is to learn a generative model that produces samples optimized to achieve the best values across these functions while ensuring the generated samples remain within the manifold~$\mathcal{X}$ (green cross in Fig.~\ref{fg:motivation}(a), {namely,  ensuring that the quality of generated samples (dubbed as \textit{generation quality}) is good\footnote{{In other words, the generated samples is as realistic as samples in the given dataset~$X$.}}.}

The multi-objective generation problem introduced above inherently requires reconciling the optimization challenges in two spaces: the functionality space and the sample space as shown in Fig.~\ref{fg:motivation}(a). 
{Given the need to deal with multiple conflicting objectives in order to achieve the generation with desired properties}, one challenge is how to produce samples that cannot be further improved simultaneously across the objectives, a.k.a. \emph{Pareto optimality}~\citep{chinchuluun2007survey} (the Pareto front in Fig.~\ref{fg:motivation}(a)). 
The second challenge arises from the manifold assumption that the generated samples should lie within the data manifold~$\mathcal{X}$, namely, generated samples are supposed to be of good quality~\citep{sanchez2018inverse}. Optimizing multiple objectives without considering generation quality could result in Pareto solutions outside of the data manifold (i.e., invalid samples on the Pareto front of Fig.~\ref{fg:motivation}(a)). The third challenge relates to the coordination of generation quality and multi-property optimization. {To guarantee generation quality, generative models typically define a divergence between the distribution of generated data and that of real training data~$X$~\citep{yang2022diffusion,goodfellow2014generative}}, which tends to disperse the generated data throughout the whole data manifold~$\mathcal{X}$ (the purple plane in Fig.~\ref{fg:motivation}(a)). However, since only a limited fraction of the samples on the data manifold lie on the Pareto front, there inevitably exists some distribution gap between the generated data and the training data, leading to \emph{compromise of generation quality}, when achieving Pareto optimality.

A large number of studies~\citep{klys2018learning,deng2020disentangled,wang2022controllable,li2022editvae} attempt to design controllable generative models with multiple properties by simply assuming that these properties are independent and aggregating the multiple property objectives into a single one~$\sum_{i=1}^{m}f_i$ for controlled generation.
Notably, a very recent study~\citep{gruver2023protein} takes into consideration the trade-offs between multiple properties by incorporating the multi-objective optimization techniques into the generative models. It modified the gradient of sampling in vanilla diffusion models as a linear combination of the original diffusion gradient and the gradient solved by the multi-objective Bayesian optimization.
However, the adopted fixed coefficient is challenging to effectively coordinate the generation quality and the optimization of multiple property objectives. This results in an unnecessary loss of generation quality while achieving Pareto optimality for the property objectives.


In this work, we propose PaRetO-gUided Diffusion model (PROUD) for multi-objective generation. PROUD is formulated as a constrained optimization that minimizes the Kullback–Leibler (KL) divergence between the distribution of the generated data and that of the training data, where the distribution of the generated data is also constrained to be close to the distribution of Pareto solutions under the KL divergence. {This guarantees that generated samples are moved towards the Pareto set and then the quality of these generated samples is optimized to the best within a neighborhood of the Pareto set.} Specifically, constrained optimization is implemented during the generative process of a pre-trained unconditional diffusion model. Multiple gradient descents for the multiple objectives and the original diffusion gradient are adaptively weighted to denoise samples. The contributions of this work are summarized as follows:
\begin{itemize}
    \item \revision{We propose a novel constrained optimization formulation for controllable generation adhering to multiple properties, defined as multi-objective generation, which can better coordinate the generation quality and the optimization for multi-objectives.}
    \item \revision{A new controllable diffusion model (PROUD) is introduced to {solve the constrained optimization formulation}. The guidance of multiple objectives is adaptively integrated with that of data likelihood, which can reduce the needless comprise of generation quality while achieving Pareto optimality in terms of multiple property objectives.}
    \item We apply our PROUD to optimizing multiple objectives in the tasks of controllable image generation and protein design. {Additionally, we establish various baselines based on diffusion models to demonstrate the superiority of our PROUD.}
\end{itemize}

\section{Related Work}
In the section, we summarize the related works based on their strategies for integrating the optimization of multiple property objectives into deep generative models.

\textbf{Single-objective generation (SOG)} refers to approaches that simply combine multiple objectives into a single one to guide the generation. Extensive efforts have been devoted to controllable generation with multiple properties independent of each other~\citep{klys2018learning,guo2020property,jin2020multi,deng2020disentangled,wang2022controllable,li2022editvae}. Nevertheless, these methods fail to capture the correlation between properties and ignore the conflicting nature among properties, leading to an insufficient exploration of the solution space.

\textbf{Multi-objective Generation (MOG)} refers to approaches that introduce multi-objective optimization techniques into generative models. 
\citet{wang2022multi} adopted a weighted-sum strategy to deal with the trade-offs between properties, which can only work in cases of convex Pareto fronts and a uniformly distributed grid of weighting cannot guarantee uniform points on the Pareto front~\citep{sener2018multi,liu2021profiling}. 
\citet{stanton2022accelerating} proposed LaMBO (Latent Multi-objective Bayesian Optimization), which applies multi-objective Bayesian optimization in the latent space of denoising autoencoder to optimize the generated samples with multiple black-box objectives. Although it can characterize the Pareto front, the data generated by denoising autoencoder is of inferior quality. 
\citet{gruver2023protein} further applied LaMBO to the latent space of discrete diffusion models. It generalized classifier-guided diffusion models~\citep{dhariwal2021diffusion} by replacing the classifier gradient with the gradient obtained by LaMBO. The combination of the score estimate of a diffusion model and the classifier gradient necessitates manual tuning of the combination coefficient,  which is theoretically inappropriate for non-convex functions~\citep{gong2021bi}. 
\citet{tagasovska2022pareto} proposed to use multiple gradient descent~\citep{desideri2012multiple} for sampling within compositional energy-based models (EBMs) where each EBM is conditioned on one specific property, but training multiple conditional EBMs requires much more supervision than training discriminative models. Moreover, this kind of paradigm cannot enjoy post-hoc controls upon the pre-trained unconditional generative models.
Multi-objective generative flow networks (GFlowNets)~\citep{jain2023multi} fully integrated guidance from multiple objectives into the training process. So, they must be retrained whenever the objectives change and are also not suitable for use with pre-trained generative models. In addition, this kind of models are usually difficult to train~\citep{shen2023towards}.

Diffusion models~\citep{ho2020denoising, sohl2015deep, song2019generative, song2021score} represent the state-of-the-art (SOTA) in deep generative models. Therefore, we build our multiple-objective generation model based on diffusion models. While most related works design their methods based on other deep generative models, we apply their ideas to the diffusion model as much as possible for the sake of comparison. Please refer to Section~\ref{sec:experiments} for more details.

\section{Preliminaries} 
Before delving into our method, we introduce the technical background about multi-objective optimization in Section~\ref{MOO_back} and diffusion models in Section~\ref{generation_back}, respectively.

\subsection{Multi-objective Optimization}\label{MOO_back}

Let $x \in \mathbb{R}^d$ be a decision variable. Assuming that
$F(x)=\left[f_1(x), f_2(x),\ldots, f_m(x)\right]$ be a set of $m$ objective functions, each of which represents a property and is preferred to have a smaller value. The multi-objective optimization problem\revision{~\citep{chinchuluun2007survey,deb2001multi}} can be conventionally expressed as:
\begin{equation}\label{MOO_obj}
    \min_{x\in \mathbb{R}^d} F(x) = \min_{x \in \mathbb{R}^d} \left[f_1(x), f_2(x), \ldots, f_m(x)\right].
\end{equation}
In this context, for $x_1, x_2 \in \mathbb{R}^d$, $x_1$ is said to dominate $x_2$, i.e., $x_1 \prec x_2$, iff $f_i(x_1) \leq f_i(x_2), \forall i =1,2,\ldots, m$, and $F(x_1) \neq F(x_2)$.

\begin{definition}[Pareto optimality] \label{df:pareto_optimal}
 A point $x^{\ast} \in \mathbb{R}^d$ is called Pareto optimal iff there exists no any other $x' \in \mathbb{R}^d$ such that $x' \prec x^{\ast}$. The collection of Pareto optimal points are called \textit{Pareto set}, denoted as $\mathcal{P}^{\ast}$. The collection of function values $F(x^{\ast})$ of the Pareto set is called the \textit{Pareto front}\revision{~\citep{van1998evolutionary,borghi2023adaptive}}.
\end{definition}

\begin{definition}[Pareto stationarity] \label{df:pareto_stationary}
Pareto stationarity is a necessary condition for Pareto optimality. A point $x$ is called Pareto stationary if there exists a set of scalar $\omega_i, i =1,2,\ldots, m$, such that $\sum_{i=1}^m \omega_i \nabla f_i(x)= \mathbf{0}, \sum_{i=1}^m \omega_i=1, \omega_i>0, \forall i =1,2\ldots, m$.
\end{definition}

\citet{desideri2012multiple} proposed Multiple Gradient Descent (MGD) to find the Pareto optimal solutions of Eq.\eqref{MOO_obj}. To be specific, given any initial point $x \in \mathbb{R}^d$, we can iteratively update  $x$ according to:
\begin{equation}\label{eq:x_update}
    x_{t+1} = x_t-\eta v_t,
\end{equation}
where $t$ is the iteration step. The update direction $v_t$ is expected to be close to each gradient $\nabla f_i(x)$ $\forall i =1,2,\ldots, m$ as much as possible, which is therefore formulated into the following problem:
\begin{equation}\label{eq:v_t}
    \underset{v \in \mathbb{R}^d}{\max }\left\{\min _{i} \nabla f_i\left(x\right)^{\top} v-\frac{1}{2}\|v\|^2\right\}.
\end{equation}
Through Lagrange strong duality, the solution to Eq.\eqref{eq:v_t} can be framed into
\begin{equation}\label{MOO_graident}
v(x) = \nabla F(x) = \sum_{i=1}^m \omega_i^\ast \nabla f_i\left(x\right), 
\end{equation}
where  $\{\omega_i^\ast\}_{i=1}^m = \arg\min\limits_{\{\omega_i\}_{i=1}^m} \|\sum_{i=1}^m \omega_i \nabla f_i\left(x\right)\|^2$ under the constraint that $\sum_{i=1}^m \omega_i=1, \omega_i>0, \forall i =1,2\ldots, m$.

\subsection{Diffusion Models}\label{generation_back}
The idea of Diffusion models is to progressively diffuse data to noise, and then learn to reverse this process for sample generation. Considering a sequence of prescribed noise scales $0<\beta_1< \beta_2<\ldots<\beta_T<1$, Denoising Diffusion Probabilistic Model (DDPM)~\citep{ho2020denoising} diffuses data $x_0 \sim q_{\text{data}}(x)$ to noise via constructing a discrete Markov chain $\{x_0, x_1, \ldots,  x_T\}$, where $q(x_t | x_{t-1})=\mathcal{N}(x_t;  \sqrt{1-\beta_t}x_{t-1}, \beta_t \mathbf{I}), x_T \sim \mathcal{N}(\mathbf{0}, \mathbf{I})$. This process is called the forwarded process or diffusion process. In particular, $q(x_t|x_0)=\mathcal{N}(x_{t}; \sqrt{\alpha_t}x_0, (1-\alpha_t)\mathbf{I})$, where $\alpha_t=\prod_{i=1}^t\left(1-\beta_t\right)$. 

The key of diffusion-based generative models is to train a reverse Markov chain so that we can generate data starting from a Gaussian noise $p(x_T)\sim \mathcal{N}(\mathbf{0}, \mathbf{I})$. 
The training loss of the reverse diffusion process, a.k.a. generative process, is to minimize a simplified variational bound of negative log likelihood. Namely,
\begin{equation}
    \mathbb{E}_{x_0\sim q_\text{data}(x), \epsilon\sim \mathcal{N}(\mathbf{0}, \mathbf{I})} \left[\|\epsilon-\epsilon_\theta\left(\sqrt{\alpha_t} x_0+\sqrt{1-\alpha_t} \epsilon, t\right)\|^2\right],
\end{equation}
where $\epsilon_\theta(x_t, t)$ is a neural network-based approximator to predict the noise~$\epsilon$ from $x_t=\sqrt{\alpha_t} x_0+\sqrt{1-\alpha_t} \epsilon$.

After training the neural network parameterized  by $\theta$ to obtain the optimal $\epsilon_{\theta}^{\ast}(x_t, t)$, samples can be generated by starting from $x_T \sim \mathcal{N}(\mathbf{0}, \mathbf{I})$ and reversing the Markov chain:
\begin{equation}\label{eq:diffusion}
    x_{t-1}  = \frac{1}{\sqrt{1-\beta_t}}\left(x_t-\frac{\beta_t}{\sqrt{1-\alpha_t}} \epsilon_{\theta}^{\ast}\left(x_t, t\right)\right)+\sqrt{\beta_t} z_t,
\end{equation}
where $z_t \sim \mathcal{N}(\mathbf{0}, \mathbf{I})$ and $t=T, T-1, \ldots, 1$. 
More variants of diffusion models can be seen in~\citet{yang2022diffusion}. 

Existing attempts for incorporating multiple desired properties into the diffusion model~\citep{gruver2023protein} can be straightforwardly adding the derived MGD $\nabla F(x)$ in Eq.\eqref{MOO_graident} to the noise predictor $\epsilon_{\theta}^{\ast}(x_t, t)$ at each denoising step, namely,
\begin{equation}\label{eq:diffusion_mgd}
    x_{t-1}  = \frac{1}{\sqrt{1-\beta_t}}\left(x_t-\frac{\beta_t}{\sqrt{1-\alpha_t}} \Big(\epsilon_{\theta}^{\ast}\left(x_t, t\right)+\lambda \nabla F(x)\Big)\right)+\sqrt{\beta_t} z_t,
\end{equation}
where $t=T, T-1, \ldots, 1$. $\lambda$ is a trade-off hyper-parameter which balances the generation quality (i.e., the noise predictor $\epsilon_{\theta}^{\ast}(x_t, t)$) and multiple-objectives (i.e., the MGD $\nabla F(x)$). Note that an inappropriate $\lambda$ may lead to unsatisfied samples which either suffer from low quality or fail to possess required properties (Refer to experimental observations in Section~\ref{sec:experiments}).


\section{Multi-Objective Generation}



As discussed above, optimizing generative models in terms of $m$ objectives aims to produce samples that cannot be simultaneously improved for all objectives, namely, \textit{Pareto optimality} (see Definition~\ref{df:pareto_optimal}). Meanwhile, the generated samples are required to be as realistic as the training samples, which is usually achieved by enforcing distribution alignment between the generated samples and the training samples.

\paragraph{MOG compared with MOO} 
{\color{black}{
As shown in Table~\ref{tab:mooVSmog}, {both the MOO and MOG share the same objectives~$F(x)$ but differ in the space that $x$ resides in, which is termed as ``decision space'' or ``solution space'' in the MOO problem~\citep{chinchuluun2007survey} and is termed as ``data space'' in the MOG problem~\citep{gruver2023protein,wang2022controllable}}. To be specific, the decision space of the MOO problem is defined as the whole space of $\mathbb{R}^d$~\citep{cheng2017benchmark}, while the data space of the MOG problem only resides in a low-dimensional manifold~$\mathcal{X}$ embedded in $\mathbb{R}^d$ (a.k.a. the ambient space)~\citep{fefferman2016testing,roweis2000nonlinear,mcinnes2018umap}. 
Such a difference highlights that the objectives to be optimized for MOG are only meaningful within the data manifold. 
When simply applying MOO algorithms to search for solutions in the high-dimensional sample space, the obtained solutions cannot guarantee residing within the data manifold, thus resulting in very low data quality (i.e., invalid samples in Fig.~\ref{fg:motivation}(a)) and a loss of practicability~\citep{sanchez2018inverse}.

To sum up, the necessity to concurrently consider generation quality distinguishes the MOG problem from the MOO problem. Specifically, a dataset with real samples is required to define the data manifold on which the generated samples are expected to reside (Eq.\eqref{eq:constrained_obj}).
}}

\begin{table}[!t]
\centering
\caption{\revision{The MOO problem vs. the MOG problem. The generation quality in MOG is usually modeled based on the given dataset~$X\subset \mathcal{X}$, where $\mathcal{X}$ denotes a low-dimensional manifold embedded in the high dimensional space~$\mathbb{R}^d$.}}
\begin{tabular}{c|c|c|c}
\toprule[1.3pt]
 & objectives & decision/data space & generation quality \\
\midrule[1.0pt]
MOO & \multirow{2}{*}{$F(x)=[f_1(x), f_2(x), \ldots, f_m(x)]$} & $x \in \mathbb{R}^d$ & \XSolidBrush \\ 
MOG &  & $x \in \mathcal{X}, \mathcal{X} \subset \mathbb{R}^d$ & \Checkmark \\
\bottomrule[1.3pt]
\end{tabular}
\label{tab:mooVSmog}
\end{table}
\subsection{Constrained Optimization for MOG}

A straightforward solution of MOG is to take consideration of \revision{generation quality} as an additional objective and formulate it into a $m+1$ objectives problem. However, the heterogeneity of multiple objective optimization (usually defined w.r.t. a single sample) and the distribution alignment (defined w.r.t. a dataset) would bring out the optimization difficulty for the resultant MOO.
Although it is feasible to simplify the distribution divergence w.r.t. a dataset as quality scores for individual samples in some deep generative models~\citep{arjovsky2017wasserstein}, it is still challenging to obtain desired solutions that achieve Pareto optimality on $m$ objectives from the optimization of $m+1$ objectives which explore a much larger space, as empirically verified in the experiments. In addition, the complexity of multi-objective optimization increases significantly with the number of objectives~\citep{ishibuchi2008evolutionary}. 

Instead of formulating a complex and ineffective $m+1$ objective problem, we implement the multi-objective generation through a tailor-designed constrained optimization problem upon $m$ property objectives. Such a formulation also allows us to stress respective significance of data generation and $m$-objective optimization, instead of treating them equally important. 
Specifically, let $p_\theta(x)$ denote the target data distribution parameterized by $\theta$, and $p_ 0$ denote the distribution of the solution samples on the Pareto front, our constrained optimization problem can be formulated as follows
\begin{equation}\label{eq:constrained_obj}
\begin{aligned}
 \min_{\theta} D\left[q_\text{data}(x)||p_\theta(x)\right] \quad s.t. \ D\left[p_0(x)||p_\theta(x)\right] \leq \varepsilon.
\end{aligned}
\end{equation}
where $D(\cdot,\cdot)$ denotes the distribution divergence and $\varepsilon$ is a  small positive value. 

The loss in Eq.\eqref{eq:constrained_obj} controls the generation quality, which ensures the quality of the generated data as realistic as possible. The constraint in Eq.\eqref{eq:constrained_obj} ensures the generated data $x\sim p_\theta(x)$ to be Pareto optimal (with a small bearable error). Overall, Eq.\eqref{eq:constrained_obj} provides certain quality assurance while obtaining samples that can approach Pareto optimality of multiple property objectives. 

\subsection{Langevin Dynamics for Data Distribution Approximation}
It is difficult to directly solve Eq.\eqref{eq:constrained_obj} when both $q_{\text{data}}(x)$ and $p_0(x)$ are unknown.  
Motivated by those widely-developed techniques of sampling algorithms for approximating data distribution~\citep{andrieu2003introduction,song2019generative,liu2021profiling}, we develop Langevin dynamic-based sampling techniques to solve Eq.\eqref{eq:constrained_obj}. 
Specifically, Langevin dynamics are capable of generating samples from a given probability distribution~$q(x)$ solely by utilizing its score function~$\nabla \log q(x)$. Given an initial value $x_T \sim \mathcal{N}(\mathbf{0}, \mathbf{I})$, the Langevin method recursively computes the following:
\begin{equation}\label{eq:LD}
    x_{t-1} = x_t-\kappa g(x_t)+\sqrt{2\kappa} z, \quad t=T, T-1, \ldots, 0,
\end{equation} 
where $\kappa$ is the step size and can be fixed or dynamic, $z$ is sampled from the standard normal distribution $\mathcal{N}(\mathbf{0}, \mathbf{I})$ and $g(x_t)$ is the update direction for $x_t$, equal to $\nabla \log q(x_t)$. The distribution of $x_0$ will be close to the given data distribution~$q(x)$ when $\kappa \rightarrow 0$ and $T \rightarrow \infty$ under some regularity conditions~\citep{welling2011bayesian}. 

Before deriving the proper gradient~$g(x_t)$ to approximate the distribution optimized in Eq.\eqref{eq:constrained_obj} as a whole, we investigate the gradient-based strategies to optimize $D\left[q_\text{data}(x)||p_\theta(x)\right]$ and $D\left[p_0(x)||p_\theta(x)\right]$ via Langevin dynamics, separately.

\textit{Optimization of $D\left[q_\text{data}(x)||p_\theta(x)\right]$ in Eq.\eqref{eq:constrained_obj}.} Actually, various generative models are deduced to approximate the minimization of the KL divergence between the data distribution $q_\text{data}(x)$ and the model distribution~$p_\theta(x)$~\citep{DBLP:journals/corr/KingmaW13,song2021maximum,papamakarios2021normalizing}. Here, we choose diffusion models as the representative for optimizing $D\left[q_\text{data}(x)||p_\theta(x)\right]$ given their equivalent form to Eq.\eqref{eq:LD}~\citep{ho2020denoising,song2021score}. Particularly, the time-dependent predicted noise~$\epsilon_{\theta}^{\ast}\left(x_t, t\right)$ in Eq.\eqref{eq:diffusion} is the update direction $g(x_t)$ in anneal Langevin dynamics with a dynamic step size~$\eta_t$:
\begin{equation}\label{eq:guided_diffusion}
    x_{t-1} = x_t-\eta_t \epsilon_{\theta}^{\ast}\left(x_t, t\right)+\sqrt{2\eta_t} z.
\end{equation}
Consequently, the distribution of $p_\theta(x_0)$ will approach $q_\text{data}(x)$~\citep{song2021maximum}.

\textit{Optimization of $D\left[p_0(x)||p_\theta(x)\right]$ in Eq.\eqref{eq:constrained_obj}.} On the other hand, we can integrate MGD (Eq.\eqref{MOO_graident}) into Langevin dynamics to optimize $D\left[p_0(x)||p_\theta(x)\right]$, aiming to approximate the distribution of the Pareto set $p_0(x)$ upon convergence. Namely,
\begin{equation}\label{eq:MOO-LD}
    x_{t-1} = x_t-\eta  \nabla F(x_t)+\sqrt{2\eta} z,
\end{equation}
where $\eta $ is a fixed step size. The distribution of $x_0$ will converge to $p_0(x)$, as demonstrated in Theorem 3.3 of~\citet{liu2021profiling}. 

\subsection{Pareto-guided Diffusion Model}
Based on the above analysis, the key to solving the constrained optimization problem (Eq.\eqref{eq:constrained_obj}) is to design a proper strategy for unifying the optimization of $D\left[q_\text{data}(x)||p_\theta(x)\right]$ and $D\left[p_0(x)||p_\theta(x)\right]$ within the framework of Langevin dynamic sampling.
Therefore, we can indirectly solve Eq.\eqref{eq:constrained_obj} by designing the following strategies to update the gradient $g(x_t)$ in Eq.\eqref{eq:LD}:
\begin{itemize}
    \item[1)] If the sample~$x_t$ is far away from the Pareto front (\textit{constraint violation}), $g(x_t)$ is chosen to assure Pareto improvement (i.e., decreasing all the $m$ objectives) to $x_t$. The amount of Pareto improvement is determinant by the distance of $x_t$ to the Pareto front.
    \item[2)] If there are multiple directions that can yield Pareto improvement (\textit{constraint violation}), the direction of Pareto improvement that decreases $D\left[q_\text{data}(x)||p_\theta(x)\right]$ most (\textit{reducing loss}) is chosen as $g(x_t)$.
    \item[3)] If $x_t$ is close to the Pareto front (\textit{constraint satisfaction}), i.e., having a small $\|\nabla F\left(x_t\right)\|$ according to Definition~\ref{df:pareto_stationary}, $g(x_t)$ is chosen to fully optimize $D\left[q_\text{data}(x)||p_\theta(x)\right]$ (\textit{reducing loss}).
\end{itemize}

Following~\citet{ye2022pareto}, we design a new objective based on the gradients to achieve the above conditions. To be specific, since $\epsilon_{\theta}^{\ast}\left(x_t, t\right)$ is the gradient for optimizing $D\left[q_\text{data}(x)||p_\theta(x)\right]$, and $\nabla F(x)$ is the gradient for optimizing $D\left[p_0(x)||p_\theta(x)\right]$, the integrated gradient $g(x_t)$ can be solved by the following objective:
\begin{equation}\label{eq:constraint_v}
\begin{aligned}
    & g(x_t) = \arg\min_{g} \frac{1}{2} \|g-\epsilon_{\theta}^{\ast}\left(x_t, t\right)\|^2 \\
    & s.t. \quad \nabla f_i(x)^T g \geq \phi_t, \quad \forall i = 1,2,\ldots, m,\\
    &\qquad  \phi_t= \begin{cases} \alpha \|\nabla F\left(x_t\right)\| & \text {if } \|\nabla F\left(x_t\right)\| >  e \\ \quad -\infty  & \text{otherwise}\end{cases},
\end{aligned}
\end{equation}
where $\alpha$ and $e$ are positive hyper-parameters.
The constraint in Eq.\eqref{eq:constrained_obj} can be approximated by the small gradient norm~$\nabla F(x)$ due to Pareto stationarity (Definition~\ref{df:pareto_stationary}). 
In particular, when $\|\nabla F\left(x_t\right)\|> e$, $\phi_t$ is set to be proportionate to $\|\nabla F\left(x_t\right)\|$. This will encourage the gradient~$g(x_t)$ to have positive inner products with all $\nabla f_i(x)$, approximating $\nabla F(x)$. Meanwhile, the amount of Pareto improvement is based on the distance of $x_t$ to the Pareto front. If $\|\nabla F\left(x_t\right)\|$ has a very small norm, which means that the sample~$x_t$ is close to the Pareto front, we will have $g_t(x)=\epsilon_{\theta}^{\ast}\left(x_t, t\right)$ with $\phi_t=-\infty$. Therefore, samples will be updated with a pure gradient descent on $D\left[q_\text{data}(x)||p_\theta(x)\right]$ without taking into account the $m$ objectives $\{f_i(x)\}_{i=1}^{m}$, namely, $\lambda_{i,t}=0, \forall i \in [m]$.

At the situation of $\|\nabla F\left(x_t\right)\|> e$, the solution $g(x_t)$ of Eq.\eqref{eq:constraint_v} is expressed as:
\begin{equation}\label{eq:poud_diff}
    g(x_t) = \epsilon_{\theta}^{\ast}\left(x_t, t\right) + \sum_{i=1}^{m} \lambda_{i, t} \nabla f_i(x_t),
\end{equation}
where $\{\lambda_{i, t}\}_{i=1}^m$ is the solution of the following dual problem:
\begin{equation}\label{eq:lambda}
    \max_{\lambda_{i,t} \in \mathbb{R}_{+}^m} -\frac{1}{2}\|\epsilon_{\theta}^{\ast}\left(x_t, t\right) + \sum_{i=1}^{m} \lambda_{i, t} \nabla f_i(x_t)\|^2+\sum_{i=1}^m \lambda_{i,t} \phi_t.
\end{equation}
Substituting the derived gradient $g(x_t)$ (Eq.\eqref{eq:poud_diff}) into Eq.\eqref{eq:LD} and adopting a dynamic step size~$\eta_t$, we can obtain a new kind of controllable diffusion modeling, which is named as PaRetO-gUided Diffusion model (PROUD):
\begin{equation}\label{eq:pond}
    x_{t-1} = x_t-\eta_t \left(\epsilon_{\theta}^{\ast}\left(x_t, t\right) + \sum_{i=1}^{m} \lambda_{i, t} \nabla f_i(x_t)\right)+\sqrt{2\eta_t} z.
\end{equation}
PROUD does not modify the training process of diffusion models but only updates gradients during the generative process, as summarized in Algorithm~\ref{alg}. Therefore, our PROUD can be plugged into any pre-trained diffusion model to gain post-hoc control during the generative process.

In contrast to existing methods that crudely combine generative models with multi-objective optimization techniques using a predefined balance coefficient, our constrained optimization formulation (Eq.\eqref{eq:constrained_obj}) allows to dynamically infer the balance coefficient (Eq.\eqref{eq:lambda}), prioritizing the guarantee of Pareto optimality.
 
\begin{algorithm}[tb]
\caption{\label{alg} Pareto-guided Reverse Diffusion Process for a Single Sample}
\begin{algorithmic}[1]
  \State {\bfseries Input:} a pre-trained unconditional diffusion model~$\epsilon_\theta^{\ast}$, the dynamic step size~$\{\eta_t\}_{t=1}^{T}$, multiple property objectives~$\{f_i\}_{i=1}^{m}$.
    \State {\bfseries Hyper-parameters:} $\alpha$ and $e$ in Eq.\eqref{eq:constraint_v}.
  \State {\bfseries Initialize:} $x_T \sim \mathcal{N}(\mathbf{0}, \mathbf{I})$.
  \For{t = $T,T-1,\ldots, 0$}
      \State calculate the multiple gradient descent: $\nabla F(x_t)$ based on Eq.\eqref{MOO_graident};
      \If{$\|\nabla F(x_t)\| > e$} \# calculate the weight coefficients
      \State $\{\lambda_{i,t}\}_{i=1}^{m}$ takes the solution of Eq.\eqref{eq:lambda} with $\phi_t=\alpha \|\nabla F(x)\|$;
      \Else
      \State $\lambda_{i,t}=0, \forall i \in [m]$; 
      \EndIf
      \State \mbox{calculate the denoising gradient: 
      $g(x_t) = \epsilon_{\theta}^{\ast}\left(x_t, t\right) + \sum_{i=1}^{m} \lambda_{i, t} \nabla f_i(x_t)$ as Eq.\eqref{eq:poud_diff};}
      \State sample $z \sim \mathcal{N}(\mathbf{0}, \mathbf{I})$;
      \State denoise the sample: $x_{t-1} = x_t -\eta_t g(x_t)+\sqrt{2\eta_t}z$;
  \EndFor
  \State {\bfseries Output:} the sample $x_0$ which meets Pareto optimality of $m$ objectives.
\end{algorithmic}
\end{algorithm}

\subsection{Diversity Regularization for Diversiﬁed Pareto Solutions}
In practice, MGD integrated with Langevin dynamics fails to obtain diversified Pareto solutions although it can be guaranteed to obtain solutions on the Pareto front~\citep{liu2021profiling}. To make the solutions be evenly distributed on the Pareto front, we consider adding a diversity regularization, which can be enforced either in the sample space or the functionality space. Because we are interested in high-dimensional data generation, imposing larger distances between samples can be challenging. Furthermore, a significant separation between samples does not necessarily ensure a substantial distinction between their respective functionalities. Therefore, we define the diversity regularization based on the objective values. 

Suppose there are $N$ particles $\{x^1, x^2, \ldots, x^N\}$ in each step of our PROUD. We omit the subscript~$t$ of the time step for simplicity. The diversity loss is defined to encourage the dissimilarity of the objective values: 
\begin{align}\label{eq:diversity}
    l(x^1, x^2, \ldots, x^N) &= \sum_{i \neq j} \frac{1}{\|F(x^i)-F(x^j)\|^2}.
\end{align}
The diversity loss Eq.\eqref{eq:diversity} is added to the main objective in Eq.\eqref{eq:constrained_obj} with a weight coefficient~$\gamma$.


\section{Experiments}\label{sec:experiments}
In this section, we evaluate the effectiveness of our PROUD in optimizing image generation and protein generation with multiple conflicting objectives. We study white-box multi-objectives in this work and particularly focus on using MGD as the MOO technique to obtain the gradient from multi-objectives. The exploration of the black-box setting, as mentioned  in~\cite{stanton2022accelerating}, is discussed in the conclusion and remains for future work.

\textit{Dataset.} In the task of image generation, we use the CIFAR10~\citep{krizhevsky2009learning} dataset, which consists of 60,000 color images, each with a size of $3 \times 32 \times 32$, distributed across 10 classes. Regarding protein generation, following~\citet{gruver2023protein}, the experiment was conducted on the paired Observed Antibody Space (pOAS) dataset~\citep{olsen2022observed}, which comprises $90,990$ antibody sequences, each processed to a fixed length of 300. 

\textit{Baselines.} First, we include the most closely-related and SOTA work in MOG that applies the MOO technique to the deep generative model~\citep{gruver2023protein}. This baseline is termed as ``DM+$m$-MGD'', where the MGD of $m$ objectives is used to guide the generation of diffusion models (DM). We also include the baseline regarding single-objective generation, termed as ``DM+single''. It fuses multiple objectives into a single objective and uses the gradient of the obtained single objective to guide the generation of diffusion models. Another considered baseline is ``$m+1$-MGD''. It treats the objective of the diffusion model as an additional objective and formulates multi-objective generation as the optimization of $m+1$ objectives. MGD is then applied directly for the resultant $m+1$ objectives. To stress the necessity of quality assurance in the generation problem, which is the core difference between MOG and MOO, we include the MGD of $m$ objectives as the baseline, called ``m-MGD''. 

For all methods equipped with MGD, the diversity regularization (Eq.\eqref{eq:diversity}) is included except for $m+1$-MGD since its extra objective $f_{m+1}(x)$, i.e., data likelihood, is not accessible for the diffusion models.

\textit{Metrics.} In terms of generation quality, the Frechet Inception Distance (FID)~\citep{heusel2017gans} is adopted as the metric for image quality, while the log-likelihood assigned by ProtGPT~\citep{ferruz2022protgpt2} is considered as the metric for the quality of protein sequences following~\citet{gruver2023protein}. Concerning Pareto optimality, Hypervolume (HV)~\citep{zitzler1999multiobjective} is adopted to measure how well the methods approximate the Pareto set.

\subsection{Image Generation}
\label{sect:cifar10}
{We follow~\citet{liu2021sampling}\footnote{{As demonstrated in Section 3 and Figure 3(b) of their study, an objective that forces the center of generated images to be a black square can be used for constrained sampling on CIFAR10. Accordingly, they obtain samples that lie on the CIFAR10 data manifold and exhibit the black square in the middle, such as ``black plane'' and ``black dog'' images which contain a black square (smaller size than the object) in the middle. This task can be considered as image outpainting~\citep{yao2022outpainting}, namely, extrapolating images based on specified color patches on CIFAR10.}} to optimize CIFAR10 images with the objectives that force the middle of an image to be a specified color square.} 
\\
(1) Controllable generation on CIFAR10 with two objectives ({Fig.~\ref{fg:motivation}(b)}):
\begin{itemize}[leftmargin=1cm]
    \item[$\bullet$] $f_1(x) = \|x_\Omega-1_\Omega\|_2^2$, where $x$ represents the entire image, and $x_\Omega \subseteq x$ is an image patch in the region $\Omega$, corresponding to the square at the center of the image. \revision{Similar to the practical relevance shown in \citet{liu2021sampling}, this objective is to restrict the center of the generated images to be a white square, which is to sample CIFAR10 images that exhibit white color in their middle}. The patch size is set to $3\times 8 \times 8$ in the experiment.
    \item[$\bullet$] $f_2(x) = \|x_\Omega-0.5_\Omega\|_2^2$ with the similar setting. This objective is to constrain the center to be a grey square.
\end{itemize}
The desired generation for these two objectives would be those CIFAR10-like images with patches in normalized RGB color values\footnote{RGB values [0, 255] are divided by 255.} between [0.5, 0.5, 0.5] (grey) and [1, 1, 1] (white), in the middle, according to~\citet{ishibuchi2013many,li2017multiline}. Please refer to Appendix B for more details.\\
(2) Controllable generation on CIFAR10 with three objectives:
\begin{itemize}[leftmargin=1cm]
    \item[$\bullet$] $f_1(x) = \|x_\Omega-a_\Omega\|_2^2$, where $x$ represents the entire image, and $x_\Omega \subseteq x$ is an image patch in the region $\Omega$, corresponding to the square at the center of the image. This objective is to restrict the center of the generated images to be a black square. The patch size is set to $3\times 8 \times 8$ in the experiment. $a_\Omega=[0, 0, 0]_{8\times8}$.
    \item[$\bullet$] $f_2(x) = \|x_\Omega-b_\Omega\|_2^2$ with the similar setting. This objective is to constrain the center to be a deep red square. $b_\Omega=[0.5, 0, 0]_{8\times8}$.
    \item[$\bullet$] $f_3(x) = \|x_\Omega-c_\Omega\|_2^2$ with the similar setting. This objective is to constrain the center to be a deep yellow square. $c_\Omega=[0.5, 0.5, 0]_{8\times8}$.
\end{itemize}
The desired generation for these three objectives would be those CIFAR10-like images with patches in normalized RGB color values belonging to the convex triangle formed by the points [0, 0, 0] (black), [0.5, 0, 0] (deep red) and [0.5, 0.5, 0] (deep yellow). Please refer to Appendix B for more details. 
We adopt the diffusion model used in~\citet{song2020improved} as the backbone for CIFAR10 image generation.

\begin{figure}[!tb]
\centering
	\begin{minipage}{\linewidth}
	\centerline{\includegraphics[width=1\textwidth]{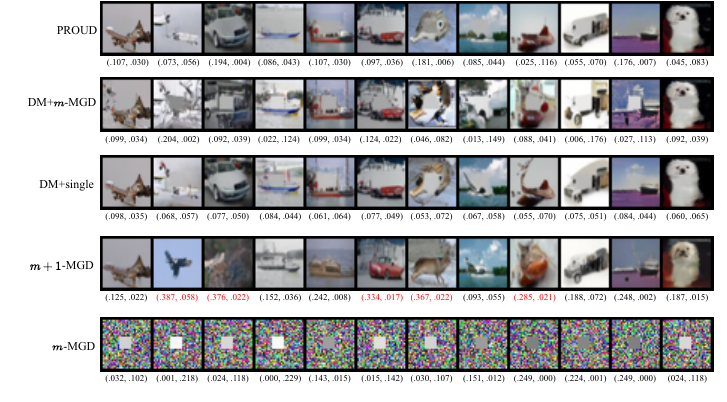}}\vskip-.1in
 	\centerline{\small(a) Two objectives}
	\end{minipage}
 	 \begin{minipage}{\linewidth}
	\centerline{\includegraphics[width=1\textwidth]{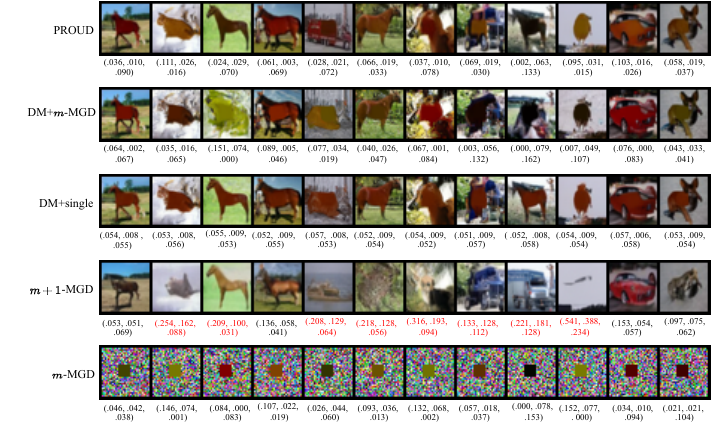}}\vskip-.1in
  	\centerline{\revision{\small(b) Three objectives}}
	\end{minipage}
\caption{\label{fg:generated_images}\revision{Generated images from our PROUD and various baselines on CIFAR10 under two/three conflicting patch-based objectives. The scores under each image refer to its objective values $[f_1(x), f_2(x)]$/$[f_1(x), f_2(x), f_3(x)]$, respectively, where those objective values do not reside on the Pareto front are marked in red.}}	\vskip-0.08in
\end{figure}

\begin{figure}[!tb]
\centering
	\begin{minipage}{0.32\linewidth}
	\centerline{\includegraphics[width=1\textwidth]{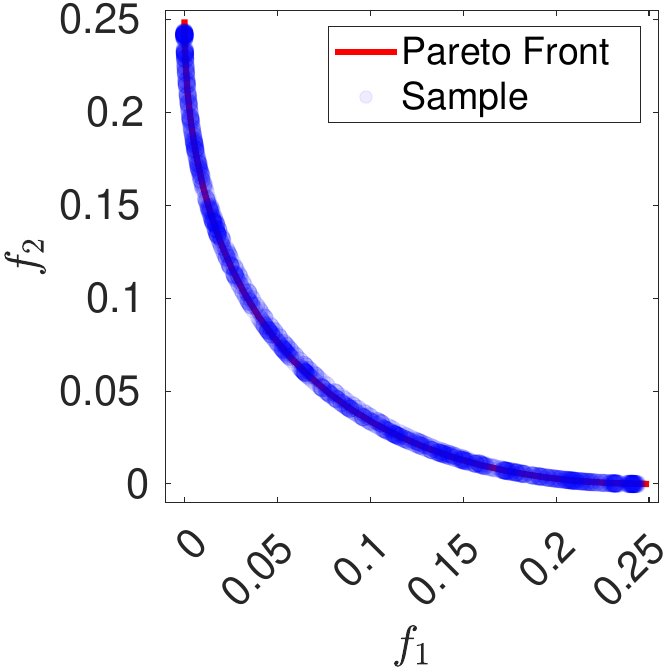}}
 	\centerline{\small(a) PROUD}
	\end{minipage}
    \hfill
 	 \begin{minipage}{0.32\linewidth}
	\centerline{\includegraphics[width=1\textwidth]{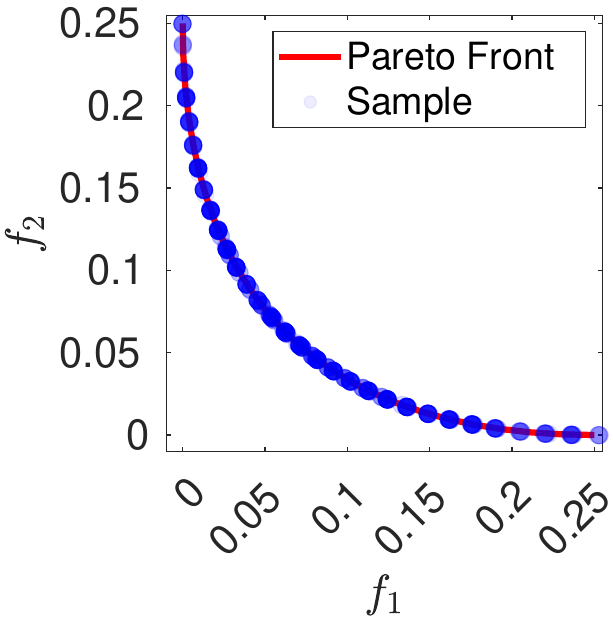}}
  	\centerline{\small(b) DM+$m$-MGD}
	\end{minipage}
    \hfill
 	\begin{minipage}{0.32\linewidth}
	\centerline{\includegraphics[width=1\textwidth]{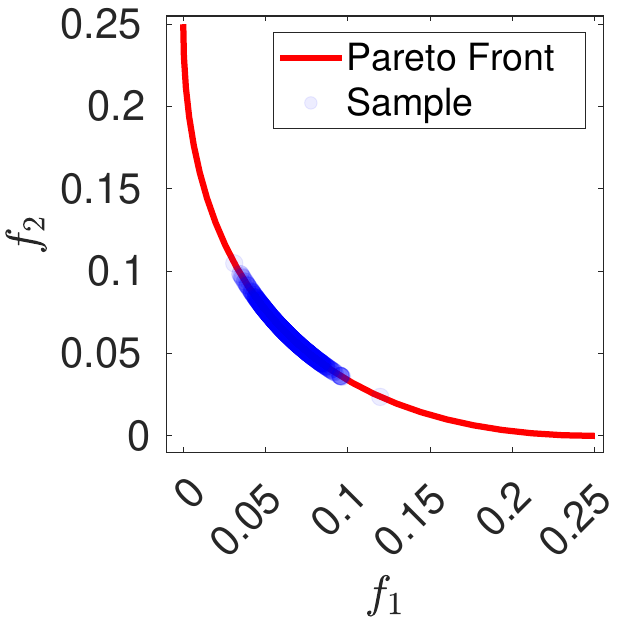}}
	\centerline{\small(c) DM+single}
	\end{minipage}
\\
    \begin{minipage}{0.32\linewidth}
	\centerline{\includegraphics[width=1\textwidth]{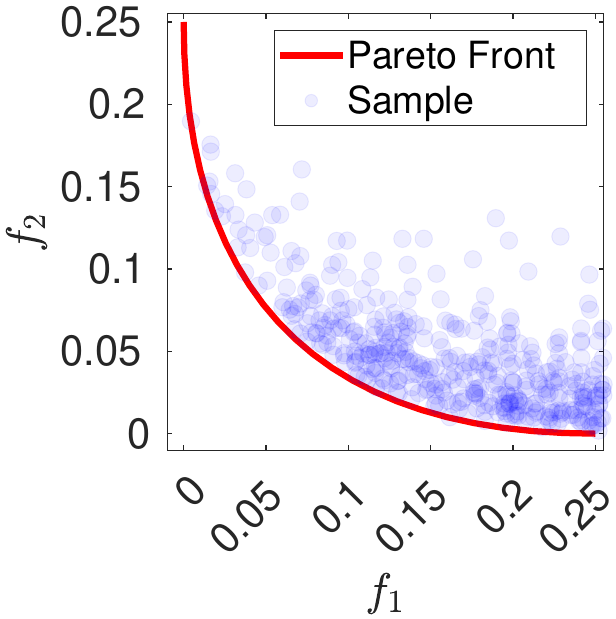}}
	\centerline{\small(d) $m+1$-MGD (Cropped)}
	\end{minipage} 
    \begin{minipage}{0.32\linewidth}
	\centerline{\includegraphics[width=1\textwidth]{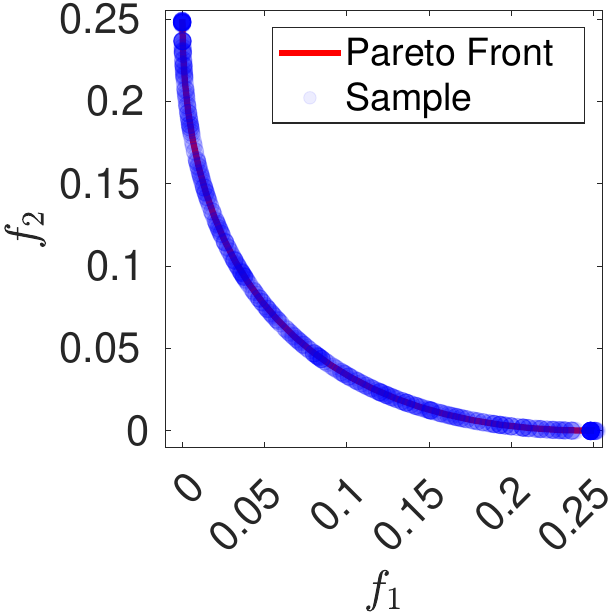}}
	\centerline{\small(e) $m$-MGD}
	\end{minipage}\hspace{2mm}
     \begin{minipage}{0.32\linewidth}
	\centerline{\includegraphics[width=1\textwidth]{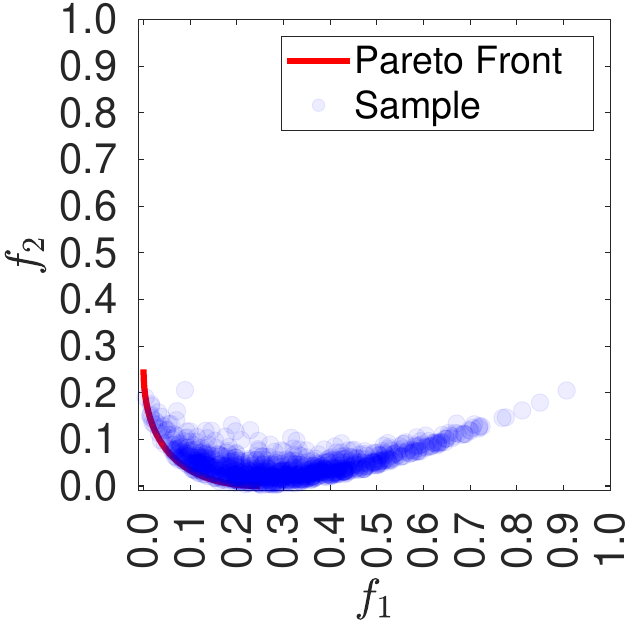}}
	\centerline{\small(f) $m+1$-MGD (full)}
	\end{minipage} 
\caption{\label{fg:pareto_front} Approximation of Pareto front of various methods on CIFAR10 optimized with two objectives. Each point denotes a generated sample, 1,000 in total, where the coordinate corresponds to its objective values. The depth of color represents sample density, the deeper the higher.}	\vskip-0.08in
\end{figure}

\begin{figure}[!tb]
\centering
	\begin{minipage}{0.32\linewidth}
	\centerline{\includegraphics[width=1\textwidth]{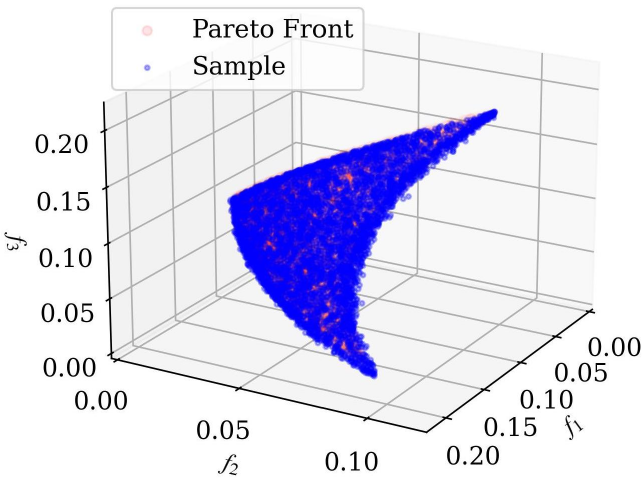}}
 	\centerline{ \begin{tabular}[c]{@{}c@{}} \small(a) PROUD \\ ($9.83 \times 10^{-5}$)\end{tabular}}
	\end{minipage}
    \hfill
 	 \begin{minipage}{0.32\linewidth}
	\centerline{\includegraphics[width=1\textwidth]{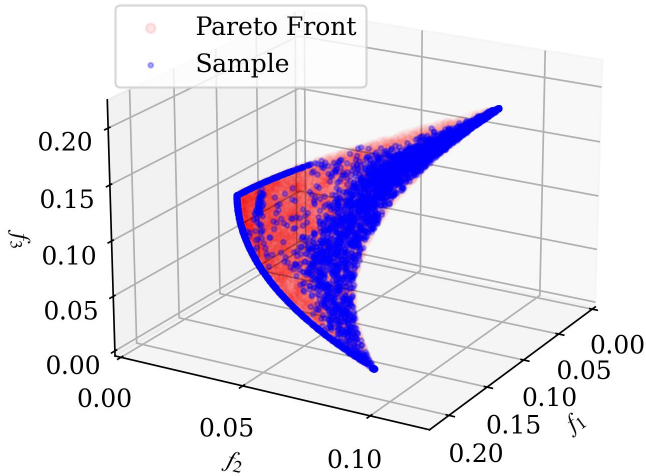}}
  	\centerline{\begin{tabular}[c]{@{}c@{}} \small(b) DM+$m$-MGD \\ ($3.10\times 10^{-4}$)\end{tabular}}
	\end{minipage}
    \hfill
 	\begin{minipage}{0.32\linewidth}
	\centerline{\includegraphics[width=1\textwidth]{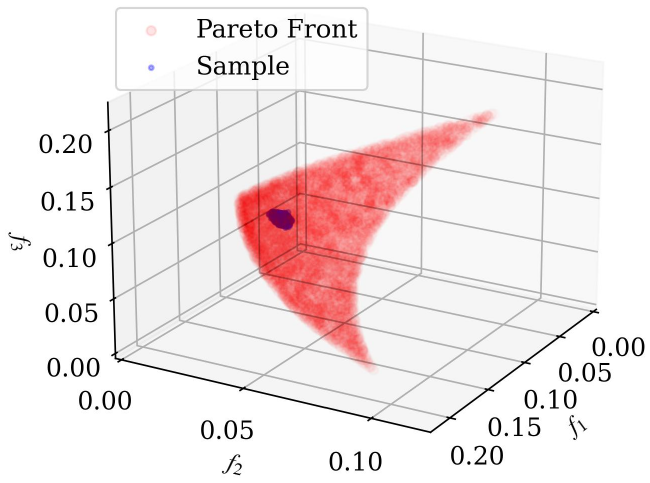}}
	\centerline{ \begin{tabular}[c]{@{}c@{}} \small(c) DM+single \\ ($6.89 \times 10^{-4}$)\end{tabular}}
	\end{minipage}
\\
    \begin{minipage}{0.32\linewidth}
	\centerline{\includegraphics[width=1\textwidth]{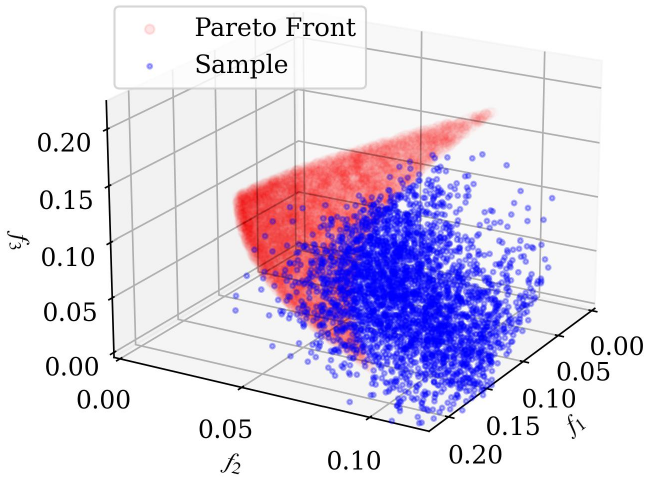}}
	\centerline{\begin{tabular}[c]{@{}c@{}} \small(d) $m+1$-MGD \\(Cropped, $2.28 \times 10^{-3}$)\end{tabular}}
	\end{minipage} 
    \begin{minipage}{0.32\linewidth}
	\centerline{\includegraphics[width=1\textwidth]{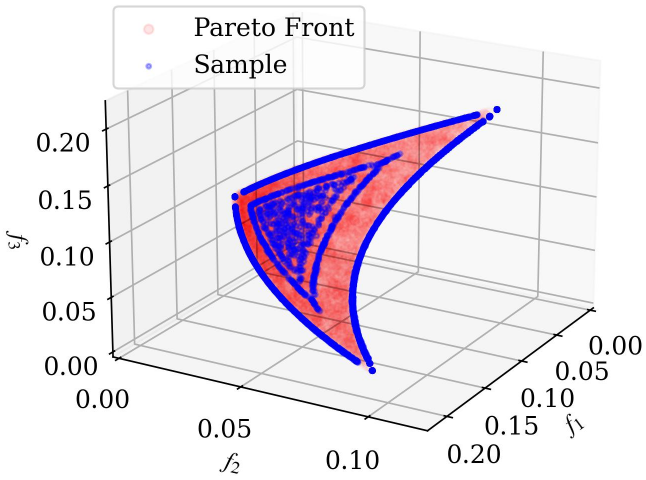}}
	\centerline{\begin{tabular}[c]{@{}c@{}}  \small(e) $m$-MGD \\($3.27\times 10^{-4}$)\end{tabular}}
	\end{minipage}\hspace{2mm}
     \begin{minipage}{0.32\linewidth}
	\centerline{\includegraphics[width=1\textwidth]{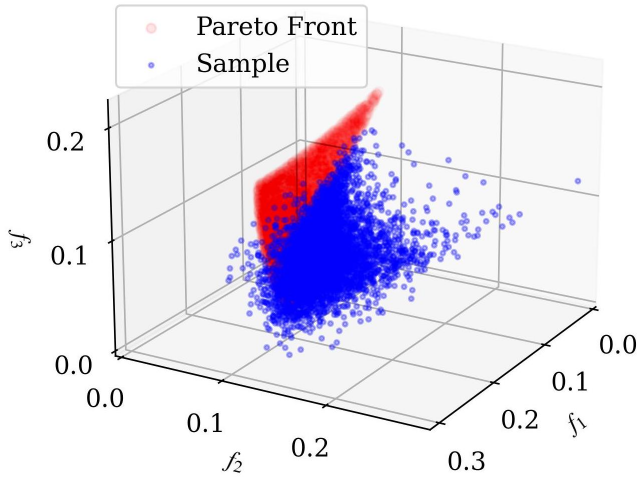}}
	\centerline{\begin{tabular}[c]{@{}c@{}}  \small(f) $m+1$-MGD \\(full, $2.50\times 10^{-3}$)\end{tabular}}
	\end{minipage} 
\caption{\label{fg:pareto_front_3obj} Approximation of Pareto front of various methods on CIFAR10 optimized with three objectives. Each point denotes each generated sample, 5,000 in total, where the coordinate corresponds to its objective values. The depth of color represents sample density, the deeper the higher. The values in the brackets are earth mover distances between the generated samples and the ground-truth Pareto solutions. We add this measure to indicate that our generated samples are indeed close to the Pareto front given the 3D visualization.}	\vskip-0.08in
\end{figure}

We sample images from our PROUD and other baselines using the same seeds for the sake of comparison. From Fig.~\ref{fg:generated_images}, we can observe that: (1) our PROUD and two baselines, DM+$m$-MGD and $m$-MGD, can successfully generate harmonious images consistent with the patch-level constraints imposed by two conflicting objectives. Among them, the generated images of our PROUD exhibit better quality than DM+$m$-MGD in some instances, as the latter tends to sacrifice generation quality to excessively meet Pareto optimality of the objectives due to the lack of a mechanism to emphasize the quality of generated samples. (2) $m+1$-MGD fails to generate satisfactory images consistent with the patch-level constraints, as the new objective (i.e, generation quality) biases the optimization of the original two objectives. Although the Pareto set of the original $m$-objectives resides within that of the $m+1$-objectives~\citep{tanabe2020easy}, the proportion is negligible even when sampling a large number of images. Refer to Fig.~\ref{fg:pareto_front}(d)\&(f) and Fig.~\ref{fg:pareto_front_3obj}(d)\&(f) for more details. (3) $m$-MGD, which does not consider generative quality in its optimization, generates meaningless images because the optimization of multiple objectives in the data generation task is only meaningful within the data manifold, as image data usually concentrate on low-dimensional manifolds embedded in a high-dimensional space.

For the MOG setting on CIFAR10 optimized with two objectives, we randomly select 1,000 generated images for each method, and calculate their objective values $[f_1(x), f_2(x)]$, respectively. Fig.~\ref{fg:pareto_front} shows that: (1) our PROUD (Fig.~\ref{fg:pareto_front}(a)) and two baselines DM+$m$-MGD (Fig.~\ref{fg:pareto_front}(b)) and $m$-MGD (Fig.~\ref{fg:pareto_front}(e)) successfully generate samples which can cover the entire Pareto front. Among them, our PROUD and $m$-MGD spread more evenly over the Pareto front. (2) DM+single only covers a partial Pareto front as shown in Fig.~\ref{fg:pareto_front}(c), because simply averaging multiple objectives into a single objective fails to explore the trade-off between multiple objectives and leads to insufficient solutions. (3) As discussed in Fig.~\ref{fg:generated_images}, $m+1$-MGD explores a much larger solution space (Fig.~\ref{fg:pareto_front}(f)), while only a few of them are located at the Pareto front of the original $m$ objectives (Fig.~\ref{fg:pareto_front}~(d)). 

For the MOG setting on CIFAR10 optimized with three objectives, we randomly select 5,000 generated images for each method and calculate their objective values [$f_1(x), f_2(x), f_3(x)$], respectively. Fig.~\ref{fg:pareto_front_3obj} shows that our PROUD exhibits significant superiority in evenly covering the Pareto front under this more challenging setting. This is because our constrained optimization formulation can better coordinate the generation quality and the optimization for multi-objectives, while ensuring sample diversity (Eq.\eqref{eq:constrained_obj}, Eq.\eqref{eq:diversity}). Although it is possible to force the two baselines DM+$m$-MGD and $m$-MGD to exhibit better diversity by setting a large diversity coefficient~$\gamma$, but this would cause the samples they generate to violate Pareto optimality, as shown in Fig.~\ref{fg:APP_dm_mgd_3obj} and Fig.~\ref{fg:APP_mgd_only_3obj} in the Appendix.

\begin{table}[!hb]
\centering
\caption{\label{tb:results} {Quantitative evaluation for Pareto approximation and generation quality. \revise{Bolded values and underlined values indicate the best results and the second best results, respectively. The Friedman \& Nemenyi test in Appendix~\ref{app_b} demonstrates that our PROUD is significantly better than other baselines.} ``-'' denotes that the value is not available as no valid data are generated.}} \vskip -0.15in
\renewcommand{\arraystretch}{1.1}
\setlength{\tabcolsep}{0.6mm}{	
\begin{tabular}{cccccccc}
\toprule[1.3pt]
\multirow{2}{*}{Method} & \multicolumn{2}{c}{CIFAR10 (2-obj)} &  \multicolumn{2}{c}{CIFAR10 (3-obj)} &  \multicolumn{2}{c}{pOAS} \\
 & HV$\uparrow$ ($10^{-2}$) & FID$\downarrow$ &  HV$\uparrow$ ($10^{-3}$) & FID$\downarrow$ &  HV$\uparrow$ & ProtGPT$\uparrow$ \\ 
\midrule
PROUD (ours) & \textbf{5.21$\pm$0.00} & \underline{31.39$\pm$0.05} &  \textbf{3.26$\pm$0.00} & \underline{44.22$\pm$0.13} &  \textbf{2472.55$\pm$60.15} & \textbf{-645.93$\pm$0.99} \\
DM+$m$-MGD & \textbf{5.20$\pm$0.01} & 38.72$\pm$0.36 &  \textbf{3.26$\pm$0.01} & 49.90$\pm$0.14 &  \underline{2289.61$\pm$65.12} & -692.80$\pm$0.34 \\
DM+single & 4.77$\pm$0.01 & 36.35$\pm$0.47 &  2.21$\pm$0.00 & 57.77$\pm$0.05 &  \underline{2302.21$\pm$58.25} & -682.26$\pm$0.49 \\
$m+1$-MGD & \underline{5.17$\pm$0.00} & \textbf{11.21$\pm$0.10} &  \underline{2.87$\pm$0.03} & \textbf{11.80$\pm$0.05} &  838.74$\pm$14.08 & \underline{-662.86$\pm$0.76} \\
$m$-MGD & \textbf{5.21$\pm$0.00} & - &  \textbf{3.26$\pm$0.01} & - &  - & - \\ 
\bottomrule[1.3pt]
\end{tabular}}\vskip-0.1in
\end{table}

To further demonstrate the superiority of our PROUD on multi-objective generation, we collect the quantitative evaluation for Pareto approximation and image quality in the left part of Table~\ref{tb:results} by sampling $50,000$ images. It shows that: our PROUD achieves the best or the second best values in both two metrics, i.e., HV for Pareto approximation and FID for image quality. It demonstrates our claim that our PROUD can provide certain quality assurance for generated samples approaching the Pareto set of multiple properties. On the contrary, either single or multiple objective generation baselines, i.e., DM+single and DM+$m$-MGD, would inevitably sacrifice \revision{generation quality} to excessively optimize the objectives.  

\subsection{Protein Sequence Generation}
To further verify our model in more challenging applications, we design multiple-objective generation task on the pOAS dataset which aims to optimize two conflicting objectives for antibody sequences:
\begin{itemize}
    \item $f_1(x)$, the solvent accessible surface area (SASA) of the protein’s predicted structure. Please refer to~\citet{ruffolo2023fast} for detailed procedures of calculating the SASA value using the protein sequences. 
    \item $f_2(x)$, the percentage of beta sheets (\%Sheets), which is measured on protein sequences directly~\citep{cock2009biopython}.
\end{itemize}
The ground-truth Pareto front is not available due to the complexity of property objectives. Since the evaluation functions for SASA and \%Sheet are not differentiable, we adopt the network predictors as differential surrogate functions for all methods. We apply the ground-truth evaluation functions for calculating the HV values on the generated samples. We adopt the discrete diffusion model in~\citet{gruver2023protein} as the backbone for protein sequence generation.

\begin{figure}[!tb]
\centering
\centerline{\includegraphics[width=0.82\textwidth]{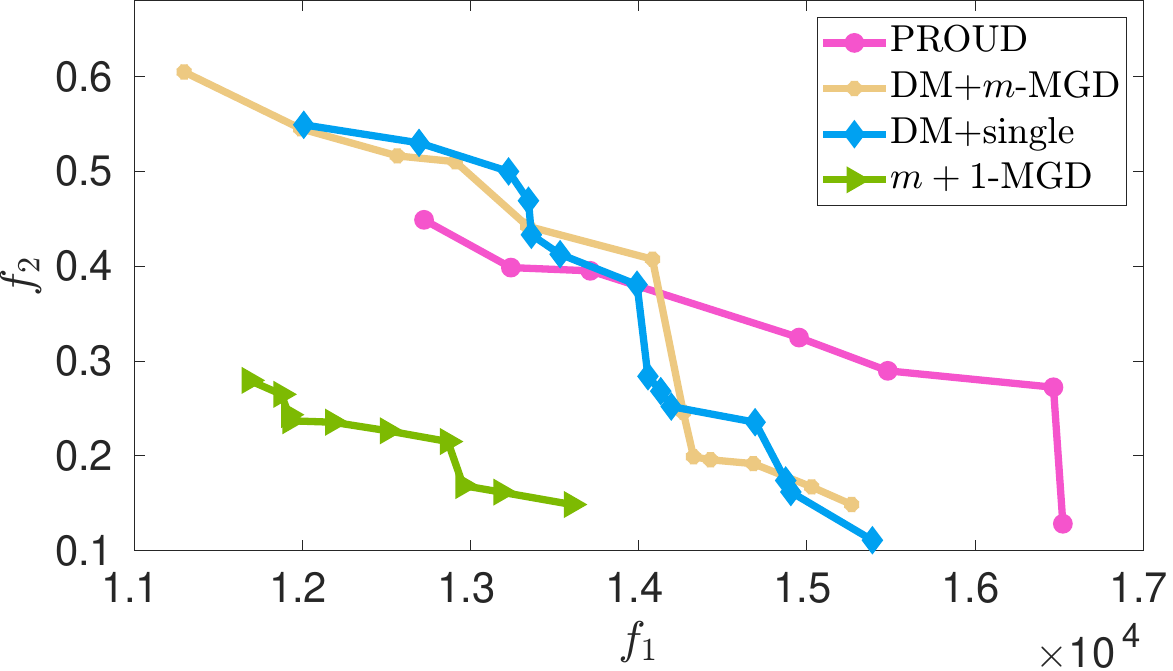}}
\caption{\label{fg:pareto_front_protein} The approximation of Pareto front (i.e., generated protein sequences) of various methods. We cannot visualize the results of $m$-MGD because all its generated protein sequences are invalid, resulting in nonexistent SASA evaluations ($f_1$).}\vskip-0.08in	
\end{figure}

To demonstrate the superiority of our PROUD in multi-objective protein generation, we initially sample $5,000$ protein sequences for each method and collect the non-dominated samples based on their two target properties, as depicted in Fig.\ref{fg:pareto_front_protein}.  The observations are as follows: (1) DM+single exhibits a wide coverage of the objective values. This could be attributed to the fact that the noise in discrete diffusion models can bring out large diversity~\citep{gruver2023protein}. By incorporating MGD into diffusion models, PROUD and DM+$m$-MGD achieve larger coverage of the objective values.  This verifies the superiority of MOG over SOG. Our PROUD and DM+$m$-MGD emphasize respective Pareto improvement  of the objectives. Nevertheless, Table~\ref{tb:results} shows that our PROUD achieves a better HV. (2) Similar to the image generation task, $m+1$-MGD demonstrates a much poorer approximation of the Pareto front for the original $m$ objectives.  Meanwhile, $m$-MGD even fails to generate any valid protein sequences, as the SASA evaluation ($f_1(x)$) for all its generated samples is nonexistent. This further highlights the difference between MOG and MOO.

Furthermore, we collect the quantitative evaluation for Pareto approximation and protein quality in the right part of Table~\ref{tb:results} by sampling $5,000$ protein sequence\footnote{We only sample $5,000$ protein sequence since the computation cost of SASA values is very high.}. Benefiting from our constrained-optimization formulation, our PROUD can avoid unnecessary loss of protein quality compared to other MOG/SOG counterparts, DM+$m$-MGD and DM+single. This improvement will greatly increase the practicality of its generated samples.

\begin{table}[!t]
\centering
\caption{\label{tb:results_alpha} Sensitivity analysis on $\alpha$ and  $e$ in Eq.\eqref{eq:constraint_v}. We retain more decimal places here to demonstrate the subtle differences between results.}
\begin{tabular}{c|ccc|ccc}
\toprule[1.3pt]
\multirow{2}{*}{Metric} & \multicolumn{3}{c|}{$\gamma=0.2,\; e=0.03$} & \multicolumn{3}{c}{$\gamma=0.2,\; \alpha=0.5$} \\
 & $\alpha=0.1$ & $\alpha=0.5$ & $\alpha=1$ & $e=0.01$    & $e=0.03$    & $e=0.05$ \\
\midrule[1pt]
FID & 31.58963073   & 31.48232218   & 31.5896311 & 31.58966697 & 31.48232218 & 31.58966696\\
HV  & 0.05211343    & 0.05211350    & 0.05211343  & 0.05211343  & 0.05211350  & 0.05211343\\
\bottomrule[1.3pt]
\end{tabular}
\end{table}

\begin{table}[!t]
\centering
\caption{\label{tb:results_gamma} Sensitivity analysis on $\gamma$. $\alpha$ and $e$ are set to $0.5$ and $0.03$, respectively. The best results are marked in bold.}
\begin{tabular}{ccccccc}
\toprule[1.3pt]
Metric & $\gamma=0$ & $\gamma=0.01$ & $\gamma=0.1$ & $\gamma=0.2$ & $\gamma=0.3$ & $\gamma=1$ \\
\midrule[1pt]
FID & 34.80      & 30.98         & 31.80        & \textbf{31.48}        & 31.63        & 33.59      \\
HV  & 0.0483     & 0.0498        & 0.0521       & \textbf{0.0521}       & 0.0521       & 0.0521     \\
\bottomrule[1.3pt]
\end{tabular}
\end{table}


\subsection{Hyper-parameter Sensitivity Study}

\begin{figure}[!htb]
\centering
	\begin{minipage}{0.243\linewidth}
	\centerline{\includegraphics[width=1\textwidth]{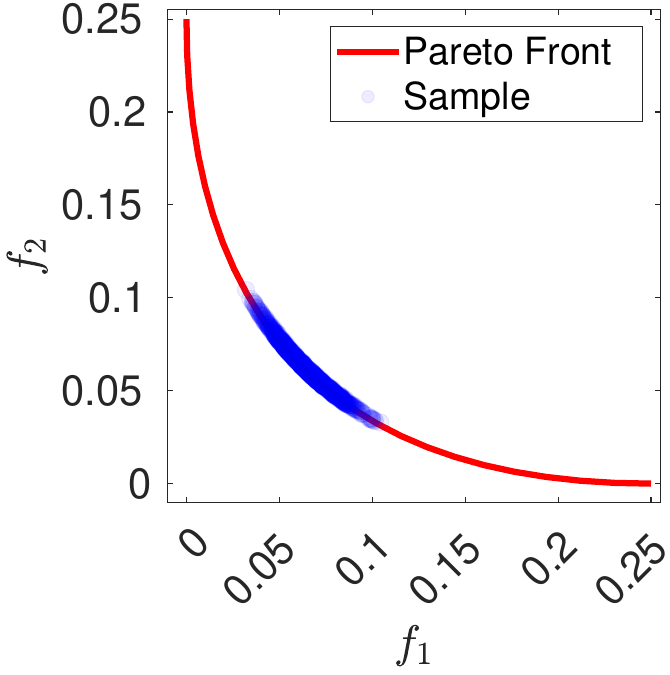}}
 	\centerline{\small(a) $\gamma=0$}
	\end{minipage}
    \hfill
 	\begin{minipage}{0.243\linewidth}
	\centerline{\includegraphics[width=1\textwidth]{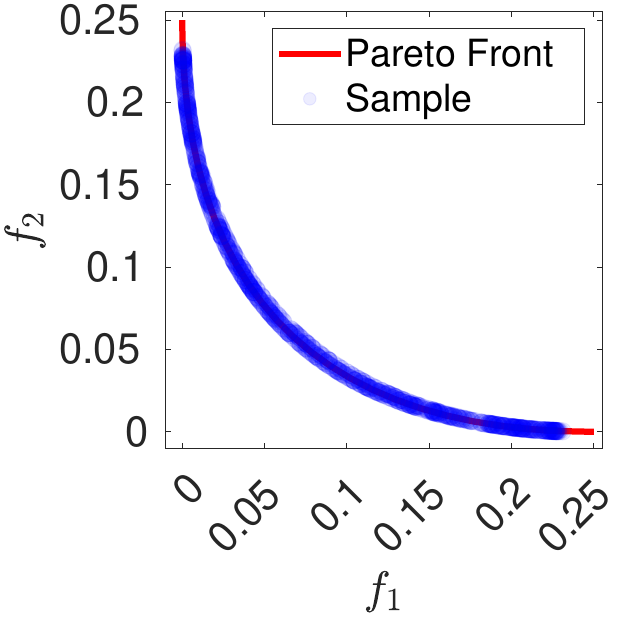}}
	\centerline{\small(b) $\gamma=0.1$}
	\end{minipage}
     \hfill
    \begin{minipage}{0.243\linewidth}
	\centerline{\includegraphics[width=1\textwidth]{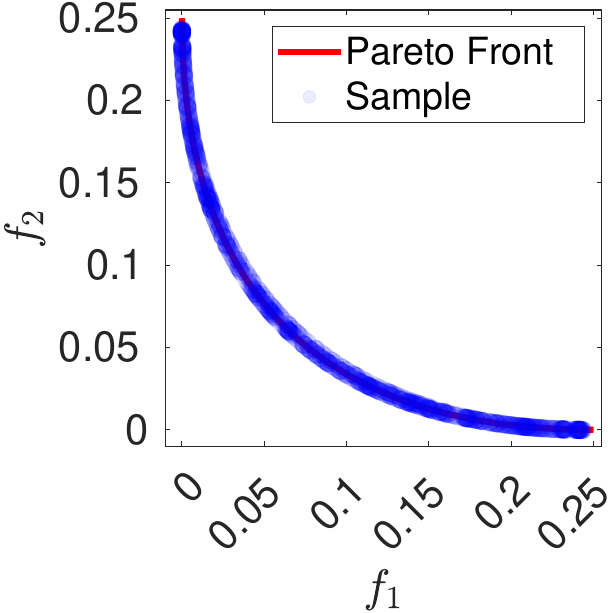}}
	\centerline{\small(c) $\gamma=0.2$}
	\end{minipage}
     \hfill
    \begin{minipage}{0.243\linewidth}
	\centerline{\includegraphics[width=1\textwidth]{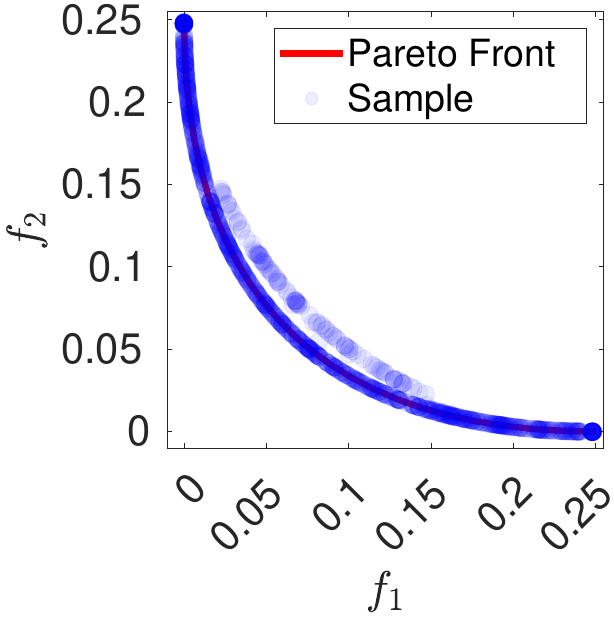}}
	\centerline{\small(d) $\gamma=1$}
	\end{minipage}
 	\begin{minipage}{0.243\linewidth}
	\centerline{\includegraphics[width=1\textwidth]{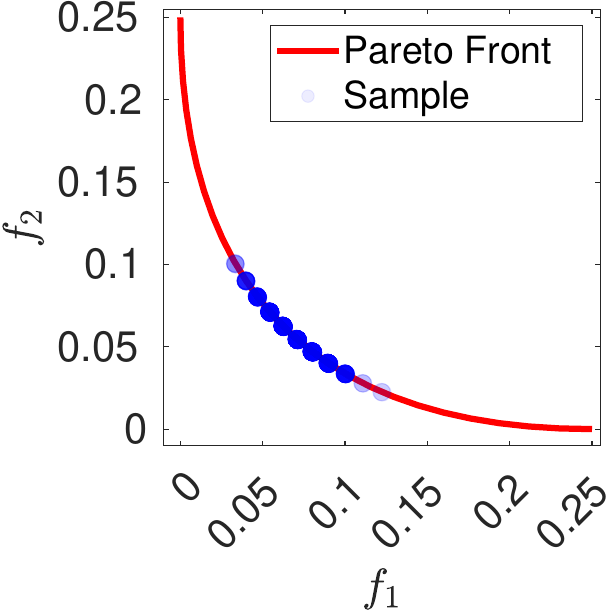}}
 	\centerline{\small(e) $\gamma=0$}
	\end{minipage}
    \hfill
 	\begin{minipage}{0.243\linewidth}
	\centerline{\includegraphics[width=1\textwidth]{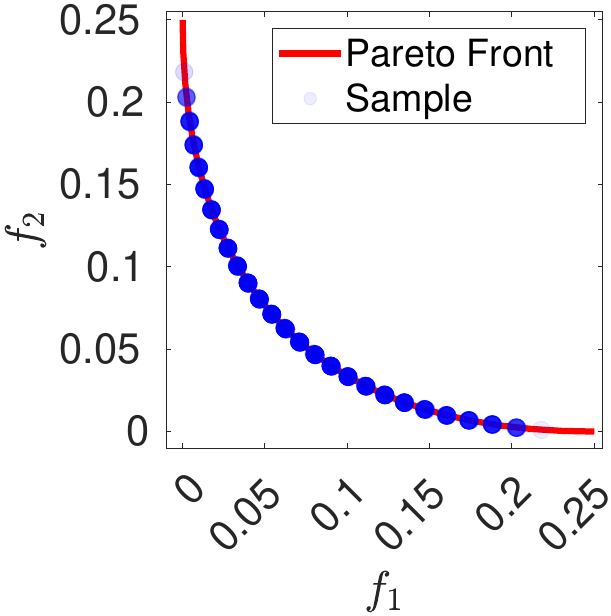}}
	\centerline{\small(f) $\gamma=0.1$}
	\end{minipage}
     \hfill
    \begin{minipage}{0.243\linewidth}
	\centerline{\includegraphics[width=1\textwidth]{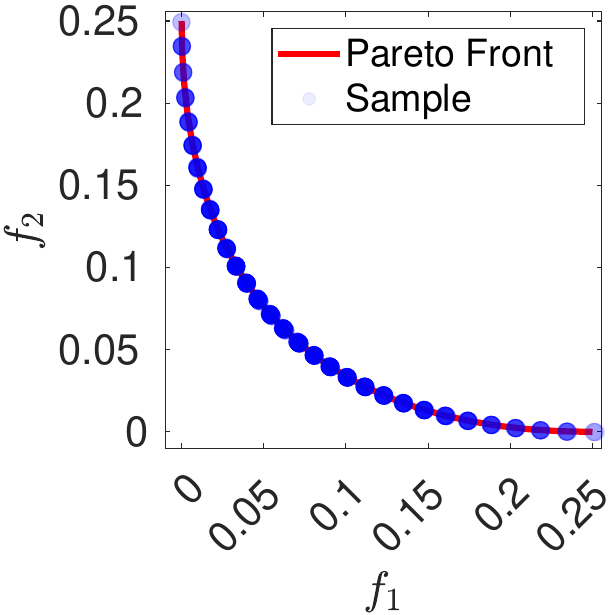}}
	\centerline{\small(g) $\gamma=0.2$}
	\end{minipage}
     \hfill
    \begin{minipage}{0.243\linewidth}
	\centerline{\includegraphics[width=1\textwidth]{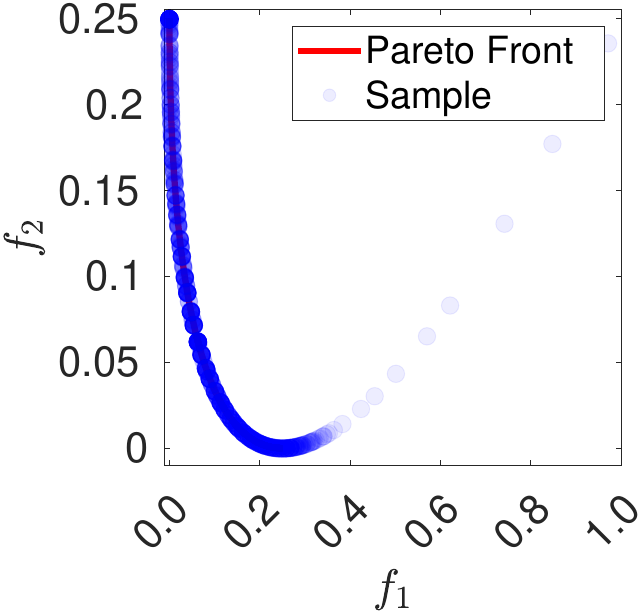}}
	\centerline{\small(h) $\gamma=1$}
	\end{minipage}
  	\begin{minipage}{0.243\linewidth}
	\centerline{\includegraphics[width=1\textwidth]{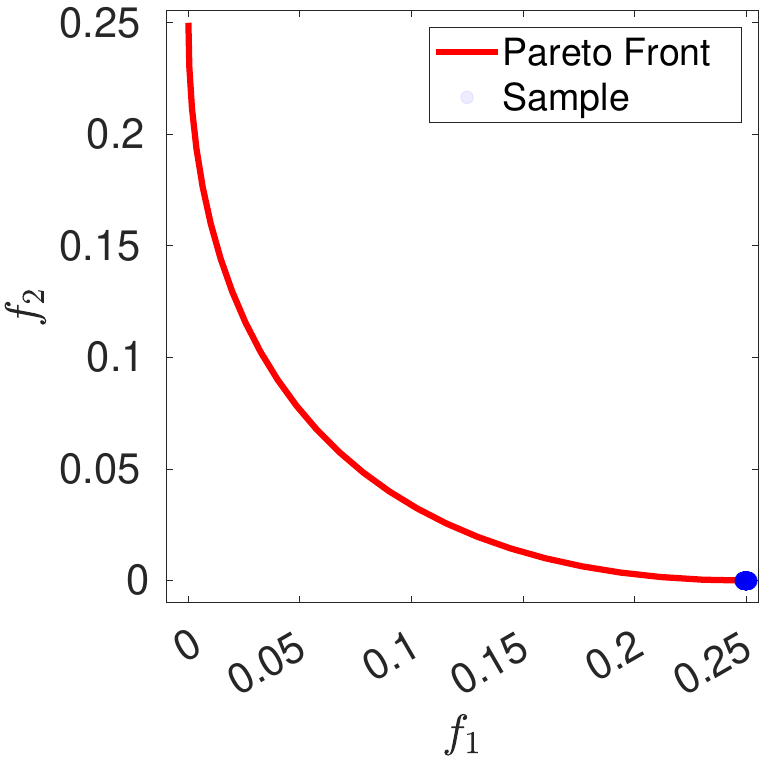}}
 	\centerline{\small(i) $w=0$}
	\end{minipage}
    \hfill
 	\begin{minipage}{0.243\linewidth}
	\centerline{\includegraphics[width=1\textwidth]{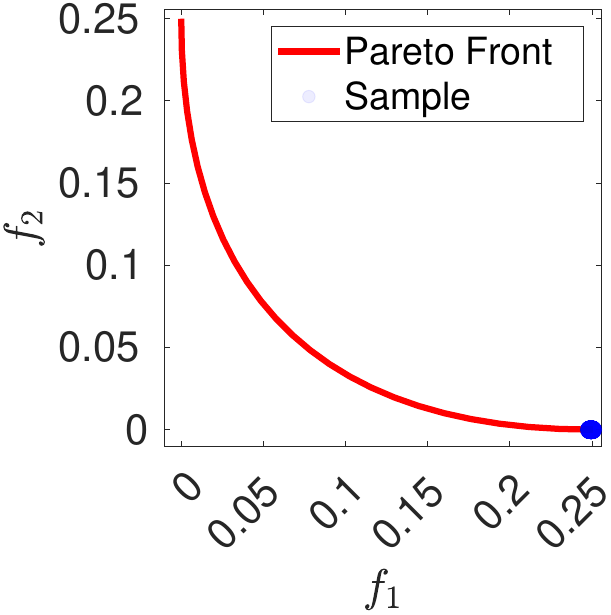}}
	\centerline{\small(j) $w=0.1$}
	\end{minipage}
     \hfill
    \begin{minipage}{0.243\linewidth}
	\centerline{\includegraphics[width=1\textwidth]{lm_div_50k.pdf}}
	\centerline{\small(k) $w=0.5$}
	\end{minipage}
     \hfill
    \begin{minipage}{0.243\linewidth}
	\centerline{\includegraphics[width=1\textwidth]{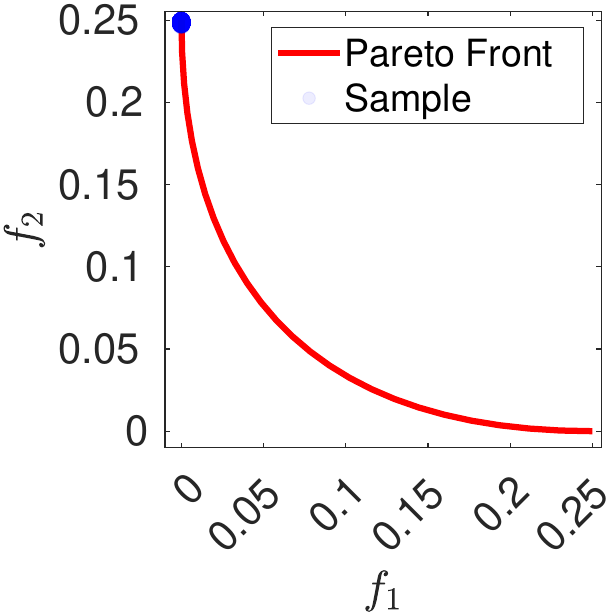}}
	\centerline{\small(l) $w=1$}
	\end{minipage}
\caption{\label{fg:ablation_study} Analysis on the effects of the diversity coefficient~$\gamma$ in Eq.\eqref{eq:diversity} to our PROUD (1st row) and DM+$m$-MGD (2nd row). As DM+single (3rd row) degenerates to SOG and does not have the diversity regularization, we conduct sensitivity analysis on its weight coefficient for combining two objectives, i.e., $(1-w)f_1 + wf_2$. The depth of color represents sample density, the deeper the higher.}
\end{figure}

We study PROUD with different configurations of the hyper-parameters, namely, $\alpha$ and $e$ in Eq.\eqref{eq:constraint_v} as well as the diversity coefficient~$\gamma$ in Eq.\eqref{eq:diversity}. The experiments are conducted on CIFAR10, with the same setting as Section~\ref{sect:cifar10}.

We set $\alpha$ as $0.1, 0.5, 1$ and $e$ as $0.01, 0.03, 0.05$, respectively. We observe in Table~\ref{tb:results_alpha} that PROUD is not sensitive to the choice of the hyper-parameters~$\alpha$ and $e$.

We set $\gamma$ as 0, 0.1, 0.2, 1. The results are summarized in Fig.~\ref{fg:ablation_study}(a) to Fig.~\ref{fg:ablation_study}(d), showing that: (1) With an appropriate diversity coefficient, our PROUD can well cover the Pareto front. (2) Without the diversity regularization, PROUD can only obtain a small set of Pareto solutions. This demonstrates the necessity of the diversity loss, consistent with the finding in the former work~\citep{liu2021profiling}. (3) With a too large value of $\gamma$, the generated samples could fall outside the Pareto front. The effect of the diversity coefficient on DM+$m$-MGD (Fig.~\ref{fg:ablation_study}(e) to Fig.~\ref{fg:ablation_study}(h)) is similar.

To further investigate the effects of the diversity coefficient on the generation quality, we collect FID results in Table~\ref{tb:results_gamma}. With~$\gamma=0.2$, PROUD obtains both the best FID and HV, which is thus set as the hyper-parameter used in Section~\ref{sect:cifar10}.

To demonstrate that the single-objective generation would fail to cover the Pareto front even with a uniform grid of weighting, we set the weight coefficient~$w$ for combining two objectives into a single objective in DM+single~``$w\times f_1(x)+(1-w)\times f_2(x)$'' as 0 to 1 with a step~$0.1$. We put the results of 0, 0.1, 0.5, 1 in Fig.~\ref{fg:ablation_study}(i) to Fig.~\ref{fg:ablation_study}(l) and rest in Appendix. 
With $w=0, 0.1, 0.2, 0.3, 0.4$, the single objective is dominated by $f_2(x)$. Consequently, the generated samples achieve the smallest value for $f_2(x)$ but the largest one for $f_1(x)$; vice versa. With an equal weight, the generated samples are supposed to obtain the comprise value between two objectives, i.e., (0.0625, 0.0625). We notice that the generated samples cover a small range around this point. This diversity could result from the diffusion noise in diffusion models.

\section{Conclusion}
This paper studies the problem of optimizing deep generative models with multiple conflicting objectives. We highlight this problem setting by treating the optimization of samples with multiple properties and the process of sample generation as a unified task. By analyzing the connections and differences from multi-objective optimization, we introduce a constrained optimization formulation to solve the multi-objective generation problem, based on which we developed PROUD. Our experiments demonstrate the efficacy of PROUD in both image and protein sequence generation. While we explored the white-box multi-objectives in this work, it would be interesting to explore our PROUD in the black-box setting in the future. The multiple gradient descent technique used can be replaced by methods such as Bayesian multi-objective optimization~\citep{stanton2022accelerating}.









\section*{Declarations}
\textbf{Funding.}
This work was supported by the Agency for Science, Technology and Research (A*STAR) Centre for Frontier AI Research, the A*STAR GAP project (Grant No.I23D1AG079), and the AISG Grand Challenge in AI for Materials Discovery (Grant No. AISG2-GC-2023-010).\\
\textbf{Competing interests.}
The authors have no financial or non-financial interests to disclose that are relevant to the content of this article.\\
\textbf{Ethics approval.}
Not applicable.\\
\textbf{Consent to participate.}
Not applicable.\\
\textbf{Consent to publish.}
Not applicable.\\
\textbf{Availability of data and materials.}
All datasets used in this work are available online and clearly cited.\\
\textbf{Code availability.}
The code of this work will be publicly released on github.\\
\textbf{Authors' contributions.}
Idea: YY; Methodology \& Experiment: YY, YP; Writing - comments/edits: all.

\bibliography{sn-bibliography}

\begin{thebibliography}{62}
\providecommand{\natexlab}[1]{#1}
\providecommand{\url}[1]{{#1}}
\providecommand{\urlprefix}{URL }
\providecommand{\doi}[1]{\url{https://doi.org/#1}}
\providecommand{\eprint}[2][]{\url{#2}}
 \bibcommenthead

\bibitem[{Afshari et~al(2019)Afshari, Hare, and Tesfamariam}]{afshari2019constrained}
Afshari H, Hare W, Tesfamariam S (2019) Constrained multi-objective optimization algorithms: Review and comparison with application in reinforced concrete structures. Applied Soft Computing 83:105631. \doi{10.1016/J.ASOC.2019.105631}

\bibitem[{Andrieu et~al(2003)Andrieu, De~Freitas, Doucet, and Jordan}]{andrieu2003introduction}
Andrieu C, De~Freitas N, Doucet A, et~al (2003) An introduction to mcmc for machine learning. Machine learning 50:5--43. \doi{10.1023/A:1020281327116}

\bibitem[{Arjovsky et~al(2017)Arjovsky, Chintala, and Bottou}]{arjovsky2017wasserstein}
Arjovsky M, Chintala S, Bottou L (2017) Wasserstein generative adversarial networks. In: International Conference on Machine Learning, pp 214--223, \urlprefix\url{https://proceedings.mlr.press/v70/arjovsky17a.html}

\bibitem[{Borghi et~al(2023)Borghi, Herty, and Pareschi}]{borghi2023adaptive}
Borghi G, Herty M, Pareschi L (2023) An adaptive consensus based method for multi-objective optimization with uniform pareto front approximation. Applied Mathematics \& Optimization 88(2):58. \doi{10.1007/s00245-023-10036-y}

\bibitem[{Cheng et~al(2017)Cheng, Li, Tian, Zhang, Yang, Jin, and Yao}]{cheng2017benchmark}
Cheng R, Li M, Tian Y, et~al (2017) A benchmark test suite for evolutionary many-objective optimization. Complex \& Intelligent Systems 3:67--81. \doi{10.1007/s40747-017-0039-7}

\bibitem[{Chinchuluun and Pardalos(2007)}]{chinchuluun2007survey}
Chinchuluun A, Pardalos PM (2007) A survey of recent developments in multiobjective optimization. Annals of Operations Research 154(1):29--50. \doi{10.1007/S10479-007-0186-0}

\bibitem[{Cock et~al(2009)Cock, Antao, Chang, Chapman, Cox, Dalke, Friedberg, Hamelryck, Kauff, Wilczynski et~al}]{cock2009biopython}
Cock PJ, Antao T, Chang JT, et~al (2009) Biopython: freely available python tools for computational molecular biology and bioinformatics. Bioinformatics 25(11):1422--1423. \doi{10.1093/bioinformatics/btp163}

\bibitem[{Dathathri et~al(2020)Dathathri, Madotto, Lan, Hung, Frank, Molino, Yosinski, and Liu}]{dathathri2019plug}
Dathathri S, Madotto A, Lan J, et~al (2020) Plug and play language models: A simple approach to controlled text generation. In: International Conference on Learning Representations, \urlprefix\url{https://openreview.net/forum?id=H1edEyBKDS}

\bibitem[{Deb(2001)}]{deb2001multi}
Deb K (2001) Multi-objective optimization using evolutionary algorithms, vol~16. John Wiley \& Sons

\bibitem[{Dem{\v{s}}ar(2006)}]{demvsar2006statistical}
Dem{\v{s}}ar J (2006) Statistical comparisons of classifiers over multiple data sets. The Journal of Machine learning research 7:1--30. \urlprefix\url{http://jmlr.org/papers/v7/demsar06a.html}

\bibitem[{Deng et~al(2020)Deng, Yang, Chen, Wen, and Tong}]{deng2020disentangled}
Deng Y, Yang J, Chen D, et~al (2020) Disentangled and controllable face image generation via 3d imitative-contrastive learning. In: IEEE/CVF Conference on Computer Vision and Pattern Recognition, pp 5154--5163, \doi{10.1109/CVPR42600.2020.00520}

\bibitem[{D{\'e}sid{\'e}ri(2012)}]{desideri2012multiple}
D{\'e}sid{\'e}ri JA (2012) Multiple-gradient descent algorithm (mgda) for multiobjective optimization. Comptes Rendus Mathematique 350(5-6):313--318. \doi{10.1016/j.crma.2012.03.014}

\bibitem[{D{\'e}sid{\'e}ri(2018)}]{desideri2018quasi}
D{\'e}sid{\'e}ri JA (2018) Quasi-riemannian multiple gradient descent algorithm for constrained multiobjective differential optimization. PhD thesis, Inria Sophia-Antipolis; Project-Team Acumes, \urlprefix\url{https://inria.hal.science/hal-01740075}

\bibitem[{Dhariwal and Nichol(2021)}]{dhariwal2021diffusion}
Dhariwal P, Nichol A (2021) Diffusion models beat gans on image synthesis. In: Advances in Neural Information Processing Systems, pp 8780--8794, \urlprefix\url{https://proceedings.neurips.cc/paper_files/paper/2021/file/49ad23d1ec9fa4bd8d77d02681df5cfa-Paper.pdf}

\bibitem[{Fefferman et~al(2016)Fefferman, Mitter, and Narayanan}]{fefferman2016testing}
Fefferman C, Mitter S, Narayanan H (2016) Testing the manifold hypothesis. Journal of the American Mathematical Society 29(4):983--1049. \doi{10.1090/jams/852}

\bibitem[{Ferruz et~al(2022)Ferruz, Schmidt, and H{\"o}cker}]{ferruz2022protgpt2}
Ferruz N, Schmidt S, H{\"o}cker B (2022) Protgpt2 is a deep unsupervised language model for protein design. Nature Communications 13(1):4348. \doi{10.1038/s41467-022-32007-7}

\bibitem[{Gong et~al(2021)Gong, Liu, and Liu}]{gong2021bi}
Gong C, Liu X, Liu Q (2021) Bi-objective trade-off with dynamic barrier gradient descent. In: Advances in Neural Information Processing Systems, pp 29630--29642, \urlprefix\url{https://proceedings.neurips.cc/paper_files/paper/2021/file/f7b027d45fd7484f6d0833823b98907e-Paper.pdf}

\bibitem[{Goodfellow et~al(2014)Goodfellow, Pouget-Abadie, Mirza, Xu, Warde-Farley, Ozair, Courville, and Bengio}]{goodfellow2014generative}
Goodfellow IJ, Pouget-Abadie J, Mirza M, et~al (2014) Generative adversarial nets. In: Advances in Neural Information Processing Systems, pp 2672--2680, \urlprefix\url{https://proceedings.neurips.cc/paper_files/paper/2014/file/5ca3e9b122f61f8f06494c97b1afccf3-Paper.pdf}

\bibitem[{Gruver et~al(2023)Gruver, Stanton, Frey, Rudner, Hotzel, Lafrance-Vanasse, Rajpal, Cho, and Wilson}]{gruver2023protein}
Gruver N, Stanton S, Frey NC, et~al (2023) Protein design with guided discrete diffusion. In: Advances in Neural Information Processing Systems, pp 12489--12517, \urlprefix\url{https://proceedings.neurips.cc/paper_files/paper/2023/file/29591f355702c3f4436991335784b503-Paper-Conference.pdf}

\bibitem[{Guo et~al(2020)Guo, Du, and Zhao}]{guo2020property}
Guo X, Du Y, Zhao L (2020) Property controllable variational autoencoder via invertible mutual dependence. In: International Conference on Learning Representations, \urlprefix\url{https://openreview.net/forum?id=tYxG\_OMs9WE}

\bibitem[{Heusel et~al(2017)Heusel, Ramsauer, Unterthiner, Nessler, and Hochreiter}]{heusel2017gans}
Heusel M, Ramsauer H, Unterthiner T, et~al (2017) Gans trained by a two time-scale update rule converge to a local nash equilibrium. In: Advances in Neural Information Processing Systems, pp 6629--6640, \urlprefix\url{https://proceedings.neurips.cc/paper_files/paper/2017/file/8a1d694707eb0fefe65871369074926d-Paper.pdf}

\bibitem[{Ho et~al(2020)Ho, Jain, and Abbeel}]{ho2020denoising}
Ho J, Jain A, Abbeel P (2020) Denoising diffusion probabilistic models. In: Advances in Neural Information Processing Systems, pp 6840--6851, \urlprefix\url{https://proceedings.neurips.cc/paper/2020/file/4c5bcfec8584af0d967f1ab10179ca4b-Paper.pdf}

\bibitem[{Ishibuchi et~al(2008)Ishibuchi, Tsukamoto, and Nojima}]{ishibuchi2008evolutionary}
Ishibuchi H, Tsukamoto N, Nojima Y (2008) Evolutionary many-objective optimization: A short review. In: IEEE Congress on Evolutionary Computation, pp 2419--2426, \doi{10.1109/CEC.2008.4631121}

\bibitem[{Ishibuchi et~al(2013)Ishibuchi, Yamane, Akedo, and Nojima}]{ishibuchi2013many}
Ishibuchi H, Yamane M, Akedo N, et~al (2013) Many-objective and many-variable test problems for visual examination of multiobjective search. In: IEEE Congress on Evolutionary Computation, pp 1491--1498, \doi{10.1109/CEC.2013.6557739}

\bibitem[{Jain et~al(2023)Jain, Raparthy, Hern{\'a}ndez-Garc{\i}a, Rector-Brooks, Bengio, Miret, and Bengio}]{jain2023multi}
Jain M, Raparthy SC, Hern{\'a}ndez-Garc{\i}a A, et~al (2023) Multi-objective gflownets. In: International Conference on Machine Learning, pp 14631--14653, \urlprefix\url{https://proceedings.mlr.press/v202/jain23a.html}

\bibitem[{Jin et~al(2020)Jin, Barzilay, and Jaakkola}]{jin2020multi}
Jin W, Barzilay R, Jaakkola T (2020) Multi-objective molecule generation using interpretable substructures. In: International Conference on Machine Learning, pp 4849--4859, \urlprefix\url{http://proceedings.mlr.press/v119/jin20b.html}

\bibitem[{Kingma and Welling(2014)}]{DBLP:journals/corr/KingmaW13}
Kingma DP, Welling M (2014) Auto-encoding variational bayes. In: International Conference on Learning Representations, \urlprefix\url{https://openreview.net/forum?id=33X9fd2-9FyZd}

\bibitem[{Klys et~al(2018)Klys, Snell, and Zemel}]{klys2018learning}
Klys J, Snell J, Zemel R (2018) Learning latent subspaces in variational autoencoders. In: Advances in Neural Information Processing Systems, pp 6445--6455, \urlprefix\url{https://proceedings.neurips.cc/paper_files/paper/2018/file/73e5080f0f3804cb9cf470a8ce895dac-Paper.pdf}

\bibitem[{Krizhevsky et~al(2009)Krizhevsky, Hinton et~al}]{krizhevsky2009learning}
Krizhevsky A, Hinton G, et~al (2009) Learning multiple layers of features from tiny images \urlprefix\url{https://www.cs.utoronto.ca/~kriz/learning-features-2009-TR.pdf}

\bibitem[{Li et~al(2017)Li, Grosan, Yang, Liu, and Yao}]{li2017multiline}
Li M, Grosan C, Yang S, et~al (2017) Multiline distance minimization: A visualized many-objective test problem suite. IEEE Transactions on Evolutionary Computation 22(1):61--78. \doi{10.1109/TEVC.2017.2655451}

\bibitem[{Li et~al(2022)Li, Liu, and Walder}]{li2022editvae}
Li S, Liu M, Walder C (2022) Editvae: Unsupervised parts-aware controllable 3d point cloud shape generation. In: AAAI Conference on Artificial Intelligence, pp 1386--1394, \doi{10.1609/AAAI.V36I2.20027}

\bibitem[{Liao et~al(2020)Liao, Schwarz, Mescheder, and Geiger}]{liao2020towards}
Liao Y, Schwarz K, Mescheder L, et~al (2020) Towards unsupervised learning of generative models for 3d controllable image synthesis. In: IEEE/CVF Conference on Computer Vision and Pattern Recognition, pp 5871--5880, \doi{10.1109/CVPR42600.2020.00591}

\bibitem[{Liu et~al(2021{\natexlab{a}})Liu, Tong, and Liu}]{liu2021profiling}
Liu X, Tong X, Liu Q (2021{\natexlab{a}}) Profiling pareto front with multi-objective stein variational gradient descent. In: Advances in Neural Information Processing Systems, pp 14721--14733, \urlprefix\url{https://proceedings.neurips.cc/paper/2021/file/7bb16972da003e87724f048d76b7e0e1-Paper.pdf}

\bibitem[{Liu et~al(2021{\natexlab{b}})Liu, Tong, and Liu}]{liu2021sampling}
Liu X, Tong X, Liu Q (2021{\natexlab{b}}) Sampling with trusthworthy constraints: A variational gradient framework. In: Advances in Neural Information Processing Systems, pp 23557--23568, \urlprefix\url{https://papers.nips.cc/paper/2021/file/c61aed648da48aa3893fb3eaadd88a7f-Paper.pdf}

\bibitem[{McInnes et~al(2018)McInnes, Healy, and Melville}]{mcinnes2018umap}
McInnes L, Healy J, Melville J (2018) Umap: Uniform manifold approximation and projection for dimension reduction. arXiv preprint \urlprefix\url{http://arxiv.org/abs/1802.03426}

\bibitem[{Olsen et~al(2022)Olsen, Boyles, and Deane}]{olsen2022observed}
Olsen TH, Boyles F, Deane CM (2022) Observed antibody space: A diverse database of cleaned, annotated, and translated unpaired and paired antibody sequences. Protein Science 31(1):141--146. \doi{10.1002/pro.4205}

\bibitem[{Papamakarios et~al(2021)Papamakarios, Nalisnick, Rezende, Mohamed, and Lakshminarayanan}]{papamakarios2021normalizing}
Papamakarios G, Nalisnick E, Rezende DJ, et~al (2021) Normalizing flows for probabilistic modeling and inference. Journal of Machine Learning Research 22(57):1--64. \urlprefix\url{http://jmlr.org/papers/v22/19-1028.html}

\bibitem[{Roweis and Saul(2000)}]{roweis2000nonlinear}
Roweis ST, Saul LK (2000) Nonlinear dimensionality reduction by locally linear embedding. Science 290(5500):2323--2326. \doi{10.1126/science.290.5500.2323}

\bibitem[{Ruffolo et~al(2023)Ruffolo, Chu, Mahajan, and Gray}]{ruffolo2023fast}
Ruffolo JA, Chu LS, Mahajan SP, et~al (2023) Fast, accurate antibody structure prediction from deep learning on massive set of natural antibodies. Nature Communications 14(1):2389. \doi{10.5281/zenodo.7709609}

\bibitem[{Sanchez-Lengeling and Aspuru-Guzik(2018)}]{sanchez2018inverse}
Sanchez-Lengeling B, Aspuru-Guzik A (2018) Inverse molecular design using machine learning: Generative models for matter engineering. Science 361(6400):360--365. \doi{10.1126/science.aat2663}

\bibitem[{Sener and Koltun(2018)}]{sener2018multi}
Sener O, Koltun V (2018) Multi-task learning as multi-objective optimization. In: Advances in Neural Information Processing Systems, pp 525--536, \urlprefix\url{https://proceedings.neurips.cc/paper/2018/file/432aca3a1e345e339f35a30c8f65edce-Paper.pdf}

\bibitem[{Shen et~al(2023)Shen, Bengio, Hajiramezanali, Loukas, Cho, and Biancalani}]{shen2023towards}
Shen MW, Bengio E, Hajiramezanali E, et~al (2023) Towards understanding and improving gflownet training. In: International Conference on Machine Learning, pp 30956--30975, \urlprefix\url{https://proceedings.mlr.press/v202/shen23a.html}

\bibitem[{Sohl-Dickstein et~al(2015)Sohl-Dickstein, Weiss, Maheswaranathan, and Ganguli}]{sohl2015deep}
Sohl-Dickstein J, Weiss E, Maheswaranathan N, et~al (2015) Deep unsupervised learning using nonequilibrium thermodynamics. In: International Conference on Machine Learning, pp 2256--2265, \urlprefix\url{http://proceedings.mlr.press/v37/sohl-dickstein15.html}

\bibitem[{Song and Ermon(2019)}]{song2019generative}
Song Y, Ermon S (2019) Generative modeling by estimating gradients of the data distribution. In: Advances in Neural Information Processing Systems, pp 11918--11930, \urlprefix\url{https://proceedings.neurips.cc/paper_files/paper/2019/file/3001ef257407d5a371a96dcd947c7d93-Paper.pdf}

\bibitem[{Song and Ermon(2020)}]{song2020improved}
Song Y, Ermon S (2020) Improved techniques for training score-based generative models. In: Advances in Neural Information Processing Systems, pp 12438--12448, \urlprefix\url{https://papers.neurips.cc/paper_files/paper/2020/file/92c3b916311a5517d9290576e3ea37ad-Paper.pdf}

\bibitem[{Song and Kingma(2021)}]{song2021train}
Song Y, Kingma DP (2021) How to train your energy-based models. arXiv preprint \urlprefix\url{https://arxiv.org/abs/2101.03288}

\bibitem[{Song et~al(2021{\natexlab{a}})Song, Durkan, Murray, and Ermon}]{song2021maximum}
Song Y, Durkan C, Murray I, et~al (2021{\natexlab{a}}) Maximum likelihood training of score-based diffusion models. In: Advances in Neural Information Processing Systems, pp 1415--1428, \urlprefix\url{https://papers.nips.cc/paper/2021/file/0a9fdbb17feb6ccb7ec405cfb85222c4-Paper.pdf}

\bibitem[{Song et~al(2021{\natexlab{b}})Song, Sohl-Dickstein, Kingma, Kumar, Ermon, and Poole}]{song2021score}
Song Y, Sohl-Dickstein J, Kingma DP, et~al (2021{\natexlab{b}}) Score-based generative modeling through stochastic differential equations. In: International Conference on Learning Representations, \urlprefix\url{https://openreview.net/forum?id=PxTIG12RRHS}

\bibitem[{Stanton et~al(2022)Stanton, Maddox, Gruver, Maffettone, Delaney, Greenside, and Wilson}]{stanton2022accelerating}
Stanton S, Maddox W, Gruver N, et~al (2022) Accelerating bayesian optimization for biological sequence design with denoising autoencoders. In: International Conference on Machine Learning, pp 20459--20478, \urlprefix\url{https://proceedings.mlr.press/v162/stanton22a.html}

\bibitem[{Tagasovska et~al(2022)Tagasovska, Frey, Loukas, Hotzel, Lafrance-Vanasse, Kelly, Wu, Rajpal, Bonneau, Cho et~al}]{tagasovska2022pareto}
Tagasovska N, Frey NC, Loukas A, et~al (2022) A pareto-optimal compositional energy-based model for sampling and optimization of protein sequences. In: NeurIPS 2022 Workshop AI for Science: Progress and Promises, \urlprefix\url{https://openreview.net/forum?id=U2rNXaTTXPQ}

\bibitem[{Tanabe and Ishibuchi(2020)}]{tanabe2020easy}
Tanabe R, Ishibuchi H (2020) An easy-to-use real-world multi-objective optimization problem suite. Applied Soft Computing 89:106078. \doi{10.1016/J.ASOC.2020.106078}

\bibitem[{Van~Veldhuizen et~al(1998)Van~Veldhuizen, Lamont et~al}]{van1998evolutionary}
Van~Veldhuizen DA, Lamont GB, et~al (1998) Evolutionary computation and convergence to a pareto front. In: Late breaking papers at the genetic programming 1998 conference, pp 221--228, \urlprefix\url{https://citeseerx.ist.psu.edu/document?repid=rep1&type=pdf&doi=f329eb18a4549daa83fae28043d19b83fe8356fa}

\bibitem[{Wang et~al(2022)Wang, Guo, Lin, Pan, Du, Wang, Ye, Petersen, Leitgeb, AlKhalifa et~al}]{wang2022multi}
Wang S, Guo X, Lin X, et~al (2022) Multi-objective deep data generation with correlated property control. In: Advances in Neural Information Processing Systems, pp 28889--28901, \urlprefix\url{https://proceedings.neurips.cc/paper_files/paper/2022/file/b9c2e8a0bbed5fcfaf62856a3a719ada-Paper-Conference.pdf}

\bibitem[{Wang et~al(2024)Wang, Du, Guo, Pan, Qin, and Zhao}]{wang2022controllable}
Wang S, Du Y, Guo X, et~al (2024) Controllable data generation by deep learning: A review. ACM Comput Surv 56(9). \doi{10.1145/3648609}

\bibitem[{Wang et~al(2023)Wang, Zhao, and Xing}]{wang2023stylediffusion}
Wang Z, Zhao L, Xing W (2023) Stylediffusion: Controllable disentangled style transfer via diffusion models. In: IEEE/CVF International Conference on Computer Vision, pp 7677--7689, \doi{10.1109/ICCV51070.2023.00706}

\bibitem[{Watson et~al(2023)Watson, Juergens, Bennett, Trippe, Yim, Eisenach, Ahern, Borst, Ragotte, Milles et~al}]{watson2023novo}
Watson JL, Juergens D, Bennett NR, et~al (2023) De novo design of protein structure and function with rfdiffusion. Nature 620(7976):1089--1100. \doi{10.1038/s41586-023-06415-8}

\bibitem[{Welling and Teh(2011)}]{welling2011bayesian}
Welling M, Teh YW (2011) Bayesian learning via stochastic gradient langevin dynamics. In: International Conference on Machine Learning, pp 681--688, \urlprefix\url{https://icml.cc/2011/papers/398\_icmlpaper.pdf}

\bibitem[{Yang et~al(2023)Yang, Zhang, Song, Hong, Xu, Zhao, Zhang, Cui, and Yang}]{yang2022diffusion}
Yang L, Zhang Z, Song Y, et~al (2023) Diffusion models: A comprehensive survey of methods and applications. ACM Computing Surveys 56(4):1--39. \doi{10.1145/3626235}

\bibitem[{Yao et~al(2022)Yao, Gao, Yang, Sun, Zhang, and Huang}]{yao2022outpainting}
Yao K, Gao P, Yang X, et~al (2022) Outpainting by queries. In: European Conference on Computer Vision, pp 153--169, \doi{10.1007/978-3-031-20050-2\_10}

\bibitem[{Ye and Liu(2022)}]{ye2022pareto}
Ye M, Liu Q (2022) Pareto navigation gradient descent: a first-order algorithm for optimization in pareto set. In: Uncertainty in Artificial Intelligence, pp 2246--2255, \urlprefix\url{https://proceedings.mlr.press/v180/ye22a.html}

\bibitem[{Zhang et~al(2023)Zhang, Qian, Huang, Zhang, Xiao, He, and Lu}]{zhang2023robust}
Zhang S, Qian Z, Huang K, et~al (2023) Robust generative adversarial network. Machine Learning 112:5135--5161. \doi{10.1007/s10994-023-06367-0}

\bibitem[{Zitzler and Thiele(1999)}]{zitzler1999multiobjective}
Zitzler E, Thiele L (1999) Multiobjective evolutionary algorithms: a comparative case study and the strength pareto approach. IEEE transactions on Evolutionary Computation 3(4):257--271. \doi{10.1109/4235.797969}

\end{thebibliography}

\newpage
\begin{appendices}

\section{Complete sensitivity analysis for single-objective generation}\label{secA1}
\begin{figure}[!htb]
\centering
    \begin{minipage}{0.243\linewidth}
	\centerline{\includegraphics[width=1\textwidth]{Ab_ls_div_0.pdf}}
	\centerline{\small(a) $w=0$}
	\end{minipage}
     \hfill
  	\begin{minipage}{0.243\linewidth}
	\centerline{\includegraphics[width=1\textwidth]{Ab_ls_div_0.1.pdf}}
 	\centerline{\small(b) $w=0.1$}
	\end{minipage}
    \hfill
 	\begin{minipage}{0.243\linewidth}
	\centerline{\includegraphics[width=1\textwidth]{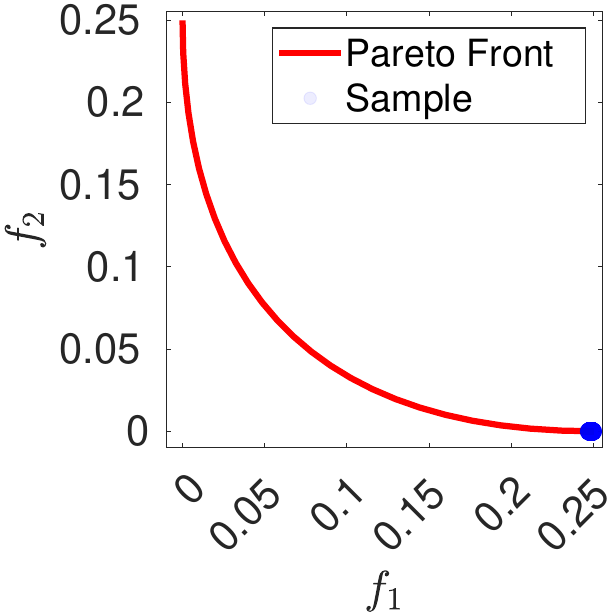}}
	\centerline{\small(c) $w=0.2$}
	\end{minipage}
     \hfill
    \begin{minipage}{0.243\linewidth}
	\centerline{\includegraphics[width=1\textwidth]{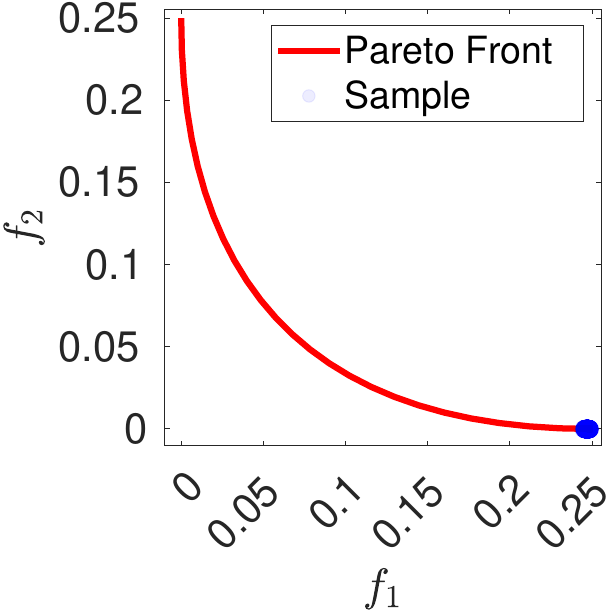}}
	\centerline{\small(d) $w=0.3$}
	\end{minipage}
     \hfill
    \begin{minipage}{0.243\linewidth}
	\centerline{\includegraphics[width=1\textwidth]{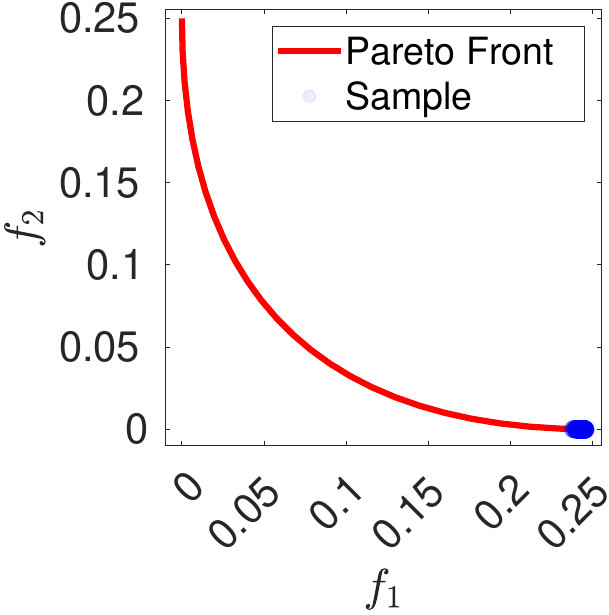}}
	\centerline{\small(e) $w=0.4$}
	\end{minipage}
     \hfill
    \begin{minipage}{0.243\linewidth}
	\centerline{\includegraphics[width=1\textwidth]{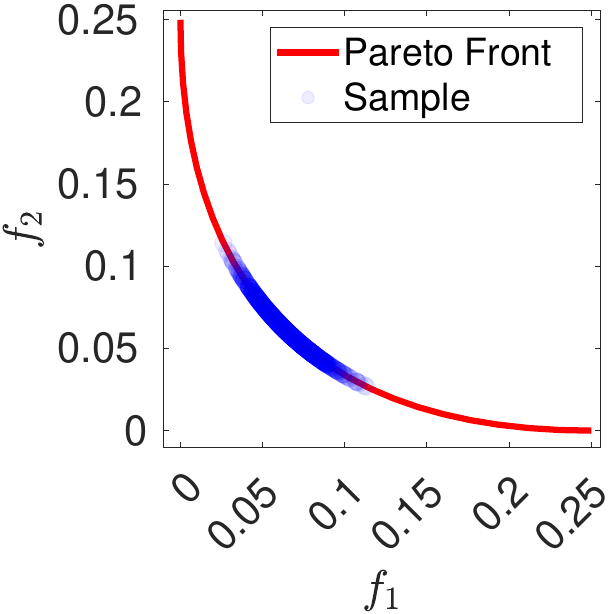}}
	\centerline{\small(f) $w=0.5$}
	\end{minipage}
     \hfill
  	\begin{minipage}{0.243\linewidth}
	\centerline{\includegraphics[width=1\textwidth]{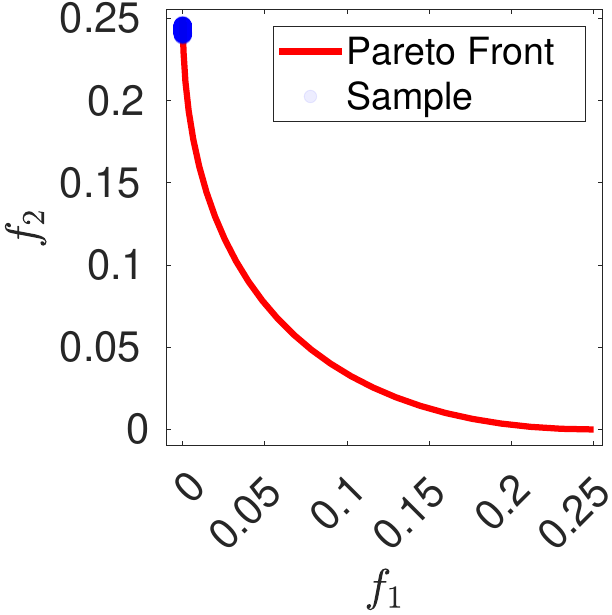}}
 	\centerline{\small(g) $w=0.6$}
	\end{minipage}
    \hfill
 	\begin{minipage}{0.243\linewidth}
	\centerline{\includegraphics[width=1\textwidth]{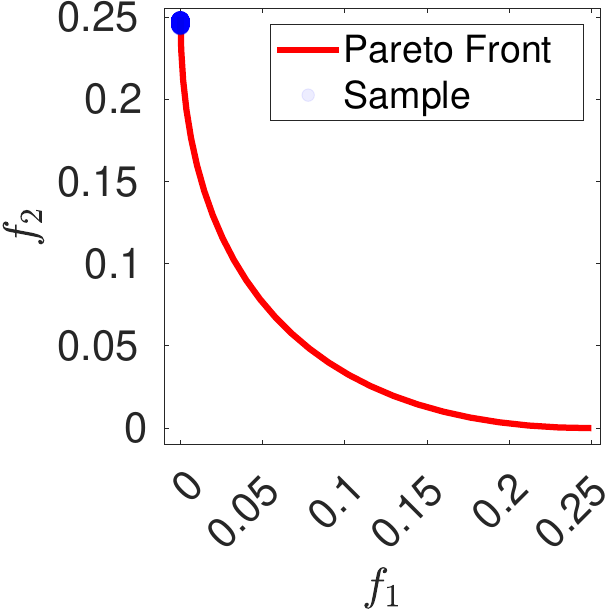}}
	\centerline{\small(h) $w=0.7$}
	\end{minipage}
     \hfill
    \begin{minipage}{0.243\linewidth}
	\centerline{\includegraphics[width=1\textwidth]{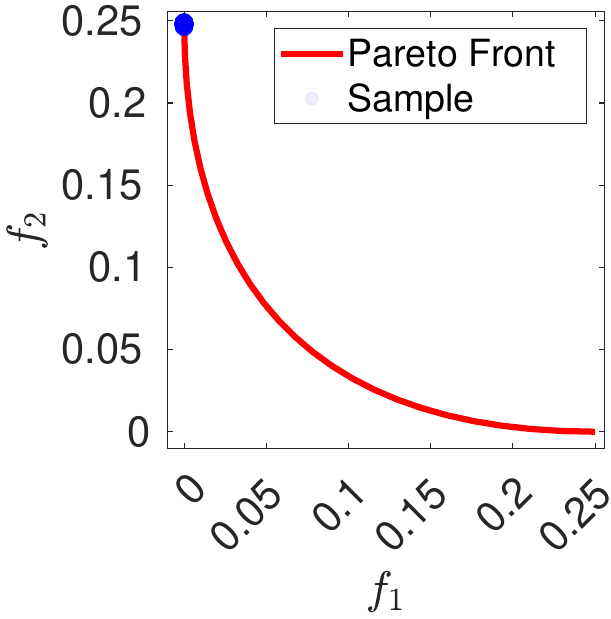}}
	\centerline{\small(i) $w=0.8$}
	\end{minipage}
     \hfill
    \begin{minipage}{0.243\linewidth}
	\centerline{\includegraphics[width=1\textwidth]{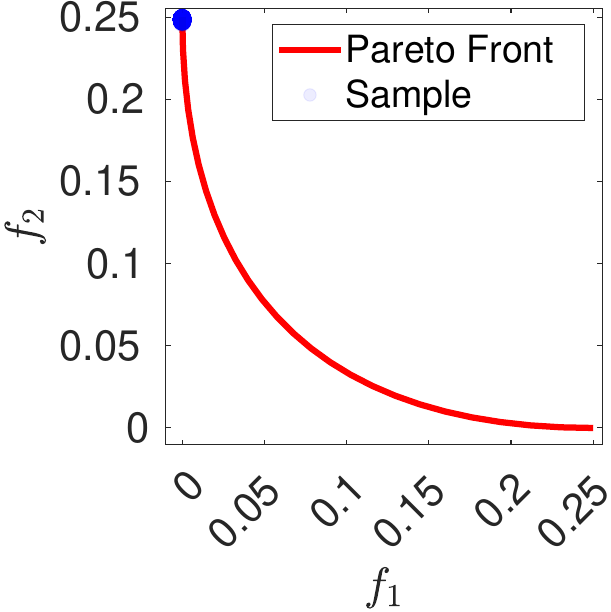}}
	\centerline{\small(j) $w=0.9$}
	\end{minipage}
     \hfill
     \begin{minipage}{0.243\linewidth}
	\centerline{\includegraphics[width=1\textwidth]{Ab_ls_div_1.pdf}}
	\centerline{\small(k) $w=1$}
	\end{minipage}
\caption{\label{fg:ablation_study_DM_single}  Sensitivity analysis on the weight coefficient for combining two objectives, i.e., $(1-w)f_1 + wf_2$ in DM+single. The depth of color represents sample density, the deeper the higher.}	
\end{figure}

We set the weight coefficient~$w$ for combining two objectives in DM+single~``$w\times f_1(x)+(1-w)\times f_2(x)$'' as 0 to 1 with a step~$0.1$. The results is shown in Fig.~\ref{fg:ablation_study_DM_single}: 
\begin{itemize}
\item  when $w < 0.5$,  the resultant final objective is dominated by $f_2(x)$. Consequently, the leading objective is optimized to the best where all the generated samples have the smallest value for $f_2(x)$ but the largest one for $f_1(x)$ . 

\item when $w > 0.5$,  the resultant final objective is dominated by $f_1(x)$. Therefore, the generated samples achieve the smallest value for the first objective but the largest one for the second objective. 

\item when $w = 0.5 =\frac{1}{m}$, the generated samples are supposed to obtain the comprise value between~$f_1(x)$ and $f_2(x)$, i.e.,  (0.0625, 0.0625). We notice that the generated samples cover a small range around this point. This diversity could result from the diffusion noise in diffusion models.
\end{itemize} 

\begin{figure}[!tb]
\centering
	\begin{minipage}{0.32\linewidth}
	\centerline{\includegraphics[width=1\textwidth]{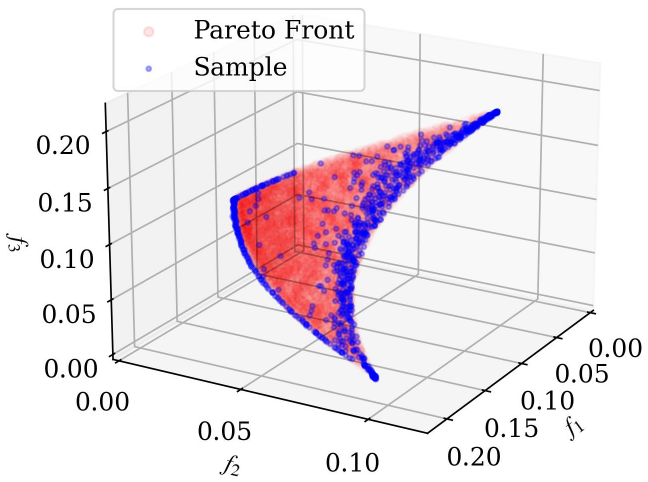}}
 	\centerline{\small(a) $\gamma=0.8$}
	\end{minipage}
    \hfill
 	 \begin{minipage}{0.32\linewidth}
	\centerline{\includegraphics[width=1\textwidth]{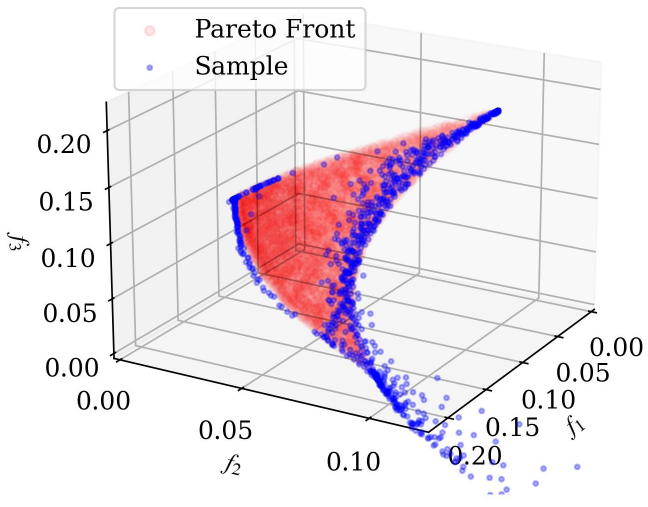}}
  	\centerline{\small(b) $\gamma=1$}
	\end{minipage}
    \hfill
 	\begin{minipage}{0.32\linewidth}
	\centerline{\includegraphics[width=1\textwidth]{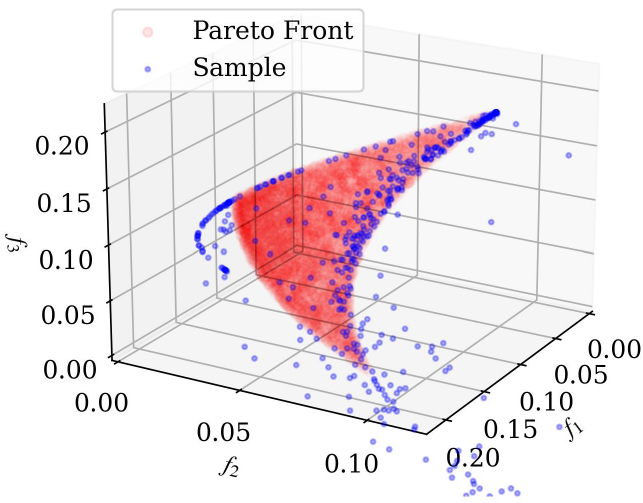}}
	\centerline{\small(c) $\gamma=1.5$}
	\end{minipage}
\caption{\label{fg:APP_dm_mgd_3obj} \revision{Different diversity coefficient~$\gamma$ for DM+$m$-MGD on CIFAR10 optimized with three objectives. 1,000 generated samples are randomly selected for visualization.}}	\vskip-0.08in
\end{figure}

\begin{figure}[!tb]
\centering
	\begin{minipage}{0.32\linewidth}
	\centerline{\includegraphics[width=1\textwidth]{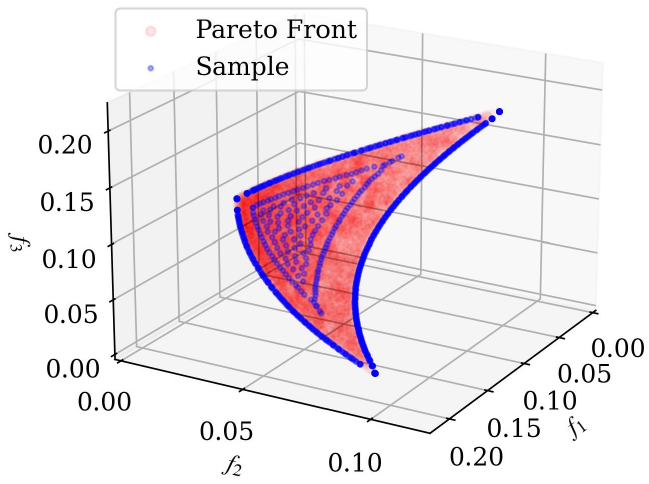}}
 	\centerline{\small(a) $\gamma=0.01$}
	\end{minipage}
    \hfill
 	 \begin{minipage}{0.32\linewidth}
	\centerline{\includegraphics[width=1\textwidth]{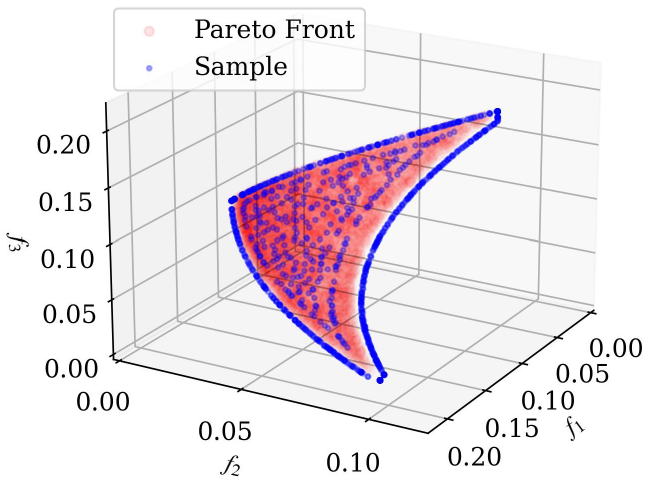}}
  	\centerline{\small(b) $\gamma=0.05$}
	\end{minipage}
    \hfill
 	\begin{minipage}{0.32\linewidth}
	\centerline{\includegraphics[width=1\textwidth]{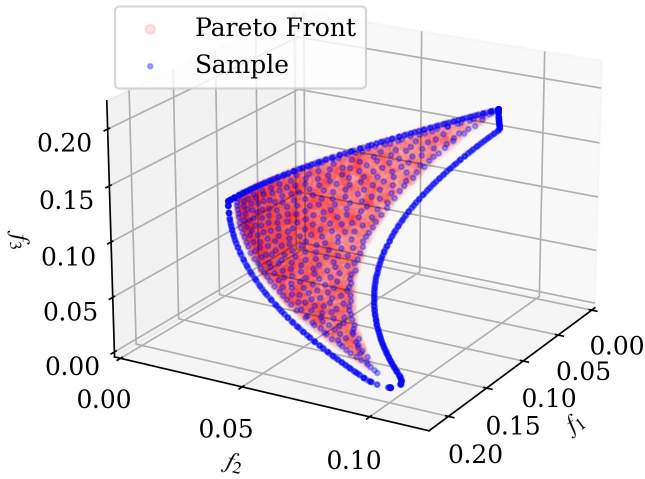}}
	\centerline{\small(c) $\gamma=0.1$}
	\end{minipage}
\caption{\label{fg:APP_mgd_only_3obj} Different diversity coefficient~$\gamma$ for $m$-MGD on CIFAR10 optimized with three objectives. 1,000 generated samples are randomly selected for visualization.}	\vskip-0.08in
\end{figure}

\begin{figure}[!tb]
\centering
	\begin{minipage}{0.32\linewidth}
	\centerline{\includegraphics[width=1\textwidth]{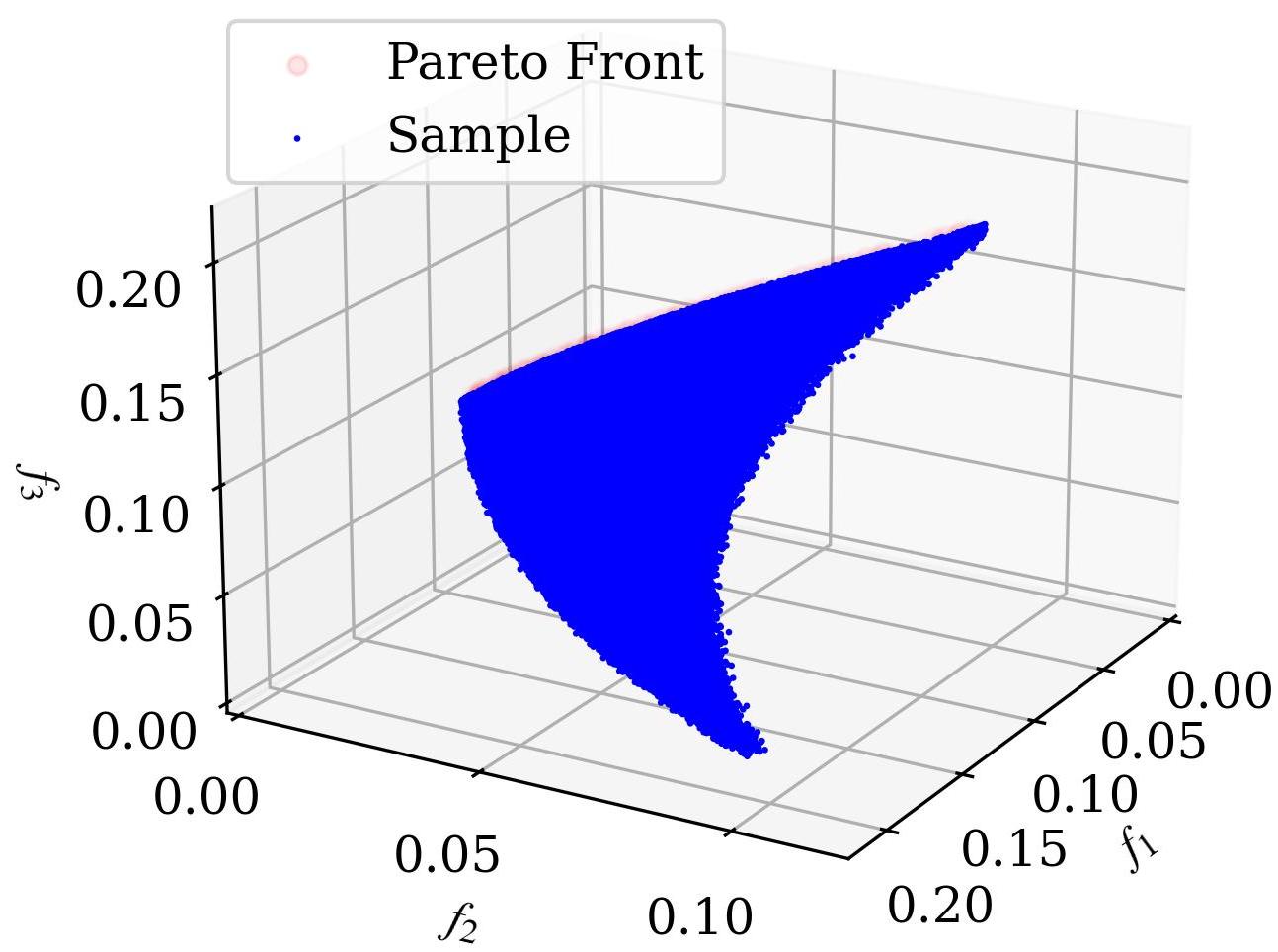}}
	\end{minipage}
    \hfill
 	 \begin{minipage}{0.32\linewidth}
	\centerline{\includegraphics[width=1\textwidth]{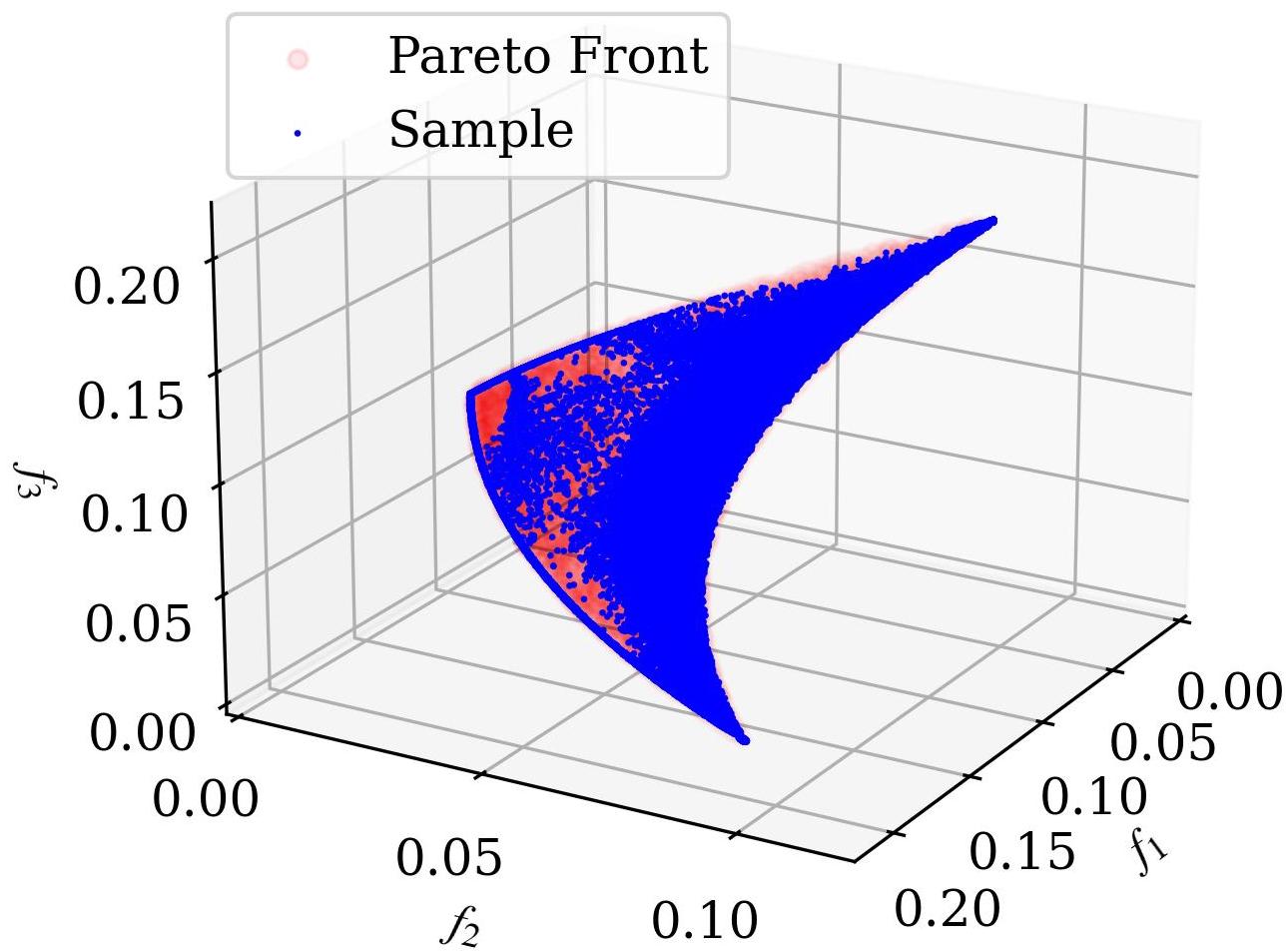}}
	\end{minipage}
    \hfill
 	\begin{minipage}{0.32\linewidth}
	\centerline{\includegraphics[width=1\textwidth]{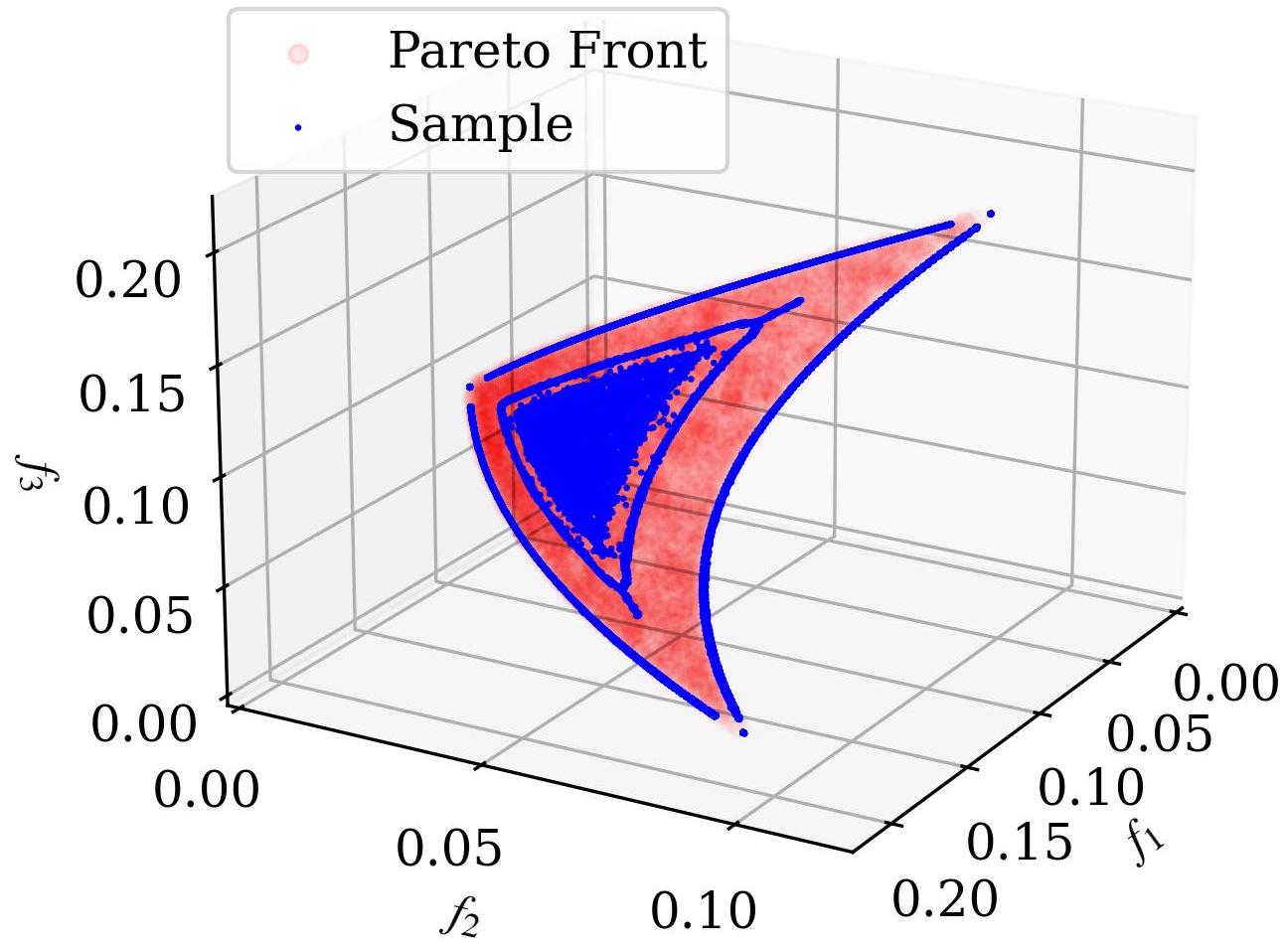}}
	\end{minipage}
    \begin{minipage}{0.32\linewidth}
	\centerline{\includegraphics[width=1\textwidth]{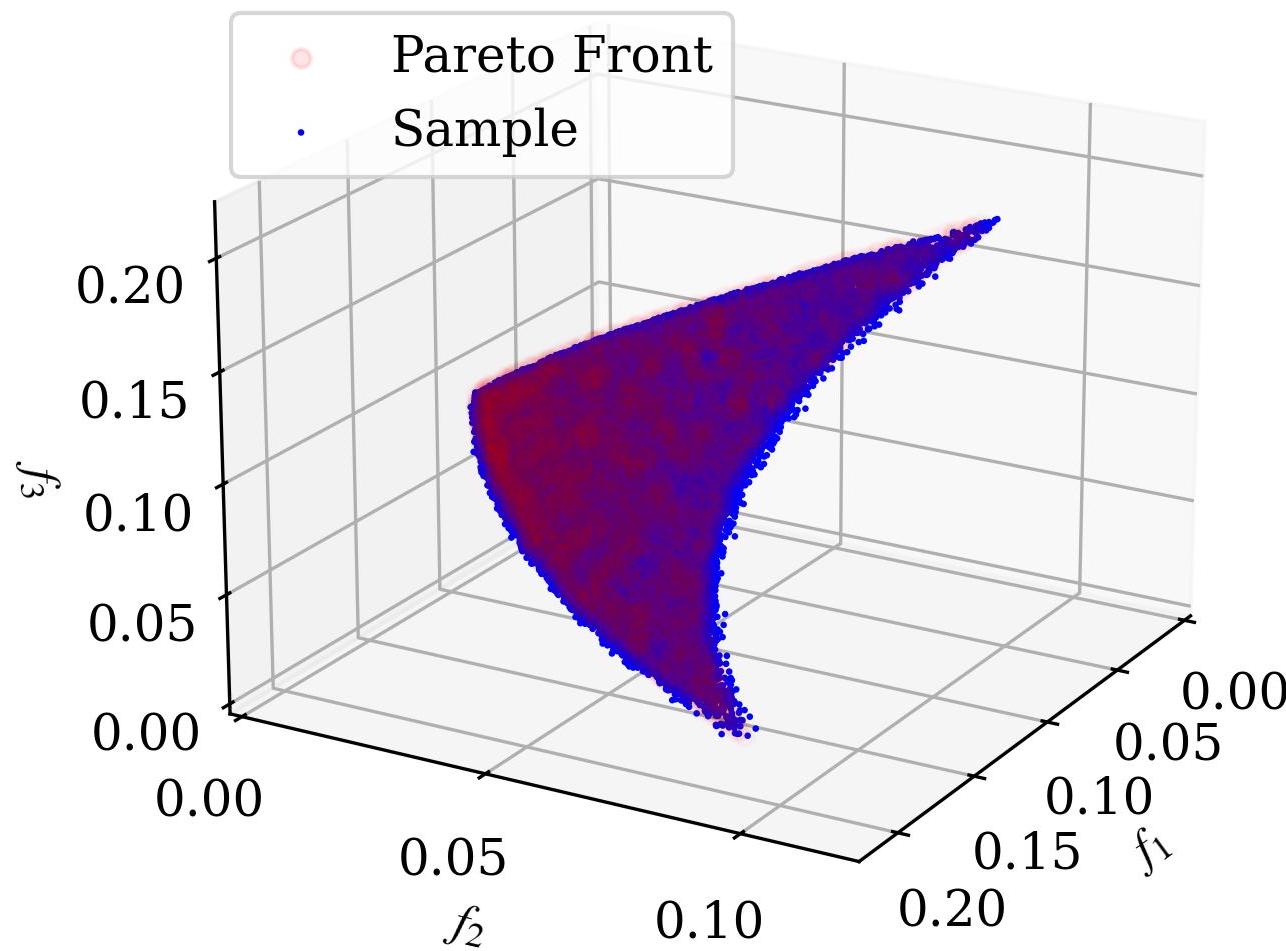}}
 	\centerline{\small(a) PROUD}
	\end{minipage}
    \hfill
 	 \begin{minipage}{0.32\linewidth}
	\centerline{\includegraphics[width=1\textwidth]{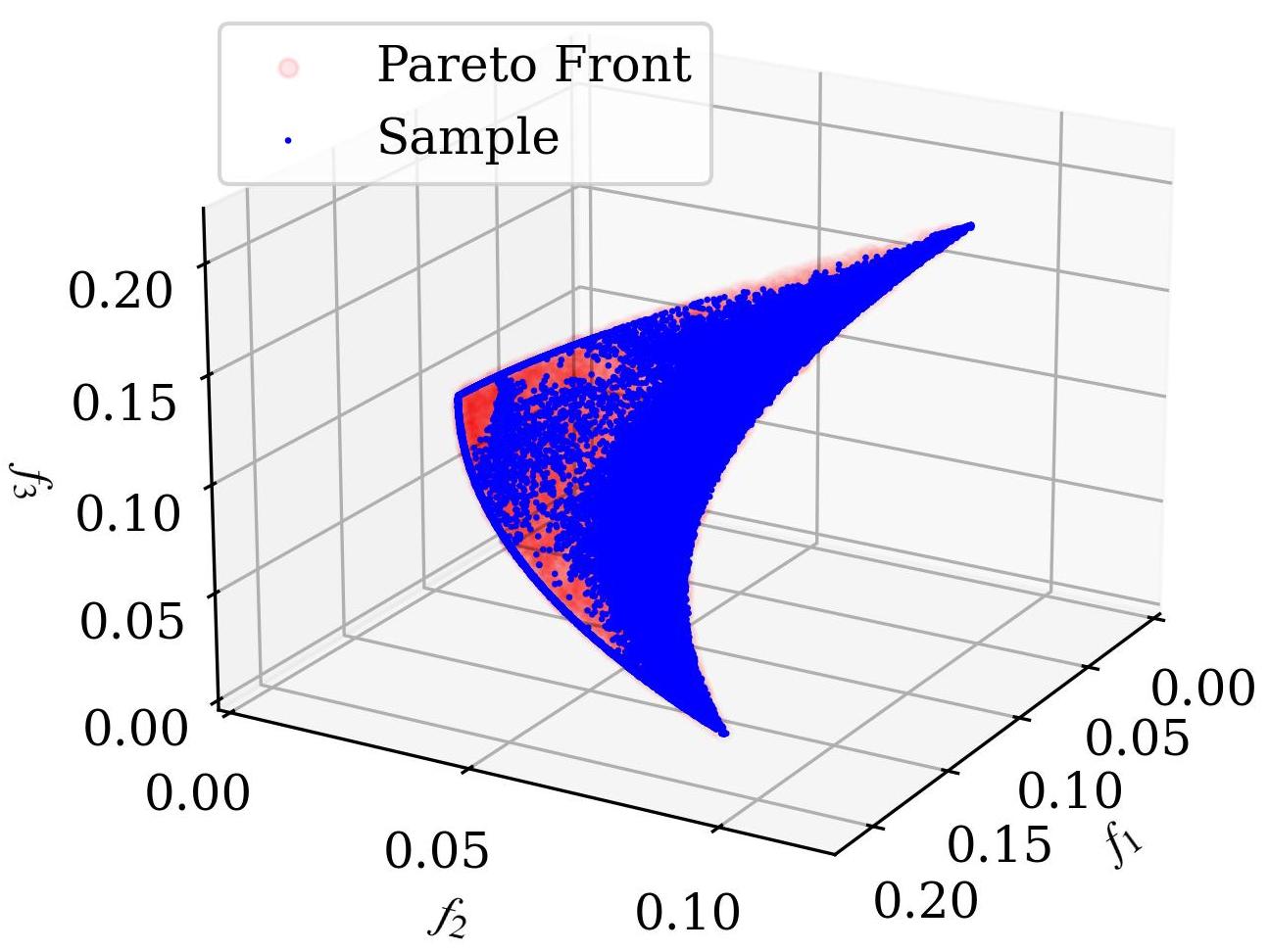}}
  	\centerline{\small(b) DM+$m$-MGD}
	\end{minipage}
    \hfill
 	\begin{minipage}{0.32\linewidth}
	\centerline{\includegraphics[width=1\textwidth]{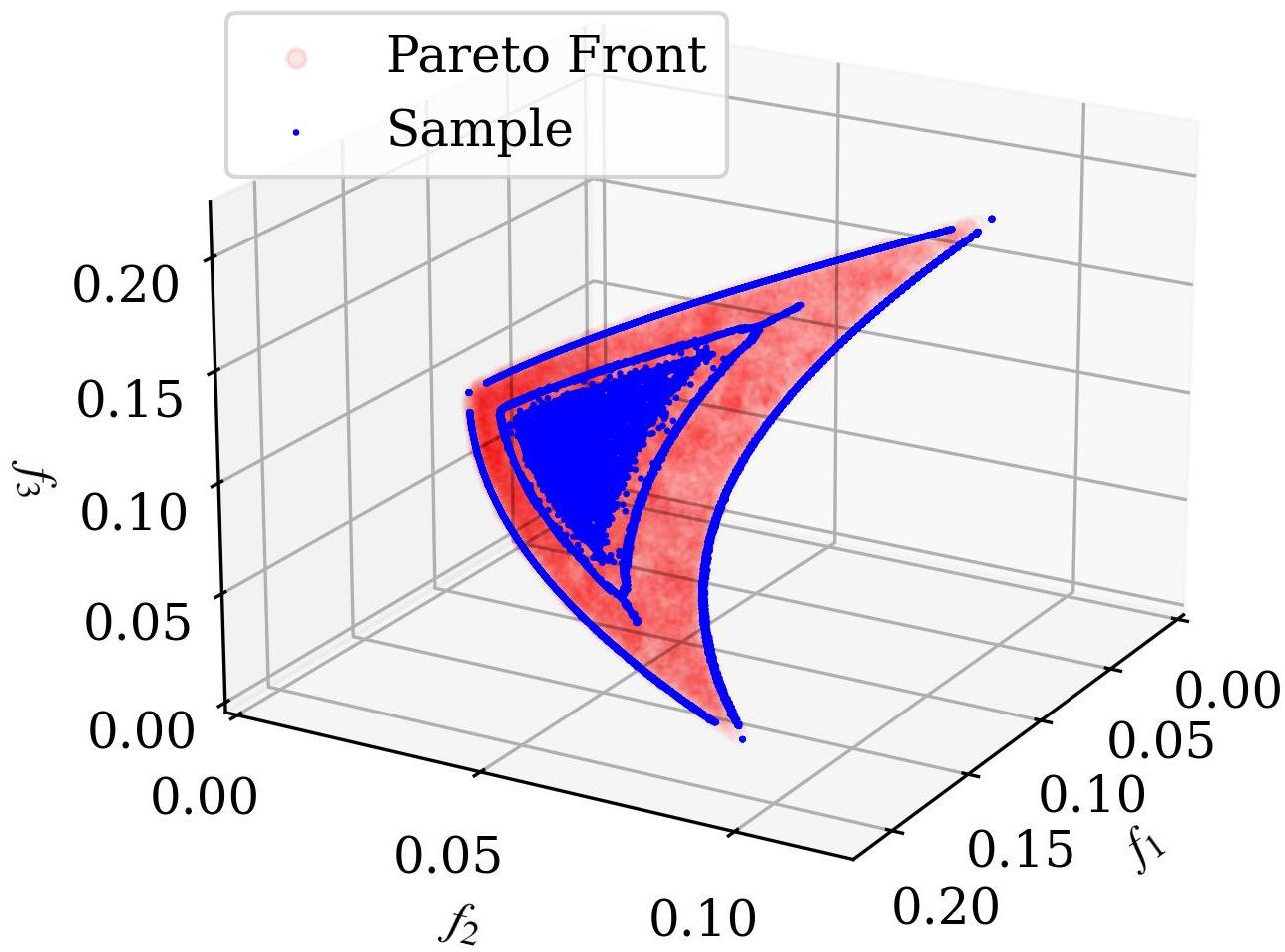}}
	\centerline{\small(c) $m$-MGD}
	\end{minipage}
\caption{\label{fg:3obj_50k} {Approximation of Pareto front of various methods on CIFAR10 optimized with three objectives. The first row presents 50,000 generated samples while the second row presents non-dominated points out of 50,000 sample points, verifying the HV results obtained in Table~\ref{tb:results}.}}\vskip-0.08in
\end{figure}

\section{\revision{More Experimental Settings and Analyses}}\label{app_b}
{\color{black}{
\paragraph{Image Generation} According to~\citet{ishibuchi2013many,li2017multiline}\footnote{\revision{Our problem setting is slightly different as we take the distance square in order to obtain a non-linear shape of the Pareto front. We also refer reviewer to example-1 in~\citet{liu2021profiling} that defines a same two-objective problem but with 1-D decision variable for easy understanding.}}, we can obtain that: (1) the Pareto solutions of the two objective setting are the points on the line between $1_{\Omega}$ and $0.5_{\Omega}$. \revise{Namely, the Pareto solutions are $\{x|x_{\Omega}=\kappa_{\Omega}, \kappa_{\Omega} \in [0.5_{\Omega}, 1_{\Omega}]\}$\footnote{We use $[0.5_{\Omega}, 1_{\Omega}]$ to denote image patches in normalized RGB color values between [0.5, 0.5, 0.5] (grey) and [1, 1, 1] (white).}. When taking images from CIFAR10 based on the Pareto set (Fig.~\ref{fg:cifar_2obj_pf_img}), we follow~\citet{liu2021sampling} to sample images in a small neighborhood around $\kappa_{\Omega}$, namely, $\|x_\Omega-\kappa_\Omega\|_2^2 \leq \epsilon$, where $\epsilon=8\times 10^{-4}$.}
 (2) The Pareto solutions of the three objective setting are the points on the convex polygonal formed by three points $a_{\Omega}, b_{\Omega}, c_{\Omega}$. 
For easy understanding, we assume $\Omega=3\times 1\times1$, which is actually to constrain the middle point of CIFAR10 images to be certain colors.  

We visualize the Pareto front of these two settings in Fig.~\ref{fg:GT}. Specifically, for the two objective setting, the Pareto optimal points lie on the line between [1, 1, 1] and [0.5, 0.5, 0.5] (Fig.~\ref{fg:GT}(a)), which physically denote RGB values (normalized, RGB values [0, 255] divided by 255). Then, we calculate the objectives values $[f_1(x), f_2(x)]$ for these points accordingly, shown in Fig.~\ref{fg:GT}(b). Fig.~\ref{fg:GT}(c) and (d) are plotted for the three objective setting in a similar way. According to their Pareto fronts, we select [0.25, 0.25] and [0.2, 0.1, 0.2] as reference points to calculate the hypervolume (HV) for the two objective setting and the three objective setting in Table~\ref{tb:results}, respectively.

\revise{We sample CIFAR10 image using the constraint with different patch sizes to demonstrate its effect in Fig.~\ref{fg:patch_size}. With a smaller size of the region~$\Omega$, more CIFAR10 images will meet the constraint.} 


\begin{figure}[!thb]\vskip-0.15in
\centering
	\begin{minipage}{0.45\linewidth}
	\centerline{\includegraphics[width=1\textwidth]{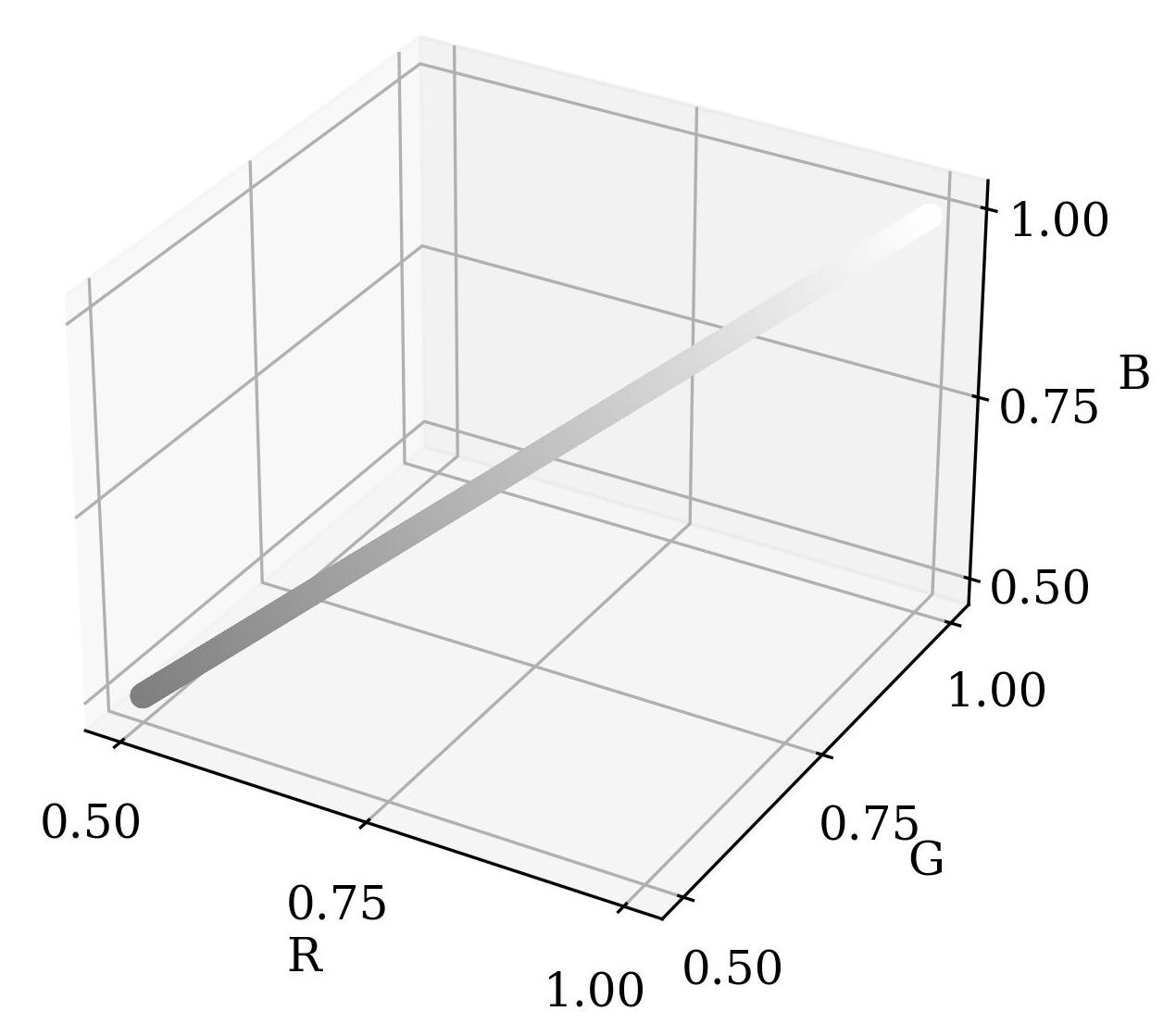}}
 	\centerline{\small(a) Two objectives (data space)}
	\end{minipage}
    \hfill
    \begin{minipage}{0.49\linewidth}
	\centerline{\includegraphics[width=1\textwidth]{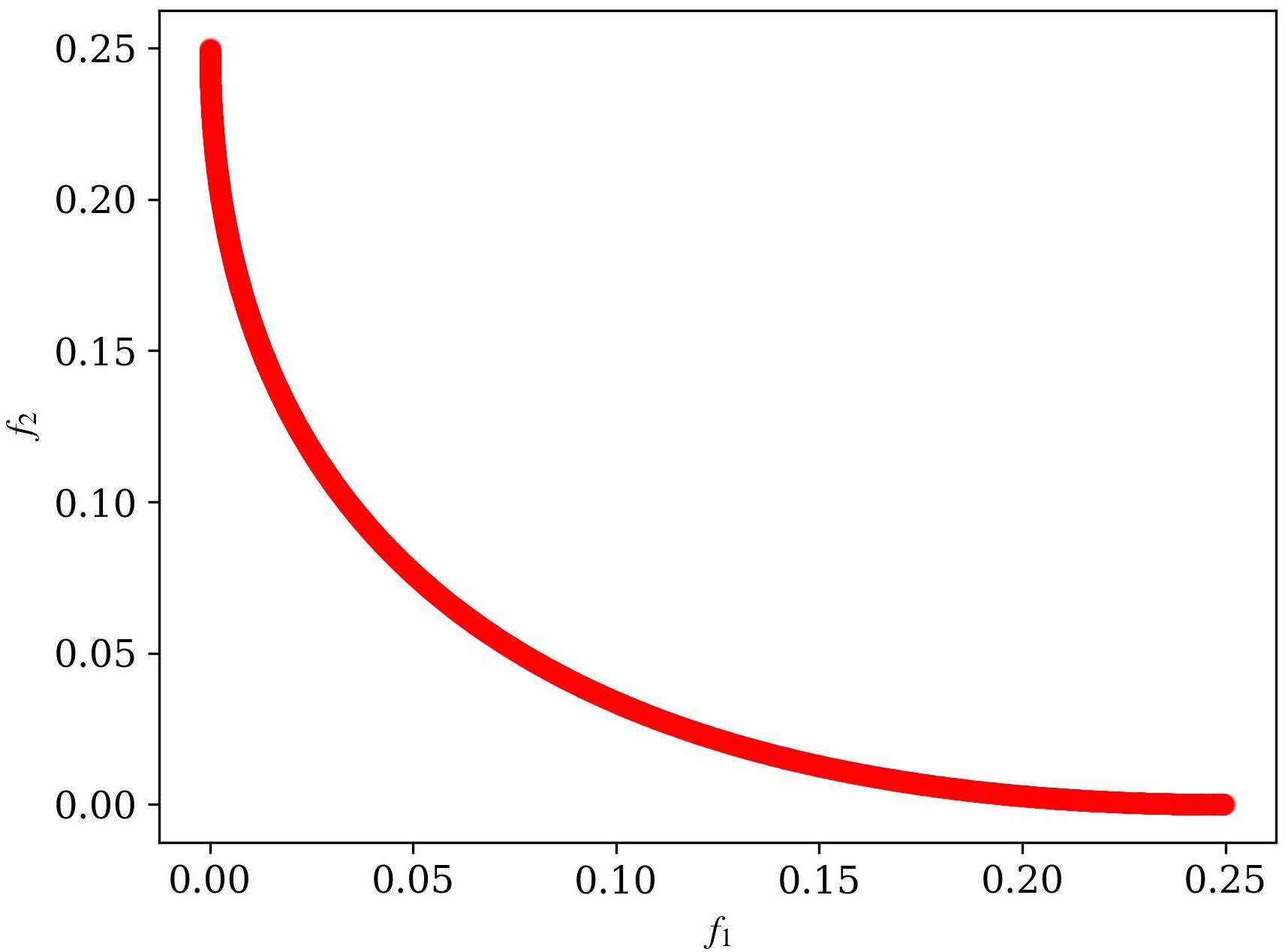}}
 	\centerline{\small(B) Two objectives (functionality space)}
	\end{minipage}
    \hfill
    \begin{minipage}{0.465\linewidth}
	\centerline{\includegraphics[width=1\textwidth]{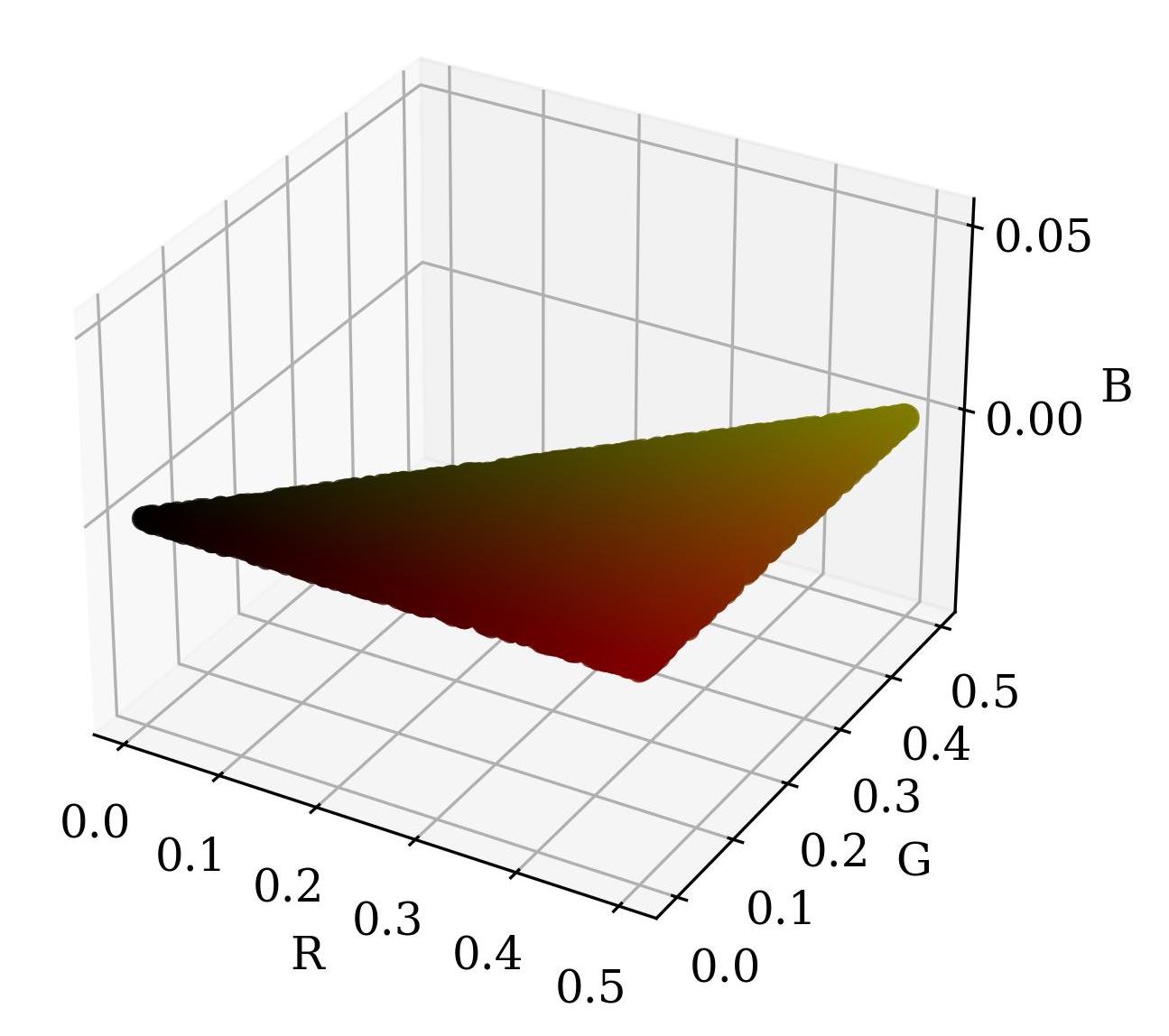}}
 	\centerline{\small(c) Three objectives (data space)}
	\end{minipage}
    \hfill
    \begin{minipage}{0.515\linewidth}
	\centerline{\includegraphics[width=1\textwidth]{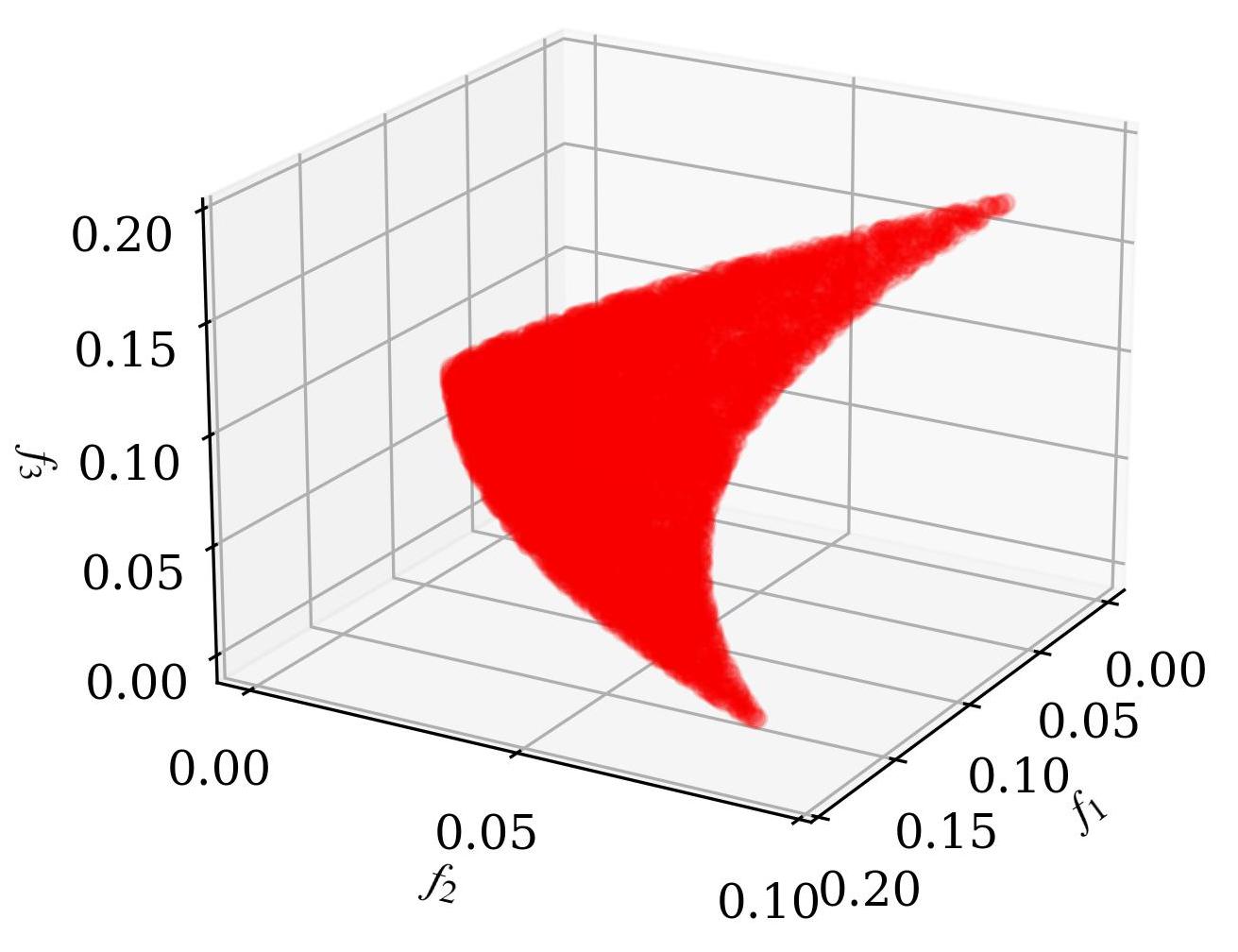}}
 	\centerline{\small(d) Three objectives (functionality space)}
	\end{minipage} 
\caption{\label{fg:GT} \revision{Pareto front of two and three objectives in data space and functionality space optimized for CIFAR10 image generation.}}
\end{figure}
\paragraph{Protein Sequence Generation} Our experiments in Section~5.2 adopted the same dataset and objectives as that in Section~5.2 of~\citet{gruver2023protein}. Note that we did not include their other experiments, because the experiment in their Section~5.1 is not a generation task equipped with property optimization and the dataset for the experiment in Section~5.3 and~5.4 has not been released due to private data. We select $[1\times 10^4, 0]$ as a reference point to calculate the HV for this task.
}}

\paragraph{Justification of Our Experiment Designs} Our experiment designs can appropriately justify the motivation of the MOG problem. Both CIFAR10 and protein datasets are real-world datasets whose data lie on low-dimensional manifolds in high-dimensional space~\citep{krizhevsky2009learning,gruver2023protein}, thus applicable to our MOG problem setting. Meanwhile, the objectives considered for CIFAR10 are indeed benchmark multi-objective optimization problems with clear evaluations~\citep{ishibuchi2013many}; the objectives considered for the protein design task represent real-world scenarios~\citep{gruver2023protein}. Lastly, Fig.~\ref{fg:generated_images} and Table~\ref{tb:results} demonstrate the necessity of considering generation quality, as the generation quality of all baseline methods suffers to some extent when optimizing multiple properties.

\paragraph{\revise{Significant Test}} \revise{We apply the Friedman test under the null hypothesis positing that all methods perform similarly, alongside the Nemenyi post-hoc test for pairwise comparisons among the four methods~\citep{demvsar2006statistical}. The number of factors was set to four, given the failure of $m$-MGD to produce qualified samples, leading to its exclusion. The dataset comprised 30 instances, with each of the four methods independently evaluated five times across three datasets, employing two evaluation criteria. The Friedman test shows that $\tau_F=18.24$, greater than the critical value $F_{3,87} = 2.709$ when $\alpha = 0.05$. Therefore, the null hypothesis is rejected, which signifies a statistically significant difference among the four methods at the significance level of 0.05. Subsequent analysis via the Nemenyi post-hoc test in Fig.~\ref{fg:Friedman_test} unequivocally demonstrates that our PROUD exhibits marked superiority over the three baseline methods.}

\begin{figure}[!ht]
    \begin{minipage}{0.53\linewidth}
	\centerline{\includegraphics[width=1\textwidth]{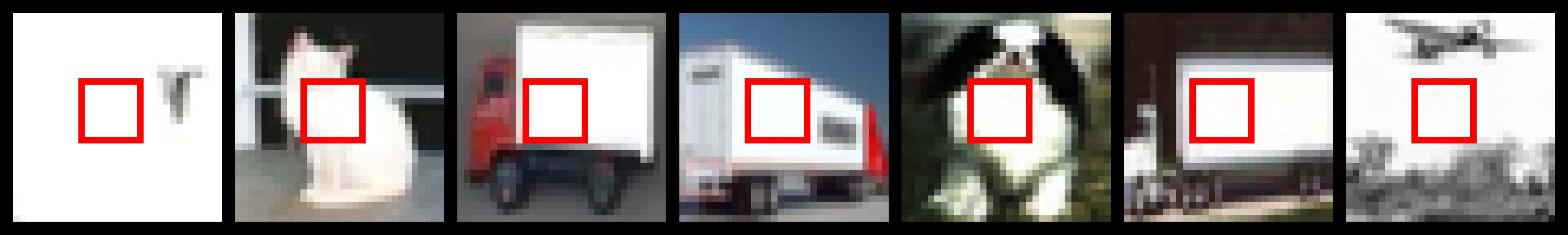}}
 	\centerline{\small[0, 0.25]}
	\end{minipage}
    \hfill
    \begin{minipage}{0.07\linewidth}
	\centerline{\includegraphics[width=1\textwidth]{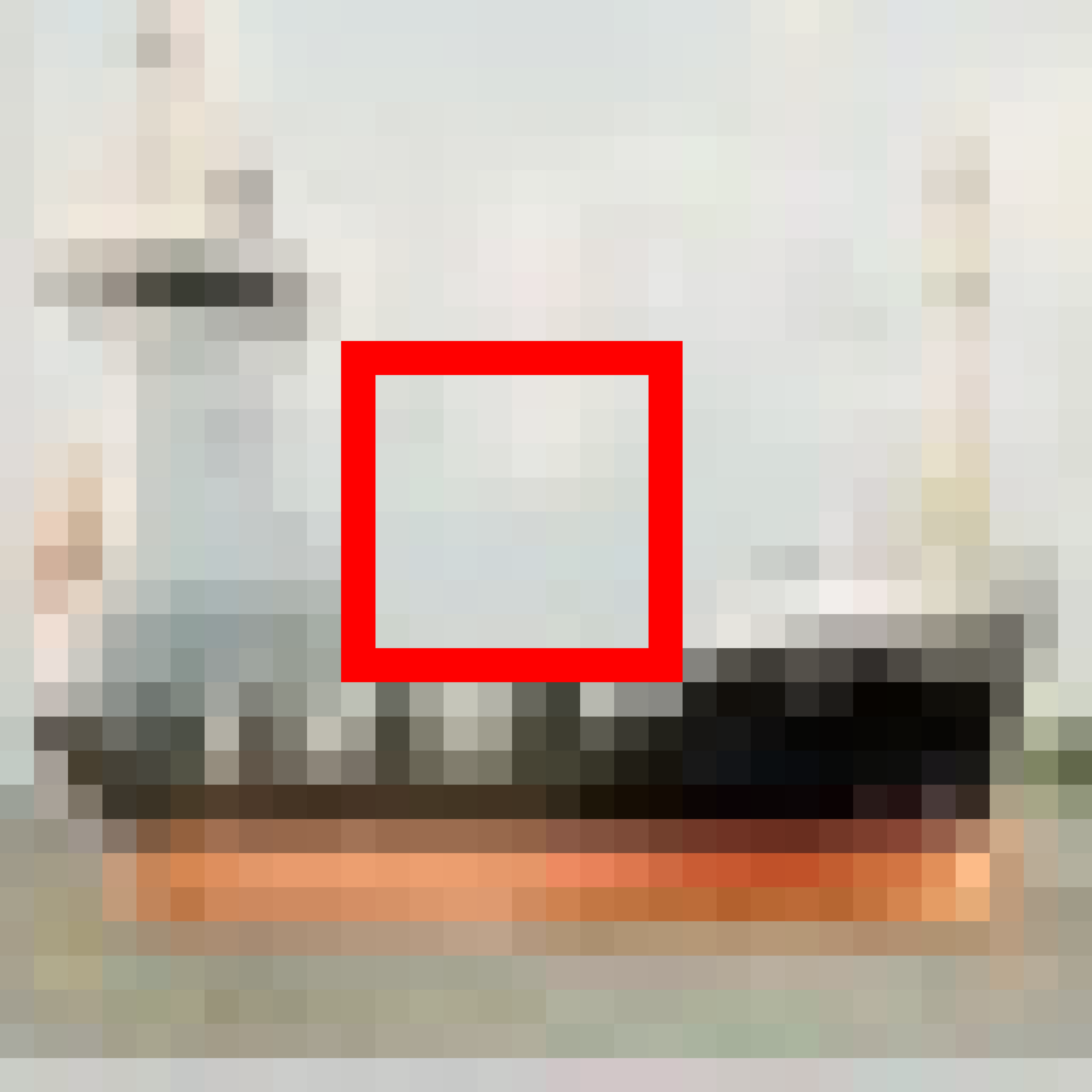}}
 	\centerline{\small[0.025, 0.1225]}
	\end{minipage}
    \hfill
    \begin{minipage}{0.3\linewidth}
	\centerline{\includegraphics[width=1\textwidth]{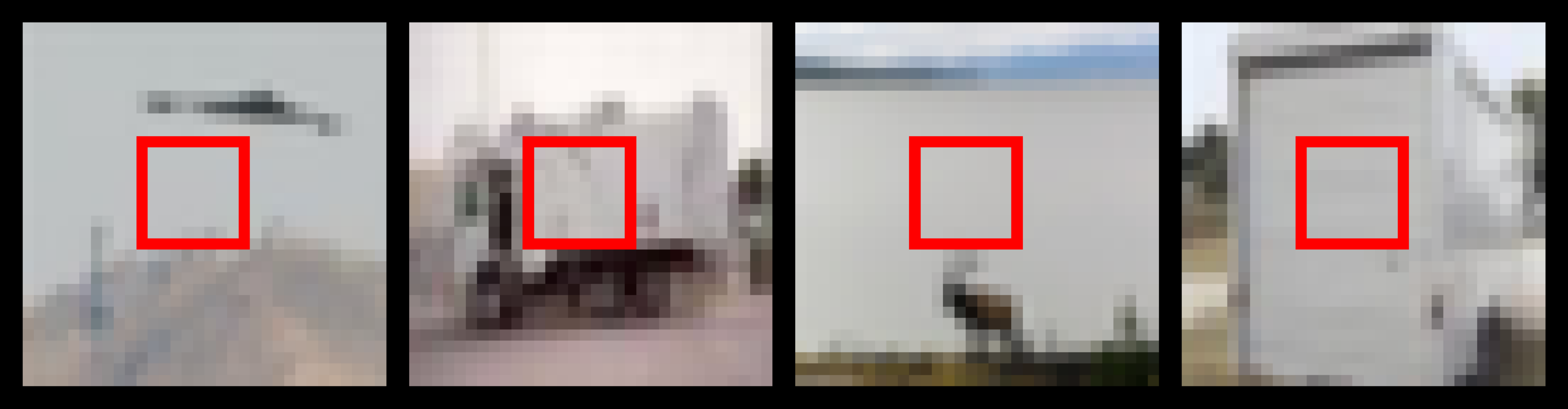}}
 	\centerline{\small[0.0625, 0.0625]}
	\end{minipage}
    \vskip0.1in
    \begin{minipage}{0.3\linewidth}
	\centerline{\includegraphics[width=1\textwidth]{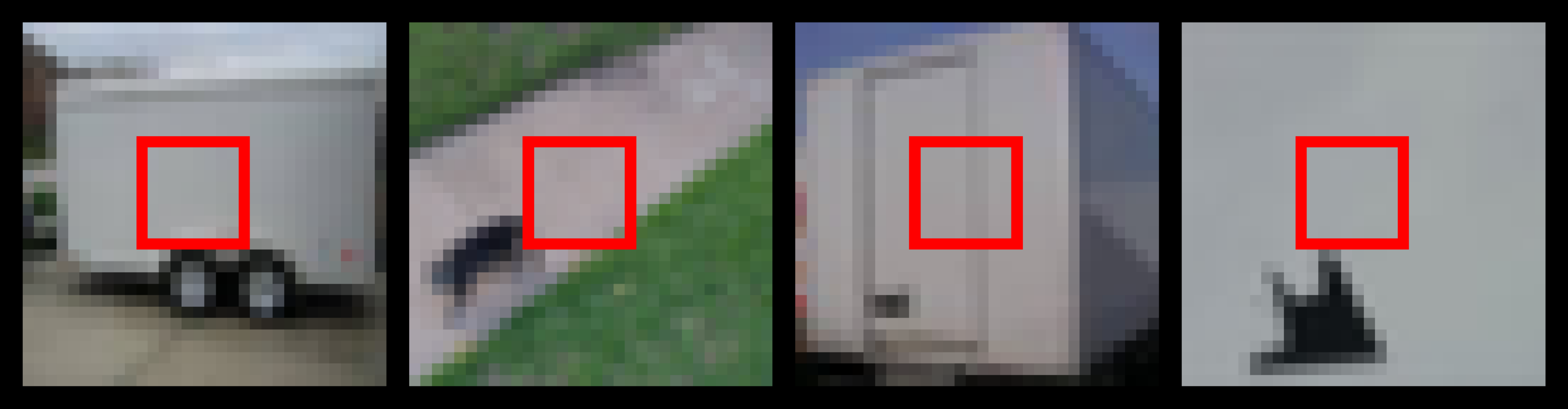}}
 	\centerline{\small[0.140625, 0.015625]}
	\end{minipage}
    \hfill
    \begin{minipage}{0.22\linewidth}
	\centerline{\includegraphics[width=1\textwidth]{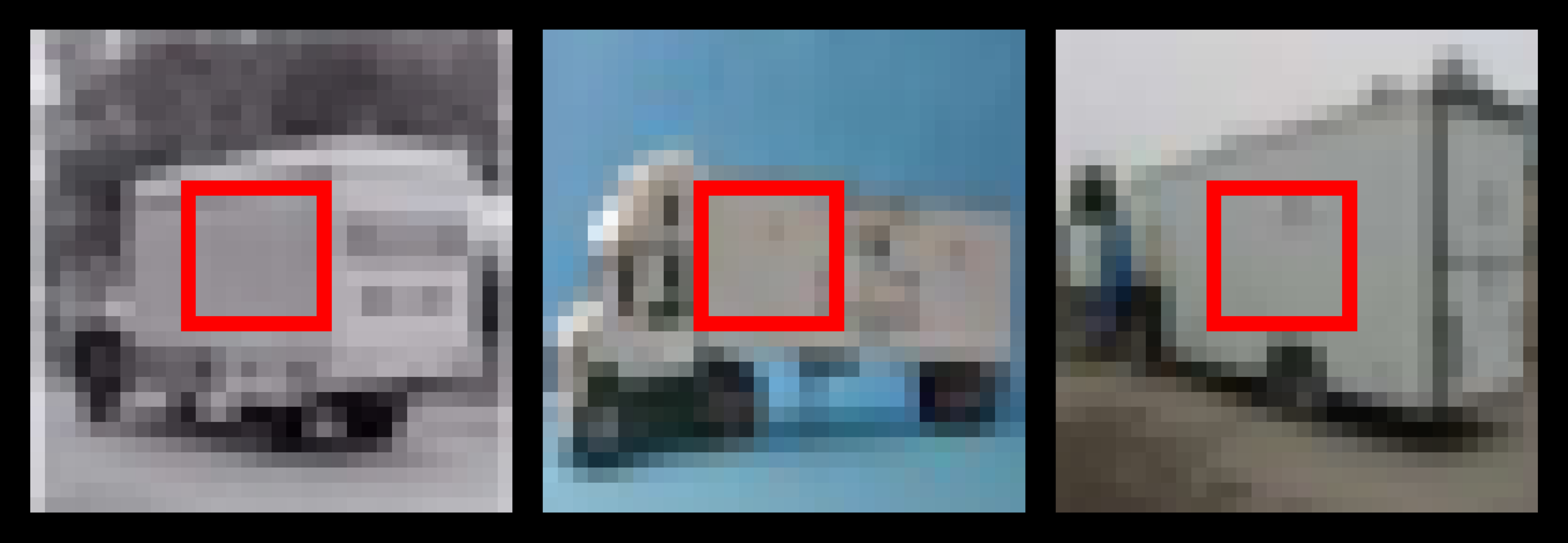}}
 	\centerline{\small[0.180625, 0.005625]}
	\end{minipage}
    \hfill
    \begin{minipage}{0.15\linewidth}
	\centerline{\includegraphics[width=1\textwidth]{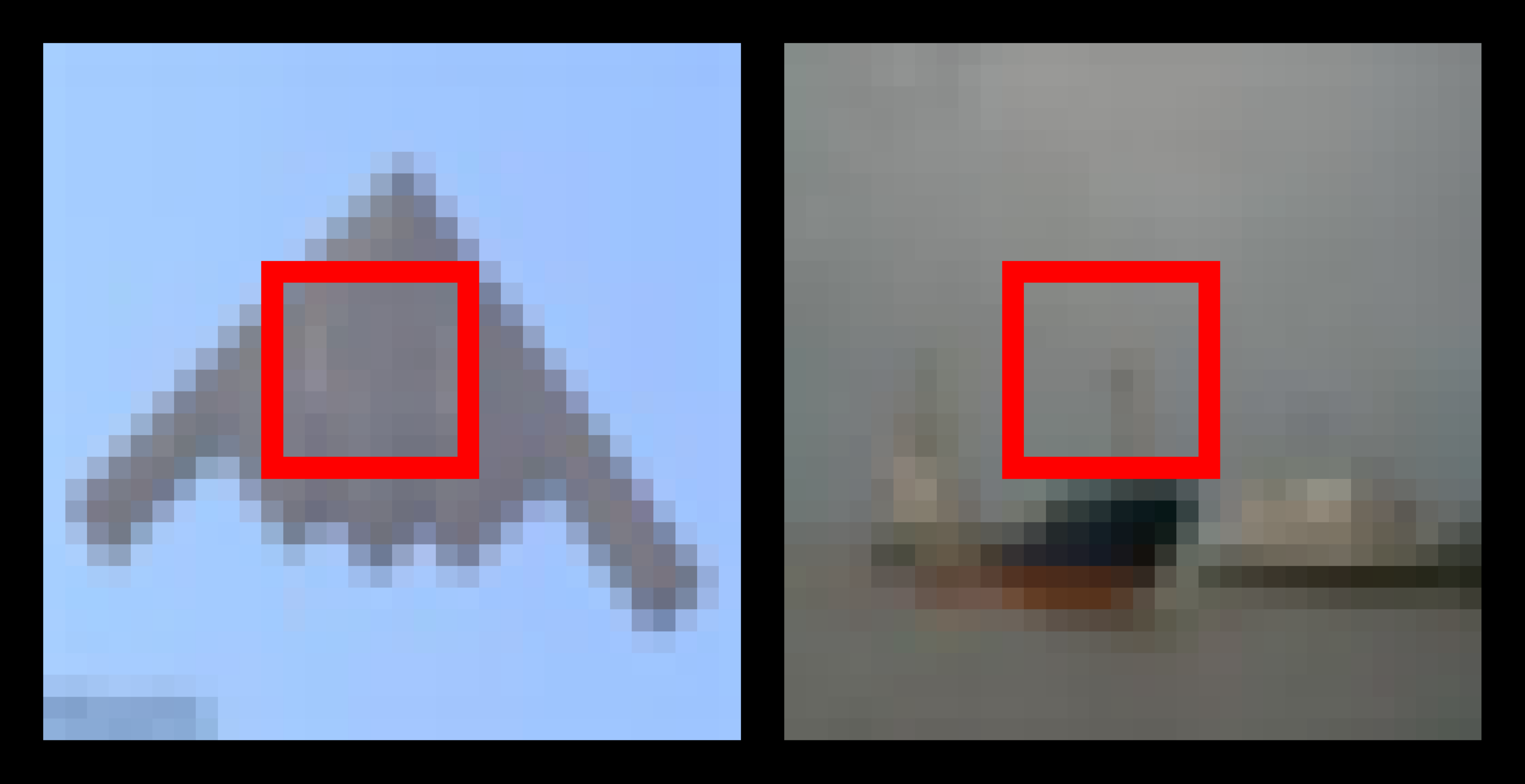}}
 	\centerline{\small[0.25, 0]}
	\end{minipage}
    \caption{\label{fg:cifar_2obj_pf_img}\revise{Full resolution CIFAR10 images ($3\times32\times32$) in Fig.~\ref{fg:motivation}(b) of the manuscript. The red box denotes the region~$\Omega$ ($3\times8\times8$) in the two objectives in Section~5.1.}}	
\end{figure}

\begin{figure}[!ht]
    \begin{minipage}{\linewidth}
	\centerline{\includegraphics[width=0.75\textwidth]{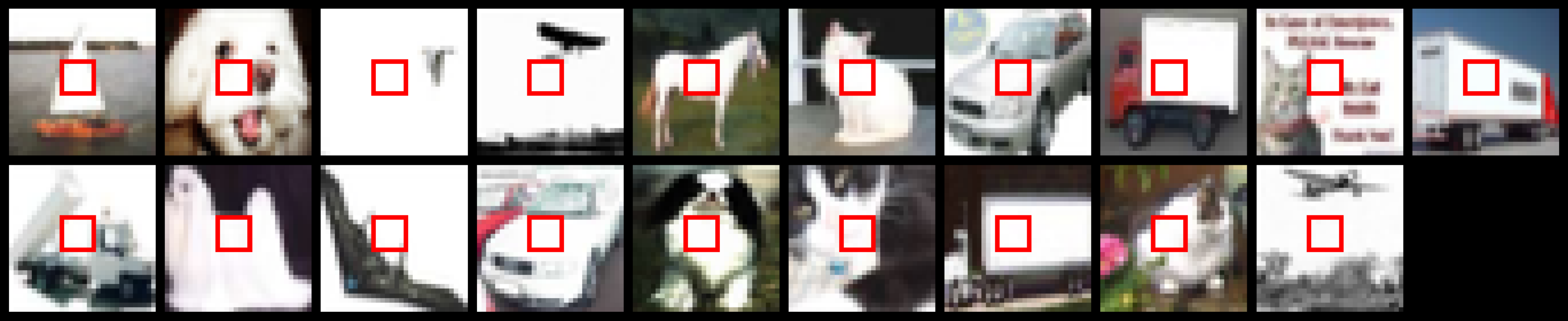}}
 	\centerline{$\Omega$ ($3\times6\times6$, 19 images)}
	\end{minipage}
    \hfill
    \begin{minipage}{\linewidth}
	\centerline{\includegraphics[width=0.5\textwidth]{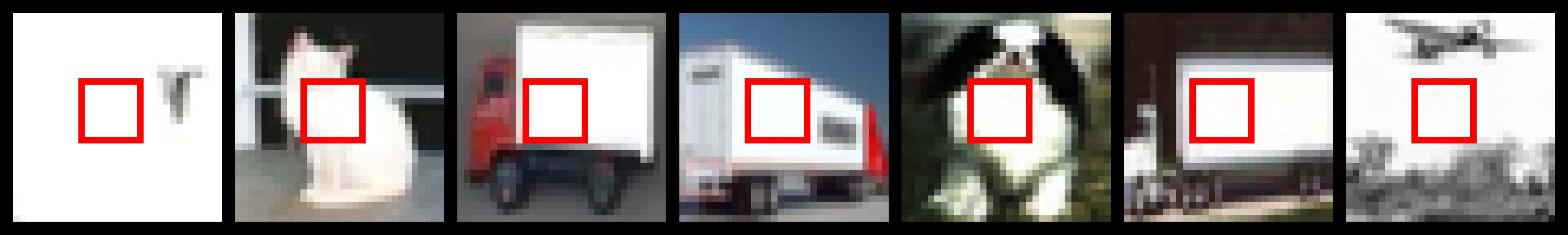}}
 	\centerline{$\Omega$ ($3\times8\times8$, 7 images)}
	\end{minipage}
    \hfill
    \begin{minipage}{\linewidth}
	\centerline{\includegraphics[width=0.2\textwidth]{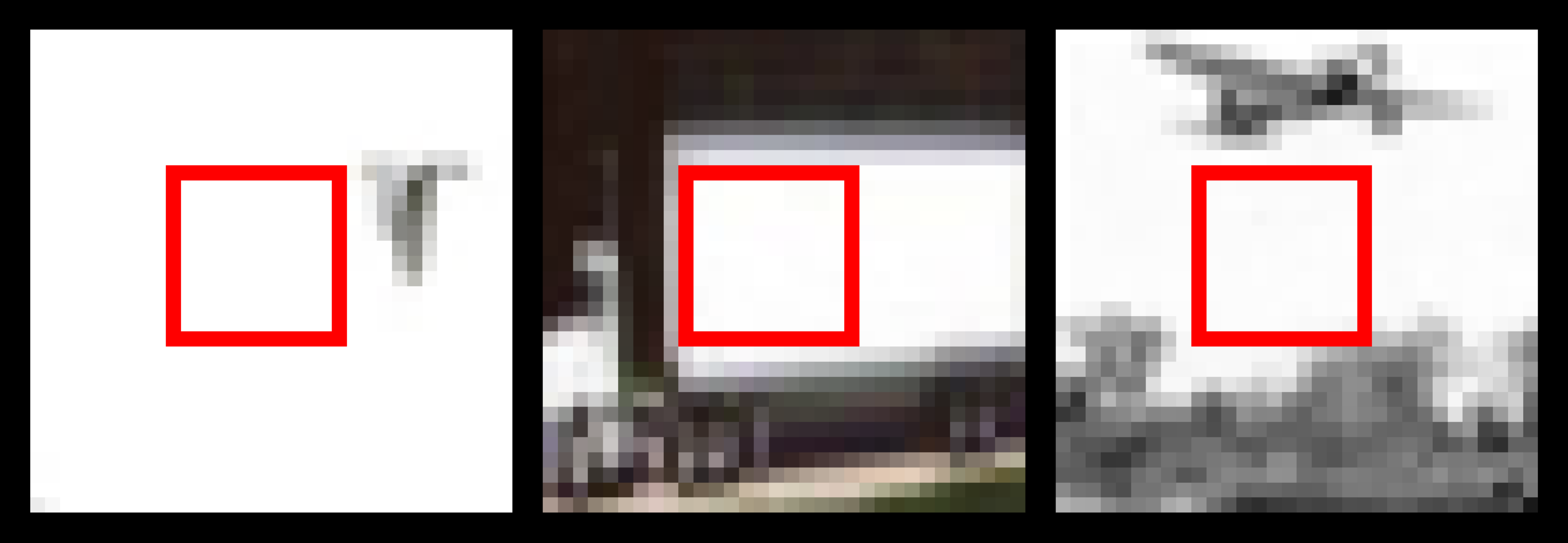}}
 	\centerline{$\Omega$ ($3\times10\times10$, 3 images)}
	\end{minipage}
    \caption{\label{fg:patch_size}\revise{Sampling CIFAR-10 images with regions of different patch sizes.}}	
\end{figure}

\begin{figure}[!t]
	\centerline{\includegraphics[width=0.65\textwidth]{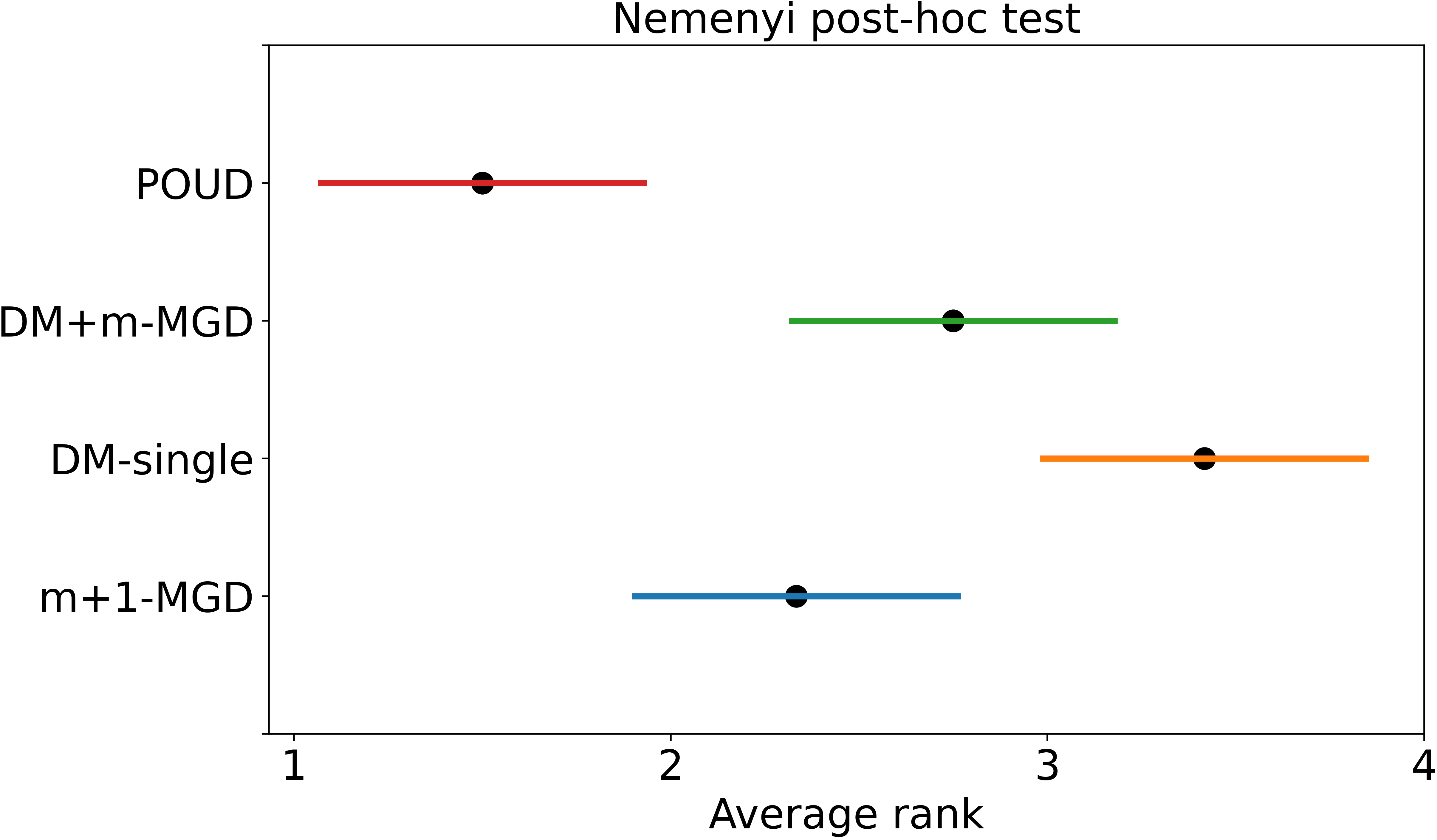}}
    \caption{\label{fg:Friedman_test} \revise{Nemenyi post-hoc test over four methods.}}	
\end{figure}


\section{Discussions}
The constrained MOO problem defines its decision space $S$ on a constrained space expressed using specified linear, nonlinear, or box constraints~\citep{afshari2019constrained,desideri2018quasi} in $\mathbb{R}^d$. Consequently, it is different from our MOG problems, whose manifold is delineated by a given dataset $\mathcal{X}$. Nevertheless, MOG problems could be understood as a type of constrained MOO problem in a broader context.

\begin{table}[!ht]
\centering
\caption{Comparison of the MOG problem with the relevant MOO problems. The generation quality in MOG is usually modeled based on a given dataset $X\subset \mathcal{X}$, where $\mathcal{X}$ denotes a low-dimensional manifold embedded in a high dimensional space~$\mathbb{R}^d$. $F(x)=[f_1(x), f_2(x), \ldots, f_m(x)]$.}
\begin{tabular}{|c|c|c|c|}
\hline
 & objectives & decision/data space & generation quality \\ \hline
MOO & $F(x)$ & $x \in \mathbb{R}^d$ & \XSolidBrush \\\hline
\begin{tabular}[c]{@{}c@{}}Constrained \\ MOO \end{tabular}  & $F(x)$ & \begin{tabular}[c]{@{}c@{}}$ x \in S, S \subset \mathbb{R}^d$ defined by \\ (non)linear or box constraints\end{tabular} & \XSolidBrush \\ \hline
MOG & $F(x)$ & $x \in \mathcal{X}, \mathcal{X} \subset \mathbb{R}^d$ & \Checkmark \\ \hline
\end{tabular}
\label{tab:mooVSmog2}
\end{table}




\end{appendices}

\end{document}